\documentclass[lettersize,journal]{IEEEtran}

\usepackage{amsmath,amsfonts}
\usepackage{array}
\usepackage[caption=false,font=normalsize,labelfont=sf,textfont=sf]{subfig}
\usepackage{textcomp}
\usepackage{stfloats}
\usepackage{url}
\usepackage{verbatim}
\usepackage{graphicx}
\def\BibTeX{{\rm B\kern-.05em{\sc i\kern-.025em b}\kern-.08em
    T\kern-.1667em\lower.7ex\hbox{E}\kern-.125emX}}
\usepackage{balance}

\usepackage{cite}
\usepackage{amssymb}
\usepackage{algorithm}
\usepackage{algpseudocode}
\usepackage{adjustbox}
\usepackage{bbm}
\usepackage[table]{xcolor}
\usepackage{tcolorbox}
\usepackage{arydshln}
\usepackage{pifont}
\usepackage{booktabs}
\usepackage{float}
\usepackage{multirow}
\usepackage{makecell}
\usepackage{cuted}
\usepackage{capt-of}
\usepackage[OT1]{fontenc}
\usepackage{paralist}
\usepackage{bm}
\usepackage{mathrsfs}
\usepackage{threeparttable}
\usepackage[commandnameprefix=always]{changes}
\usepackage{ragged2e}
\usepackage{setspace}
\usepackage{soul}
\usepackage[super]{nth}
\usepackage[switch]{lineno}

\usepackage[
    pagebackref=true,
    colorlinks=true,
    bookmarks=false,
    citecolor=blue,
    linkcolor=blue,
    urlcolor=magenta
]{hyperref}
\usepackage[capitalize]{cleveref}

\definecolor{LightCyan}{rgb}{0.5,1,1}
\definecolor{mypink3}{cmyk}{0,0.7808,0.4429,0.1412}
\definecolor{lightgray}{gray}{0.9}
\definecolor{softblue}{RGB}{220,235,255}
\definecolor{obscol}{RGB}{30,90,170}
\definecolor{evidcol}{RGB}{140,40,150}
\definecolor{deccol}{RGB}{20,120,70}
\definecolor{linkorange}{RGB}{200,110,60}

\crefname{section}{Sec.}{Secs.}
\Crefname{section}{Section}{Sections}
\crefname{table}{Tab.}{Tabs.}
\Crefname{table}{Table}{Tables}

\newcommand{\Obs}[1]{{\color{obscol}\bfseries #1}}
\newcommand{\Evi}[1]{{\color{evidcol}\bfseries #1}}
\newcommand{\Dec}[1]{{\color{deccol}\bfseries #1}}

\tcbset{
  exbox/.style={
    colback=gray!15,
    colframe=black,
    boxrule=1pt,
    arc=2pt,
    left=4pt,
    right=8pt,
    top=4pt,
    bottom=4pt
  }
}

\begin{document}

\bstctlcite{IEEEexample:BSTcontrol}

\title{So-Fake: Benchmarking Social-Media Image Forgery Detection}

\author{Zhenglin Huang,
Xiangtai Li,
Xi Yang,
Bei Peng,
Xiaowei Huang,
Baoyuan Wu,  \\
Dacheng Tao,
Ming-Hsuan Yang,
and Guangliang Cheng%
\thanks{Corresponding author: Guangliang Cheng (e-mail: guangliang.cheng@liverpool.ac.uk).
Zhenglin Huang, Guangliang Cheng, and Xiaowei Huang are with the University of Liverpool.
Xi Yang is with the Hong Kong University of Science and Technology.
Bei Peng is with the University of Sheffield.
Baoyuan Wu is with The Chinese University of Hong Kong, Shenzhen.
Xiangtai Li and Dacheng Tao are with Nanyang Technological University
(e-mail: \{xiangtai94,dacheng.tao\}@gmail.com).
Ming-Hsuan Yang is with the Department of Electrical Engineering and
Computer Science, University of California, Merced.}
}


\maketitle

\begin{strip}
  \centering
  \vspace{-35mm}
  \includegraphics[width=1.0\textwidth]{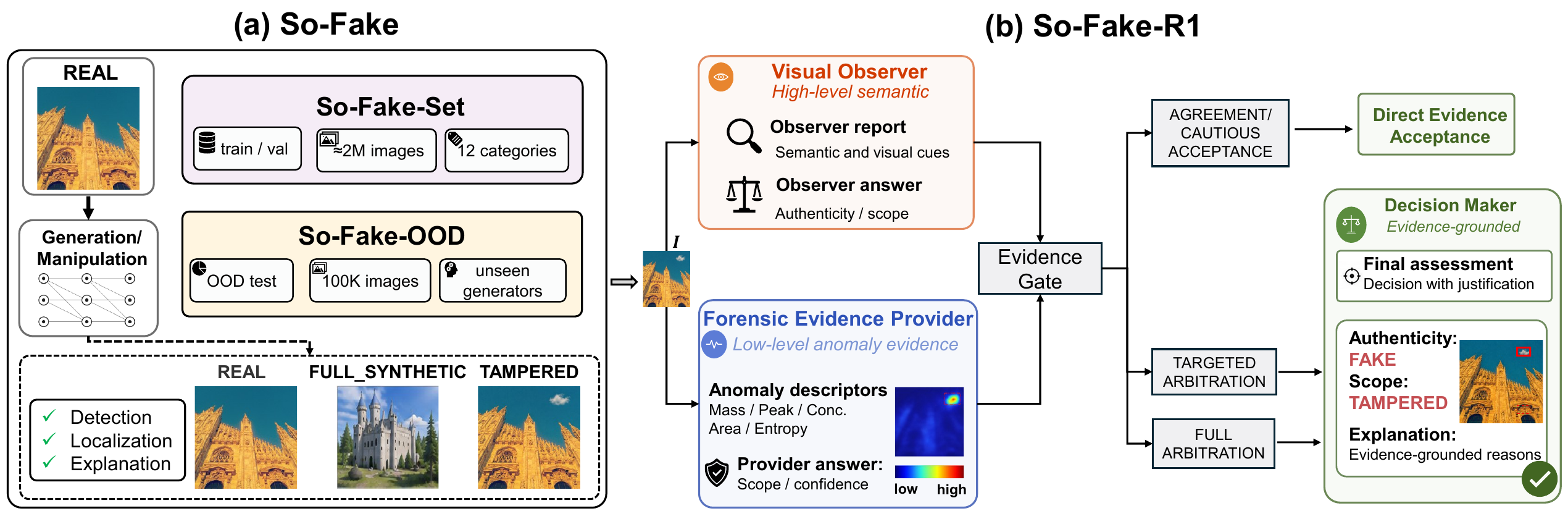}
   \vspace{-2mm}
  \captionof{figure}{\textbf{Overview of So-Fake and So-Fake-R1.} (a) \textbf{So-Fake} comprises \textbf{So-Fake-Set} (training/validation) and \textbf{So-Fake-OOD} (out-of-distribution), spanning three authenticity scopes (REAL, FULL\_SYNTHETIC, TAMPERED) to jointly evaluate detection, localization, and explanation under both in-domain and OOD settings.
(b) \textbf{So-Fake-R1} builds complementary semantic and forensic evidence views with a \textbf{Visual Observer} and a \textbf{Forensic Evidence Provider}. An evidence gate either resolves sufficiently supported cases by direct evidence acceptance or routes conflicting cases to the \textbf{Decision Maker} for image-side arbitration, producing the final authenticity label, forgery scope, localization, and explanation.}
  \label{Figure1}
\end{strip}

\begin{abstract}
Generative models now produce synthetic images that are increasingly difficult to distinguish from authentic social-media content, threatening information integrity and public trust. Robust mitigation requires both representative benchmarks and reliable detectors, yet existing resources often lack broad social-media content, recent high-fidelity generators, joint protocols, and dedicated OOD evaluation. We introduce So-Fake, a comprehensive benchmark for social-media image forgery detection with two complementary components. So-Fake-Set provides approximately 2M training and validation images across 12 semantic categories, spanning authentic images and forgeries generated by 30 diverse synthesis and editing models. So-Fake-OOD provides a 100K held-out benchmark that combines authentic imagery from five social-media platforms with forgeries from 15 disjoint commercial generators, approximating real deployment shifts. Under a unified three-way protocol over REAL, FULL SYNTHETIC, and TAMPERED images, So-Fake jointly evaluates authenticity detection, tampered-region localization, and explanation in both in-domain and OOD settings. Existing MLLM-based detectors typically entangle observation, classification, and explanation in a single generative stream, producing plausible but weakly grounded rationales. We therefore present So-Fake-R1, an evidence-grounded framework with three roles: a Visual Observer for semantic cues, a Forensic Evidence Provider for low-level anomaly evidence, and a Decision Maker for selective arbitration. A deterministic evidence gate accepts sufficiently supported evidence directly and routes conflicting or weakly supported cases to the Decision Maker for arbitration. Extensive experiments establish So-Fake as a challenging benchmark and show that So-Fake-R1 achieves the strongest OOD detection and localization among the evaluated methods, remains competitive in-domain, and obtains the best explanation metrics under our protocol. The code, model, and dataset are publicly available at \url{https://github.com/hzlsaber/So-Fake}.
\end{abstract}


\section{Introduction}
\label{sec:intro}


\IEEEPARstart{T}{he} rapid evolution of generative AI~\cite{DBLP:journals/corr/abs-2406-14555, DBLP:journals/corr/abs-2503-04641} has made it increasingly difficult to verify the authenticity of social-media images, enabling malicious actors to create deceptive content that misleads the public or spreads false information.
This has motivated the creation of large-scale datasets~\cite{DBLP:conf/nips/YanYCZFZLWDWY24,DBLP:journals/corr/abs-2404-01579,DBLP:conf/nips/ZhuCYHLLT0H023} to study and improve forgery detection.
However, they exhibit three key limitations for social-media image forgery detection:
\textbf{1) Narrow Semantic Scope.} Existing datasets~\cite{Ricker_2024,DBLP:conf/wacv/PengHKXMY24,DBLP:conf/iccv/YangCML0XWQW0W023} often focus on restricted domains such as faces, human bodies, or specific object categories, limiting their coverage of complex social-media scenarios.
 \textbf{2) Limited Cross-Domain Evaluation.} Existing datasets lack established protocols for measuring cross-domain generalization and rarely include a dedicated out-of-distribution benchmark.
 \textbf{3) Outdated Generation Quality.} Many widely used benchmarks~\cite{DBLP:conf/nips/ZhuCYHLLT0H023,DBLP:conf/icassp/CorviCZPNV23,DBLP:journals/corr/abs-2408-10072} have outdated generator coverage and rely on earlier synthesis pipelines that may not reflect recent high-fidelity commercial systems.
 While recent works~\cite{Ricker_2024,DBLP:journals/corr/abs-2412-04292,dellanna2025truefakerealworldcase} have attempted to introduce forgery detection datasets for social-media images, they face significant constraints in acquiring authentic platform content.
 Existing benchmarks approximate social-media imagery indirectly, for instance by re-uploading images to social platforms or substituting generic open datasets.
 These proxies fail to capture the fidelity and topical diversity of authentic social-media content, highlighting the need for datasets that more faithfully reflect real-world conditions.

Beyond dataset limitations, existing evaluation protocols also remain inadequate.
Because social-media forgeries range from full synthetic to locally tampered images~\cite{DBLP:journals/corr/abs-2412-04292}, 
benchmarks should evaluate detection, manipulation localization, and evidence-based explanation rather than detection alone.
Despite progress, most methods frame the task as binary classification~\cite{DBLP:conf/nips/YanYCZFZLWDWY24,DBLP:conf/nips/00020YLW23,DBLP:journals/corr/abs-2503-11389}, 
or mask prediction~\cite{DBLP:journals/pami/DongCHCL23,DBLP:conf/cvpr/GuoLRGM023,DBLP:conf/eccv/ZhangCGSB24},
providing little insight into their decisions.
This black-box nature further limits applicability in social-media settings, where diverse manipulations require explanations grounded in visible or forensic evidence.
To address this challenge, recent advances in large language models~\cite{DBLP:journals/corr/abs-2503-24267,DBLP:journals/corr/abs-2503-15264,DBLP:journals/corr/abs-2410-02761, DBLP:journals/corr/abs-2506-07045} have inspired explainability methods that generate human-readable rationales, but such methods depend on costly human annotations and cannot jointly address locally tampered and full synthetic cases.
This motivates a joint evaluation protocol for detection, localization, and evidence-grounded explanation.

To address these limitations, we introduce \textbf{So-Fake}, a comprehensive benchmark for social-media forgery detection with explicit protocols for evaluating detection, localization, and explanation.
As illustrated in Fig.~\ref{Figure1}(a), So-Fake consists of two complementary components.
\textbf{So-Fake-Set} is the main training and validation corpus, comprising approximately \textbf{2M} images across \textbf{12} diverse categories (see Fig.~\ref{Figure2}(a) and Fig.~\ref{Figure3}(a)) and extending beyond traditional face-centric datasets to include humans, animals, events, and other social-media content.
\textbf{So-Fake-OOD} is a \textbf{100K}-image out-of-distribution benchmark that pairs authentic social-media content from Reddit, X, Instagram, Bluesky, and Tumblr with synthetic imagery generated by leading commercial models selected from a public text-to-image leaderboard\footnote{\url{https://artificialanalysis.ai/text-to-image}}.
The \textbf{30} generation and editing models used in So-Fake-Set are entirely disjoint from the \textbf{15} commercial models used in So-Fake-OOD (see Fig.~\ref{Figure2}(b)), reflecting the closed-source nature of many real-world tools and enabling realistic evaluation of generalization to unseen generators.
In both datasets, images are labeled as REAL, FULL\_SYNTHETIC, or TAMPERED, reflecting the major forms of authentic and forged content encountered in social-media scenarios.
The full model list is provided in Appendix~(Sec.~\ref{suppl:generative-methods}).

Building on these complementary datasets, we further investigate how to perform evidence-grounded forgery  detection under the So-Fake benchmark.
A trustworthy authenticity verdict should rest not only on what an image depicts, but also on forensic traces and their consistency with the predicted manipulation type.
When a single generative stream performs all three functions, the contribution of explicit forensic evidence becomes difficult to audit, and plausible rationales may not reveal which cues actually drove the prediction.

Motivated by this, we propose \textbf{So-Fake-R1}, an evidence-grounded framework that assigns each aspect to a dedicated role: a \textbf{Visual Observer} reports semantic content and suspicious appearance cues, while a \textbf{Forensic Evidence Provider} contributes structured forensic descriptors, support regions, and a complementary scope hypothesis.
So-Fake-R1 then performs selective evidence arbitration: clean and consistent evidence is accepted directly, whereas conflicting or unsafe evidence is routed to a \textbf{Decision Maker} for image-side arbitration.
As shown in Fig.~\ref{Figure1}(b), the framework outputs the final authenticity label, forgery scope, optional tampered region, and an evidence-grounded explanation.

The main contributions of this paper are as follows:

\begin{compactitem}
\item We introduce \textbf{So-Fake}, a large-scale social-media forgery detection dataset with two complementary components. 
\textbf{So-Fake-Set} contains approximately 2 million images spanning real, full synthetic, and locally tampered content for training and validation, with generated samples produced by a wide range of generative models.
\textbf{So-Fake-OOD} is a 100K-image out-of-distribution benchmark drawn from real social-media platforms and commercial generators held out from training, designed to approximate deployment distribution shifts.

\item We propose \textbf{So-Fake-R1}, an evidence-grounded framework for social-media forgery detection.
It constructs complementary semantic and forensic evidence views, directly accepts evidence that meets explicit support and scope-consistency criteria through an evidence gate, and routes conflicting, weakly supported, or structurally invalid cases to a \textbf{Decision Maker} for arbitration.
On So-Fake-OOD, it reaches \textbf{72.0\%} balanced accuracy and \textbf{47.8\%} IoU, outperforming the strongest baselines by \textbf{6.4} and \textbf{6.3} points, respectively.

\item Extensive experiments demonstrate that So-Fake is a challenging benchmark and show that So-Fake-R1 achieves the best evaluated OOD performance on authenticity classification and tamper localization while generalizing robustly to out-of-distribution social-media domains.
\end{compactitem}

\section{Related Work}
\label{sec:Relatedwork}
\subsection{Image Forgery Detection Datasets}
\begin{table*}[t]
\caption{\textbf{Comparison with recent image forgery datasets.} ``–" in \#Methods or Latest Fake denotes an unspecified number of generative methods or unspecified latest generator; Column abbreviations: MultiCls = multi-class labels, Expl. = Explanation.}
\label{table1}
\resizebox{0.99\textwidth}{!}{
\begin{tabular}{@{}cccccccccc@{}}
\toprule[1.5pt]
\bf Dataset  & \bf social-media  & \bf Latest Fake  & {\bf \#Methods} & \bf Data Sources &  {\bf \#Images} & \bf MultiCls & \bf Mask & \bf Expl. &\bf OOD \\ \midrule





DMimage~\cite{DBLP:conf/icassp/CorviCZPNV23} (’23) &  \textcolor{red}{\ding{55}}  & DALL-E (’22)   & 10       & COCO, ImageNet       &0.4M+    &  \textcolor{red}{\ding{55}}            & \textcolor{red}{\ding{55}}       &  \textcolor{red}{\ding{55}}   &  \textcolor{red}{\ding{55}}         \\

AIGCD~\cite{DBLP:journals/corr/abs-2311-12397} (’23)  &  \textcolor{red}{\ding{55}}   & Wukong (’22)   & 16 & LSUN, COCO, FFHQ       & 0.7M+  &  \textcolor{red}{\ding{55}}              &  \textcolor{red}{\ding{55}}       &   \textcolor{red}{\ding{55}}    &  \textcolor{red}{\ding{55}}         \\

SynthScars~\cite{DBLP:journals/corr/abs-2503-15264} (’25)  &  \textcolor{red}{\ding{55}} & FLUX (’24)  & - & RichHF-18K, Chameleon, FFAA   &13K   & \textcolor{red}{\ding{55}}    &\textcolor{green}{\ding{51}}        & \textcolor{green}{\ding{51}}      &  \textcolor{red}{\ding{55}}      \\ 

FakeClue~\cite{DBLP:journals/corr/abs-2503-14905} (’25)  &  \textcolor{red}{\ding{55}} & FLUX (’24)  & - & GenImage, FF++, Chameleon   & 0.1M+  & \textcolor{red}{\ding{55}}   &\textcolor{red}{\ding{55}}     & \textcolor{green}{\ding{51}}    &  \textcolor{red}{\ding{55}}        \\ 

GenImage~\cite{DBLP:conf/nips/ZhuCYHLLT0H023} (’23)  &  \textcolor{red}{\ding{55}}  & Wukong (’22)    & 8      & ImageNet        &2M+  & \textcolor{red}{\ding{55}}               & \textcolor{red}{\ding{55}}        &  \textcolor{red}{\ding{55}}    &  \textcolor{red}{\ding{55}}          \\

WildFake~\cite{DBLP:conf/aaai/0001FCLZW025} (’25) &  \textcolor{red}{\ding{55}}  & DALL-E 3 (’23)    & 27      & ImageNet, COCO,  FFHQ, LSUN,  +3 more         &3M+  & \textcolor{red}{\ding{55}}               & \textcolor{red}{\ding{55}}        &  \textcolor{red}{\ding{55}}    &  \textcolor{red}{\ding{55}}          \\

Community Forensics~\cite{DBLP:conf/cvpr/ParkO25} (’25) &  \textcolor{red}{\ding{55}}  & FLUX (’24)    &  4803      & LAION, ImageNet, COCO, FFHQ, +7 more         &2M+  & \textcolor{red}{\ding{55}}               & \textcolor{red}{\ding{55}}        &  \textcolor{red}{\ding{55}}    & \textcolor{green}{\ding{51}}          \\

SID-Set~\cite{DBLP:journals/corr/abs-2412-04292} (’24)  & \textcolor{green}{\ding{51}}   & FLUX (’24)  & 2 & COCO, Flickr30k, MagicBrush   &0.3M & \textcolor{green}{\ding{51}}              &\textcolor{green}{\ding{51}}        & \textcolor{green}{\ding{51}}      &  \textcolor{red}{\ding{55}}      \\ 

Deepfake-Eval-2024~\cite{Ricker_2024} (’25)  & \textcolor{green}{\ding{51}} &
- & - & X, Tiktok, Instagram & 1975 & \textcolor{red}{\ding{55}}            & \textcolor{red}{\ding{55}}       &  \textcolor{red}{\ding{55}}   &  \textcolor{red}{\ding{55}}         \\
 
TrueFake~\cite{dellanna2025truefakerealworldcase} (’25)  & \textcolor{green}{\ding{51}}   & FLUX (’24) & 8 & FFHQ, FORLAB, Facebook, X, Telegram  & 0.6M+ & \textcolor{red}{\ding{55}}   &   \textcolor{red}{\ding{55}}    &      \textcolor{red}{\ding{55}}  &  \textcolor{red}{\ding{55}}   \\

\cdashline{1-9}

\multirow{2}{*}{\textbf{So-Fake}} & \multirow{2}{*}{\textcolor{green}{\ding{51}}}   & \multirow{2}{*}{GPT-Image-2 (’26)} & \multirow{2}{*}{45} & COCO, Flickr30k, WIDER, Tumblr, OpenForensics        & \multirow{2}{*}{2M+} & \multirow{2}{*}{\textcolor{green}{\ding{51}}} & \multirow{2}{*}{\textcolor{green}{\ding{51}}}  & \multirow{2}{*}{\textcolor{green}{\ding{51}}}   & \multirow{2}{*}{\textcolor{green}{\ding{51}}}\\
                                                  &                              & 
                                                  &
                                                  & Reddit, OpenImages, FFHQ, CelebA, X,
                                                  Instagram, Bluesky&                           

\\
\bottomrule[1.5pt]
\end{tabular}
}
\end{table*}

The landscape of image forgery detection datasets has evolved alongside generative techniques, beginning with a predominant focus on facial manipulations. 
Early datasets such as FaceForensics++~\cite{DBLP:conf/iccv/RosslerCVRTN19}, DFFD~\cite{DBLP:conf/cvpr/DangLS0020}, and  ForgeryNet~\cite{DBLP:conf/cvpr/HeGCZYSSS021} laid the foundation for deepfake detection but focused primarily on GAN-based facial forgeries.
With the advent of diffusion models, the field expanded to general AIGC detection, motivating broader benchmarks such as GenImage~\cite{DBLP:conf/nips/ZhuCYHLLT0H023} and DMimage~\cite{DBLP:conf/icassp/CorviCZPNV23}.
As detection increasingly targets real-world content, recent work has begun constructing social-media forgery datasets~\cite{Rahman2019SMIFDNS,Ricker_2024,DBLP:journals/corr/abs-2412-04292,dellanna2025truefakerealworldcase}.
However, many existing datasets are constructed before the advent of recent high-fidelity commercial generators and still lack real-world social-media diversity.
WildFake~\cite{DBLP:conf/aaai/0001FCLZW025} and Community Forensics~\cite{DBLP:conf/cvpr/ParkO25} further expand scale and generator coverage, but are still based on open repositories rather than authentic social-media distributions and do not provide multi-class labels, tampering masks, or explanations.
In contrast, \textbf{So-Fake} combines a platform-native authentic OOD subset with held-out generated forgeries, three-way labels, editing-region masks, and human-verified reference rationales, as summarized in Tab.~\ref{table1}.

\subsection{Image Forgery Detection, Localization, and Explanation}
Recent work in forgery detection has largely relied on deep neural networks to distinguish real from manipulated content.
Although these methods~\cite{DBLP:journals/corr/abs-2503-11389,DBLP:conf/cvpr/ChenZSLW22} capture subtle visual artifacts and achieve strong performance, they often fail to generalize to new manipulation types or domains.
This has motivated a growing shift toward localization-based approaches that explicitly identify tampered regions. 
Image forgery detection and localization (IFDL) ~\cite{DBLP:journals/pami/DongCHCL23,DBLP:journals/corr/abs-2410-23556,DBLP:conf/cvpr/GuillaroCSDV23,DBLP:conf/iccv/ZhangXBZL0ALSS23,DBLP:journals/tcsv/LiuLCL22} offer finer-grained spatial evidence than global classification, facilitating more detailed analysis of manipulation techniques.
However, existing localization datasets focus almost exclusively on facial forgeries~\cite{DBLP:conf/aaai/WangZZY0025,DBLP:conf/iclr/LiangLLJKC24}, overlooking the much broader range of manipulations prevalent in social-media content.
In parallel with efforts to address data limitations, interpretability has become increasingly important, with VLM-based systems such as ForgeryGPT~\cite{li2024forgerygpt}, SIDA~\cite{DBLP:journals/corr/abs-2412-04292}, FakeShield~\cite{DBLP:journals/corr/abs-2410-02761}, FakeScope~\cite{DBLP:journals/corr/abs-2503-24267}, and LEGION~\cite{DBLP:journals/corr/abs-2503-15264} generating natural-language explanations.
SIDA is closest to our setting. 
However, SID-Set relies on curated public images with two open generators and no held-out OOD evaluation, and its model entangles observation, classification, and explanation in a single generative stream.
In contrast, \textbf{So-Fake-R1} organizes detection, localization, and explanation as evidence-grounded verification.
It constructs complementary records and uses a deterministic gate for direct acceptance or routing, followed by a Decision Maker only for weakly supported cases.

\subsection{Reinforcement Learning in VLMs}
Reinforcement learning can complement supervised imitation by optimizing task-level or preference-based rewards beyond next-token matching.
Among recent advances, methods such as offline actor-critic~\cite{DBLP:conf/icml/SpringenbergA0G24} and Proximal Policy Optimization (PPO)~ \cite{DBLP:journals/corr/SchulmanWDRK17} have proven effective, while Group Relative Policy Optimization (GRPO)~\cite{DBLP:journals/corr/abs-2501-12948} is an appealing alternative when multiple sampled outputs can be compared using automatically computed rewards.
Specifically, GRPO reduces the need for human supervision through rule-based reward mechanisms that guide models toward desired behaviors without relying on extensive annotations.
This annotation-light approach makes GRPO particularly valuable for vision-language models (VLMs)~\cite{pixel_sail,huang2025vision,Mixed_R1,sa2va}, where obtaining detailed human explanations for visual reasoning is especially challenging.
Notable applications include Seg-Zero~\cite{DBLP:journals/corr/abs-2503-06520} for zero-shot segmentation, Visual-RFT~\cite{DBLP:journals/corr/abs-2503-01785} for visual question answering, and VLM-R1~\cite{shen2025vlmr1stablegeneralizabler1style} for robust object detection. 
Unlike prior RL-based VLM methods that mainly target general visual reasoning, visual question answering, or object localization, \textbf{So-Fake-R1} uses reinforcement learning to optimize evidence-grounded forensic decision-making. Specifically, GRPO trains the Decision Maker to arbitrate between semantic observations and structured anomaly evidence, so that detection, localization, and explanation are produced within a unified system.

\section{Benchmark}
\label{sec:Benchmark}

\subsection{Benchmark Design and Protocol}
\label{subsec:overview}

\begin{figure*}[t]
\begin{center}
\centerline{\includegraphics[width=1\linewidth]{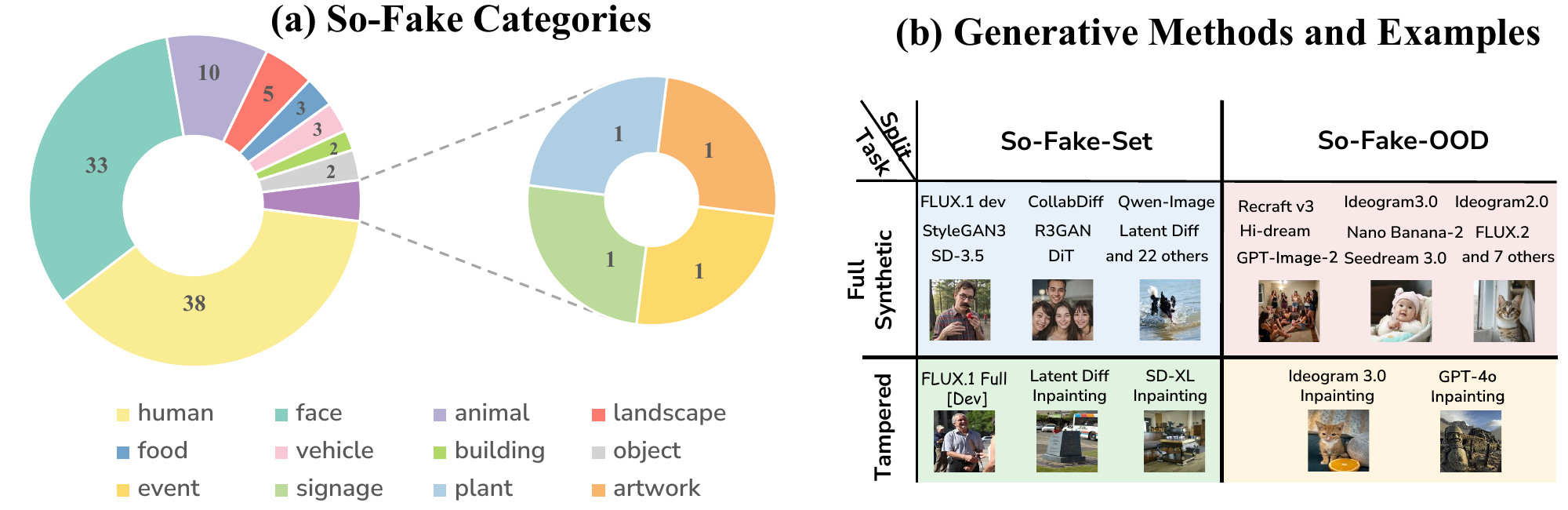}}
\caption{\textbf{Category distribution and generator coverage of So-Fake.}
(a) Distribution over the 12 semantic categories.
(b) Representative generation and editing methods used for full-synthetic
and tampered samples in So-Fake-Set and So-Fake-OOD.}
\label{Figure2}
\vspace{-2mm}
\end{center}
\end{figure*}

Social-media platforms host vast volumes of user-generated images that differ substantially from
standard academic datasets.
Unlike curated benchmarks, these images span highly diverse and informal content categories~\cite{dellanna2025truefakerealworldcase},
and often contain compression artifacts or mixed editing
traces~\cite{DBLP:conf/cvpr/CozzolinoPCNV22,DBLP:journals/corr/abs-2412-04292}.
Moreover, they are increasingly interspersed with synthetic media generated by commercial models such as
GPT-Image-2~\cite{GPT-Image-2}, Seedream~4.5~\cite{Seedream4.5}, and Nano-Banana-2~\cite{Nano-banana-2}.
These properties distinguish social-media forgery detection from conventional in-domain forgery detection: models must remain reliable under heterogeneous content, platform-dependent image statistics, and rapidly evolving manipulation tools.
This motivates a benchmark design that separates two complementary requirements: an open and scalable training corpus, and a held-out OOD testbed that preserves platform-native factors.

To address these requirements, we introduce \textbf{So-Fake}, a benchmark explicitly designed for social-media image forgery detection.
So-Fake comprises two complementary components: \textbf{So-Fake-Set}, a controlled training and validation corpus built from public image datasets, and \textbf{So-Fake-OOD}, a held-out benchmark for cross-domain robustness evaluation.
Both components follow the same three-way authenticity protocol: \textbf{REAL} images are unmanipulated authentic images, \textbf{FULL\_SYNTHETIC} images are entirely generated by image synthesis models, and \textbf{TAMPERED} images contain localized manipulations produced by editing or inpainting models.
This protocol separates full-image synthesis from localized manipulation, enabling evaluation beyond binary real/fake classification and supporting image-level detection, forgery-scope discrimination, and tamper localization.

Both So-Fake-Set and So-Fake-OOD use a unified 12-class taxonomy spanning humans, faces, animals, landscapes, food, vehicles, buildings, objects, events, signage, plants, and artwork (Fig.~\ref{Figure2}(a)).
Each image is assigned a single label by its dominant subject.
The taxonomy provides a controlled basis for separating semantic coverage from domain shift. 
In particular, So-Fake-OOD preserves the same label taxonomy while introducing a compound shift in platform source, generator family, file statistics, and tamper geometry.
This design allows us to characterize whether the OOD split preserves comparable coverage while introducing measurable distributional shift.

Concretely, \textbf{So-Fake-Set} pairs diverse public datasets with systematically generated forgeries to provide an open, scalable training corpus.
\textbf{So-Fake-OOD} combines authentic images from real social-media platforms with forgeries produced by a generator pool that is entirely disjoint from that of So-Fake-Set.
This split design reflects two deployment-oriented sources of shift: authentic platform imagery and unseen commercial generators (Fig.~\ref{Figure2}(b)).
A complete list of the models used to construct So-Fake is provided in Sec.~\ref{suppl:generative-methods} of the Appendix.
%

%

\subsection{Data Sources and Split Construction}
\label{subsec:DataCollection} 

\begin{figure*}[t]
\begin{center}
\centerline{\includegraphics[width=2\columnwidth]{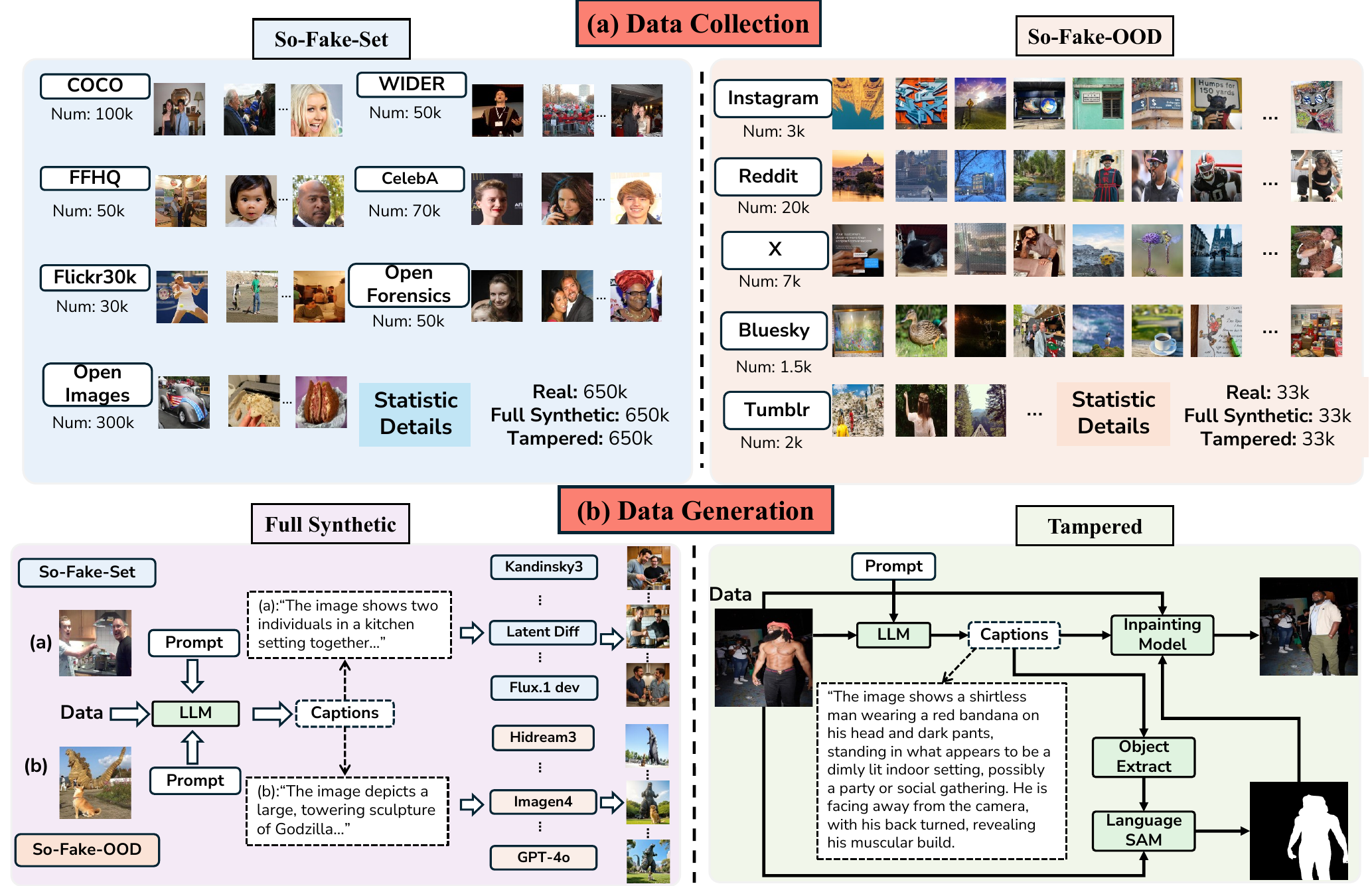}}

\caption{\textbf{Data collection and generation pipelines of So-Fake.}
(a) Data sources, representative examples, and dataset statistics for
So-Fake-Set and So-Fake-OOD.
(b) Pipelines for constructing full-synthetic and tampered images.}
\label{Figure3}
\vspace{-2mm}
\end{center}
\end{figure*}

\noindent\textbf{So-Fake-Set.} 
We construct So-Fake-Set from public image datasets to support large-scale, reproducible benchmark construction.
Specifically, we select real images from COCO~\cite{DBLP:conf/eccv/LinMBHPRDZ14}, Flickr30k~\cite{DBLP:journals/ijcv/PlummerWCCHL17}, OpenImages~\cite{DBLP:journals/corr/abs-1811-00982}, and WIDER~\cite{DBLP:conf/cvpr/XiongZLT15}, as these datasets contain complex scenes involving humans, animals, diverse environments, and daily activities commonly observed in social-media content.
We further incorporate CelebA~\cite{DBLP:conf/iccv/LiuLWT15}, OpenForensics~\cite{DBLP:conf/iccv/LeNYE21}, and FFHQ~\cite{DBLP:journals/pami/KarrasLA21} to ensure sufficient coverage of facial content, which remains a frequent subject of online image manipulation.
In total, So-Fake-Set comprises approximately 650K real images, 650K full synthetic samples, and 650K tampered samples, as shown in Fig.~\ref{Figure3}(a) (left).
The split is grouped by source image to prevent content leakage.

\noindent\textbf{So-Fake-OOD.} 
For the OOD benchmark, we collect authentic images from five social-media platforms: Reddit, Bluesky, Tumblr, X, and Instagram.
These platforms provide diverse and informal user-generated content across our 12 predefined categories, with visual styles, compression patterns, and topical distributions that differ from the public datasets used in So-Fake-Set.
This design introduces platform-dependent distributional factors for evaluating whether detectors trained on So-Fake-Set can transfer to native social-media imagery.
To ensure compliance with platform terms and usage policies, we follow the content policies of each platform.
We release images from Reddit, Bluesky, and Tumblr only where redistribution is compatible with the relevant platform policies and our release protocol; for X and Instagram, we release source hyperlinks rather than image files.
From this collection, we retain roughly 33K real images and generate corresponding full synthetic and tampered samples, yielding about 33K images per class, as shown in Fig.~\ref{Figure3}(a) (left).

\begin{table}[t]
\centering
\scriptsize
\setlength{\tabcolsep}{1.5pt}
\caption{\textbf{Comparison of image-captioning models for prompt generation.}
We report CLIP similarity between generated captions and images on 1,000 randomly sampled images.}
\begin{adjustbox}{width=1.0\columnwidth}
\begin{tabular}{lccccc}
\toprule
\cellcolor{lightgray!30} Metric &
\cellcolor{lightgray!30} Qwen2.5-VL-7B &
\cellcolor{lightgray!30} InternVL2-7B &
\cellcolor{lightgray!30} BLIP-2 &
\cellcolor{lightgray!30} InstructBLIP &
\cellcolor{lightgray!30} LLaVA-7B \\
\midrule
CLIP Score &
\textbf{0.3361$\pm$0.034} &
0.3258$\pm$0.034 &
0.3047$\pm$0.036 &
0.2996$\pm$0.034 &
0.2974$\pm$0.037 \\
\bottomrule
\end{tabular}
\vspace{-2mm}
\end{adjustbox}
\label{table_caption}
\end{table}

\subsection{Data Generation}
\label{subsec:DataGeneration} 

We generate two types of forgeries: \textbf{full synthetic} images and \textbf{tampered} images.
Full synthetic images are generated entirely by generative models, whereas tampered images contain localized manipulations applied to selected regions of otherwise authentic images.
Fig.~\ref{Figure3}(b) illustrates the two generation pipelines.

\noindent\textbf{Full Synthetic Images.}
To produce full synthetic images, we use two categories of generation techniques: \textbf{GAN-based} and \textbf{diffusion-based} models.
For GAN-based methods, we follow the official implementations and sampling protocols provided by the original authors.
For diffusion-based methods, we use a caption-conditioned text-to-image pipeline to approximately match the source-image semantics while varying the generator family.
Specifically, we evaluate five popular image-captioning vision-language models on 1,000 randomly sampled images from So-Fake-Set and compare the resulting image-text alignment using the CLIP similarity score~\cite{DBLP:conf/icml/RadfordKHRGASAM21}.
As shown in Tab.~\ref{table_caption}, Qwen2.5-VL-7B~\cite{DBLP:journals/corr/abs-2502-13923} achieves the highest average score, and we therefore adopt it as the unified caption generator for diffusion-based image synthesis.
The generated captions are then used as prompts for multiple text-to-image models.
For So-Fake-Set, we employ \textbf{30} generative architectures spanning GAN-based and diffusion-based paradigms, selected to cover diverse model families, generation years, and visual styles.
For So-Fake-OOD, we instead use \textbf{15} held-out commercial generators.
These OOD generators are excluded from So-Fake-Set, allowing the benchmark to evaluate transferability to unseen commercial models.

\noindent\textbf{Tampered Images.}
To simulate localized forgeries common in social-media imagery, we generate tampered images through an inpainting-based pipeline.
Given a source image, we first use the same captioner to produce a semantic caption describing its visible content.
An object-extraction module then parses the caption to identify candidate entities for manipulation, such as a person, animal, vehicle, or salient object.
The extracted object labels are supplied to LangSAM~\cite{LangSAM}, which produces segmentation masks for the corresponding image regions.
Each inpainting model then receives the original image, the generated mask, and the caption as inputs, and synthesizes a modified region while preserving the surrounding context, as shown in Fig.~\ref{Figure3}(b)~(right).
The generated masks are retained as ground truth for tamper localization.

For So-Fake-Set, we instantiate this pipeline with FLUX.1 Fill [dev]~\cite{flux2024}, Latent Diffusion~\cite{DBLP:conf/cvpr/RombachBLEO22}, and Stable Diffusion-XL~\cite{DBLP:conf/iclr/PodellELBDMPR24}.
These models are selected to provide visually diverse localized manipulations.
For So-Fake-OOD, we keep the same captioning, object-extraction, and mask-generation pipeline, but replace the inpainting models with held-out commercial editing models, including GPT-4o~\cite{DBLP:journals/corr/abs-2410-21276} and Ideogram~3.0~\cite{Ideogram3.0}.
This separation ensures that OOD tampered images are produced by manipulation models unseen during validation.
%


\begin{figure*}[t]
\centering
\includegraphics[width=0.9\textwidth]{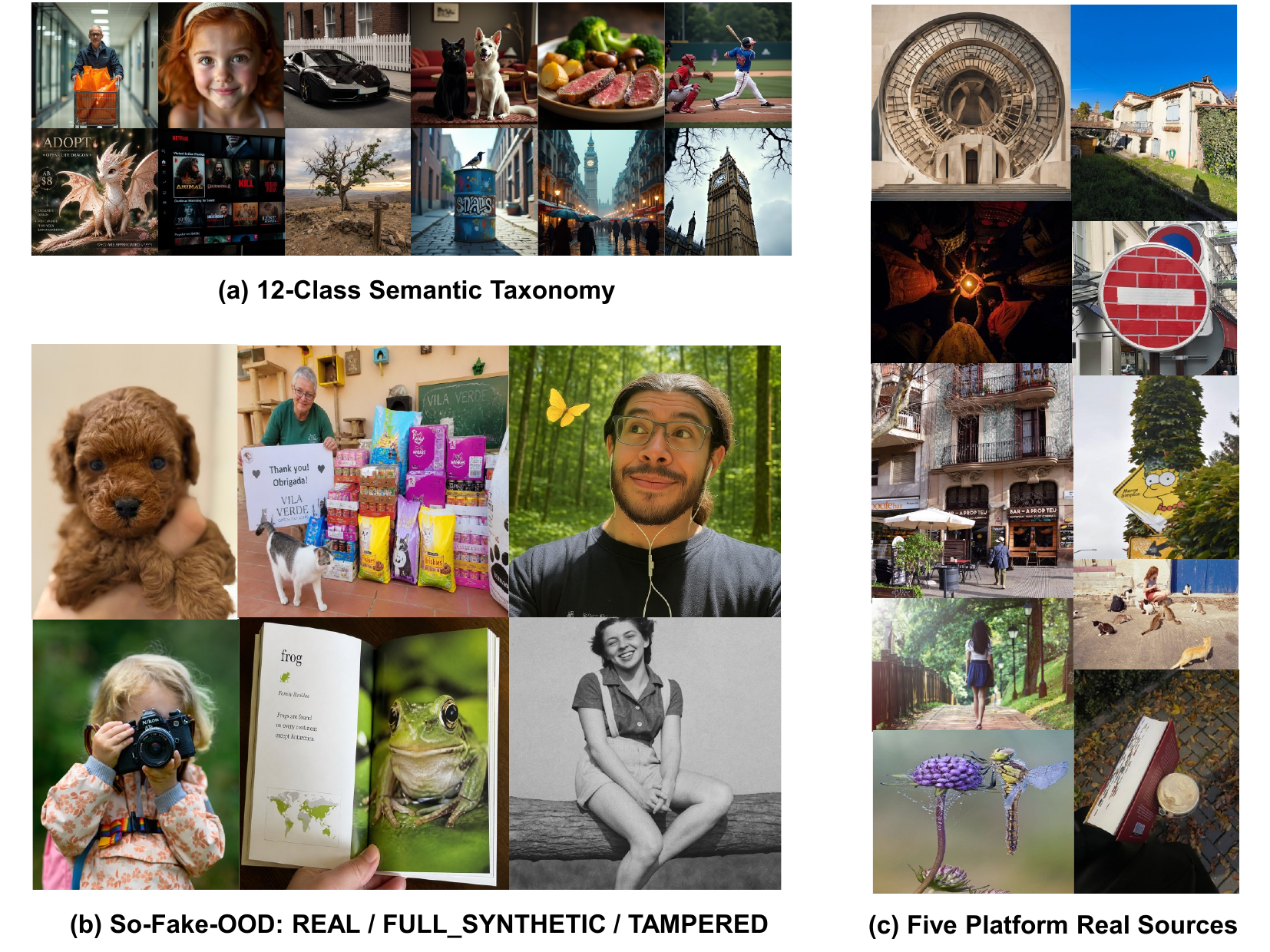}
\caption{
\textbf{Representative benchmark examples of So-Fake.}
(a) Generative examples from the 12 semantic categories used in the benchmark, shown in the order of human, face, vehicle, animal, food, event, artwork, signage, plant, object, landscape, and building.
(b) Representative So-Fake-OOD examples covering the three authenticity scopes: \textbf{REAL}, \textbf{FULL\_SYNTHETIC}, and \textbf{TAMPERED}.
(c) Platform images from the five social-media sources, ordered from top to bottom as Bluesky, Instagram, Reddit, Tumblr, and X.
}
\vspace{-2mm}
\label{fig:examples}
\end{figure*}

\noindent\textbf{Quality Control.}
We filter all generated samples with the LAION aesthetic predictor~\cite{DBLP:conf/nips/SchuhmannBVGWCC22} and discard images with scores below 5.
Because So-Fake-OOD is used as the held-out evaluation benchmark, its generated samples further undergo expert review for realism, semantic consistency, and appropriateness; about 20\% of candidate OOD generations are removed.
Additional filtering details and representative rejection examples are provided in the Appendix (Sec.~\ref{suppl:quality-control}).
%


\begin{table}[t]
\centering
\scriptsize
\setlength{\tabcolsep}{1.5pt}
\caption{\textbf{Descriptive statistics of So-Fake.}
(a) Per-scope long-side resolution and format distributions.
(b) Median mask-area ratio and proportions of small ($<5\%$),
medium ($5\%$--$25\%$), and large ($>25\%$) tampered regions.
(c) Representative examples and masks for the three size tiers.}

\label{table:dataset_stats}

\begin{tabular}{@{}c@{\hspace{2pt}}c@{}}

\begin{minipage}[c]{0.62\columnwidth}
\centering
\begin{adjustbox}{width=1.0\linewidth}
\begin{tabular}{lcc}
\toprule

\cellcolor{lightgray!30} \makecell[c]{Split\\Scope} &
\cellcolor{lightgray!30} \makecell[c]{Long Side\\Min/Med./Max} &
\cellcolor{lightgray!30} \makecell[c]{Format\\J/P/W} \\
\midrule
Train-R & 218 / 512 / 1024   & 66.7 / 33.3 / 0.0 \\
Train-F & 128 / 768 / 1328   & 10.2 / 89.8 / 0.0 \\
Train-T & 512 / 512 / 1024   & 0.0 / 100.0 / 0.0 \\
\midrule
Val-R   & 218 / 512 / 1024   & 67.8 / 32.2 / 0.0 \\
Val-F   & 224 / 768 / 1328   & 9.0 / 91.0 / 0.0 \\
Val-T   & 512 / 512 / 1024   & 0.0 / 100.0 / 0.0 \\
\midrule
OOD-R   & 225 / 1407 / 16320 & 94.8 / 4.7 / 0.5 \\
OOD-F   & 512 / 1024 / 2048  & 11.8 / 88.2 / 0.0 \\
OOD-T   & 960 / 1024 / 1472  & 0.0 / 100.0 / 0.0 \\
\bottomrule
\end{tabular}
\end{adjustbox}

\vspace{0.5mm}
{\footnotesize (a) Image statistics}
\end{minipage}
&
\begin{minipage}[c]{0.34\columnwidth}
\centering
\begin{adjustbox}{width=1.0\linewidth}
\begin{tabular}{ccccc}
\toprule
\cellcolor{lightgray!30} Split &
\cellcolor{lightgray!30} \makecell[c]{Mask\\Med.} &
\cellcolor{lightgray!30} S  &
\cellcolor{lightgray!30} M &
\cellcolor{lightgray!30} L \\
\midrule
Train & 6.9  & 45.7 & 28.8 & 25.6 \\
Val   & 6.9  & 45.7 & 28.8 & 25.6 \\
OOD   & 22.2 & 15.3 & 39.4 & 45.3 \\
\bottomrule
\end{tabular}
\end{adjustbox}
\vspace{0.5mm}
{\footnotesize (b) Tamper statistics}

\vspace{1mm}
\includegraphics[width=\linewidth]{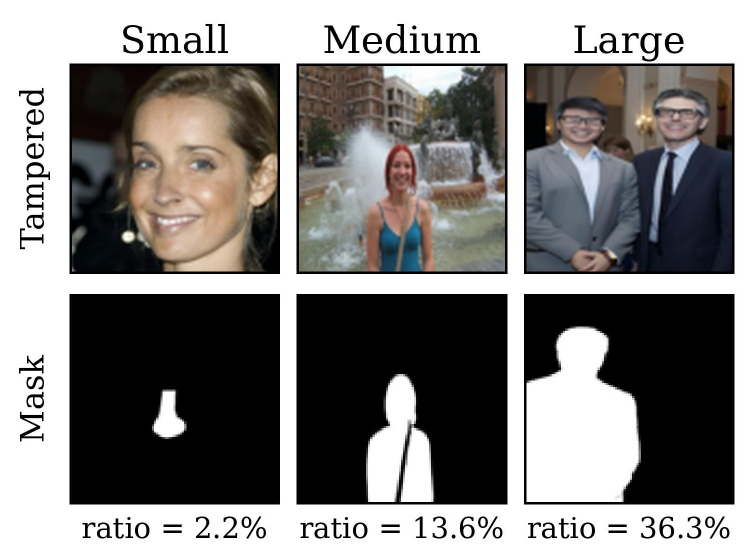}
\vspace{0.3mm}
{\footnotesize (c) Tampered examples}
\end{minipage}

\end{tabular}
\vspace{-2mm}
\end{table}

\subsection{Dataset Statistics and Diagnostics}
\label{subsec:dataset_statistics}
We release each split with a dataset card documenting file organization, release policy, and descriptive statistics.
Fig.~\ref{fig:examples} shows representative examples across the 12 semantic categories, the three So-Fake-OOD authenticity scopes, and the five platform-native real-image sources.

\noindent\textbf{Release format.}
Each split is accompanied by JSONL annotations containing source metadata, release policy, and generator/editor identity when applicable.
Depending on platform policy, samples are released either as image files or as source hyperlinks.
For TAMPERED samples, we provide binary masks and axis-aligned boxes in the released annotation files.

\begin{figure*}[t]
\centering
\includegraphics[width=1\textwidth]{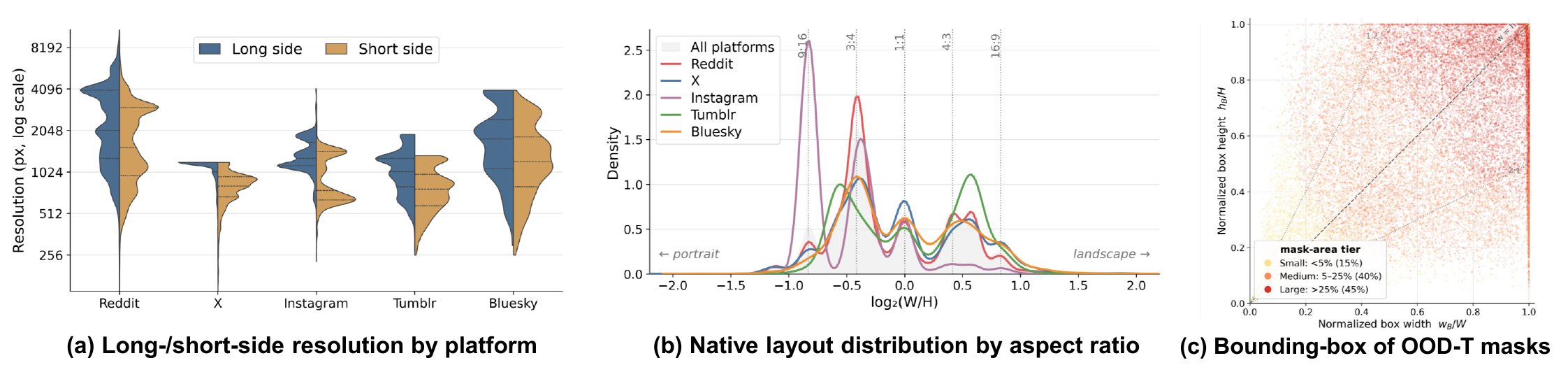}
\caption{
\textbf{OOD diagnostics of So-Fake.}
(a) Long-/short-side resolution distributions of authentic OOD images across platforms.
(b) Native aspect-ratio distributions of authentic OOD images.
(c) Normalized bounding-box width and height of OOD tamper masks.
These diagnostics characterize platform-native image-statistics shifts and diverse localization geometry.
}
\label{fig:ood_diagnostics}
\end{figure*}

\noindent\textbf{Image-level statistics.}
Tab.~\ref{table:dataset_stats}(a) reports the long-side resolution range and file-format distribution for each split and authenticity scope.
So-Fake-Set is relatively standardized in resolution, whereas So-Fake-OOD preserves broader native resolution variation from social-media platforms.
Fig.~\ref{fig:ood_diagnostics}(a) further shows platform-dependent long- and short-side resolution profiles for authentic OOD images, and Fig.~\ref{fig:ood_diagnostics}(b) shows multiple native aspect-ratio modes, including portrait, square, and landscape layouts.
Although native format and resolution distributions are correlated with authenticity scope, they do not constitute a primary detection shortcut: after uniform normalization, the relative method ranking remains unchanged and So-Fake-R1 drops by only 1.7 points in balanced accuracy.
Full results are provided in Appendix Sec.~\ref{suppl:low-level-statistics}.

\noindent\textbf{Tamper-region statistics and geometry.}
For TAMPERED samples, we characterize manipulated regions by the mask area ratio $|M|/(WH)$, where $M$ denotes foreground pixels and $W,H$ are the image width and height.
Samples are grouped into small ($<5\%$), medium ($5$--$25\%$), and large ($>25\%$) regions.
As shown in Tab.~\ref{table:dataset_stats}(b), So-Fake-Set contains many small local edits, whereas So-Fake-OOD shifts toward medium and large manipulated regions.
Fig.~\ref{fig:ood_diagnostics}(c) further shows that normalized OOD tamper boxes span broad scales and aspect ratios, indicating that OOD localization cannot be reduced to fixed-size or fixed-shape prediction.
Representative examples are included in Tab.~\ref{table:dataset_stats}(c).

Together, these diagnostics show that So-Fake-OOD captures deployment-oriented shifts beyond held-out generators.

\section{Method}
\label{sec:Method}

We introduce So-Fake-R1, an evidence-grounded forensic vision-language framework that organizes forgery detection into complementary semantic and forensic evidence views.
As shown in Fig.~\ref{Figure}, a Visual Observer provides an image-level semantic hypothesis, a Forensic Evidence Provider supplies low-level anomaly evidence, and a deterministic evidence
gate either accepts consistent evidence directly or routes conflicting cases to a Decision Maker, which arbitrates using the image
and both evidence records.

\subsection{Visual Observer}
\label{subsec:visual_observer}
The Visual Observer operates exclusively on the original image and produces a semantic evidence record based on visible content.
Its role is not to make the final forensic verdict, but to provide an image-level hypothesis and the visual rationale supporting that hypothesis.
Formally, given an input image $I\in\mathbb{R}^{H\times W\times 3}$, the observer is a mapping $\Phi_{\mathrm{obs}}$
\begin{equation}
    \mathcal{O} = \Phi_{\mathrm{obs}}(I)
    =
    \left(
    r^{\mathrm{obs}},
    \hat{a}^{\mathrm{obs}},
    \hat{s}^{\mathrm{obs}},
    \hat{B}^{\mathrm{obs}}
    \right),
    \label{eq:observer-record}
\end{equation}
where $r^{\mathrm{obs}}$ is the visual reasoning report,
$\hat{a}^{\mathrm{obs}}\in\{\texttt{REAL},\texttt{FAKE}\}$ is the authenticity hypothesis,
$\hat{s}^{\mathrm{obs}}\in\mathcal{Y}$ is the manipulation-scope hypothesis, and
$\hat{B}^{\mathrm{obs}}$ is an optional bounding box when the observer predicts local tampering.
The scope label space is:
\begin{equation}
    \mathcal{Y}=
    \{\texttt{REAL}, \texttt{FULL\_SYNTHETIC}, \texttt{TAMPERED}\}.
    \label{eq:space}
\end{equation}
For downstream arbitration, the observer provides both an authenticity
hypothesis $\hat a_{\mathrm{obs}}$ and a manipulation-scope hypothesis
$\hat s_{\mathrm{obs}}$. REAL corresponds to 
$\hat a_{\mathrm{obs}}=\mathrm{REAL}$ and 
$\hat s_{\mathrm{obs}}=\mathrm{REAL}$; FULL\_SYNTHETIC and TAMPERED correspond to fake images with image-level generation and local manipulation, respectively.

We instantiate $\Phi_{\mathrm{obs}}$ as Qwen2.5-VL-7B~\cite{DBLP:journals/corr/abs-2502-13923}, fine-tuned with LoRA~\cite{DBLP:journals/corr/abs-2106-09685} on So-Fake-Set and conditioned only on the original image.
The resulting tuple $\mathcal{O}$ serves as the semantic evidence record for subsequent routing and arbitration.

\begin{figure}[t]
\centering
\includegraphics[width=\columnwidth]{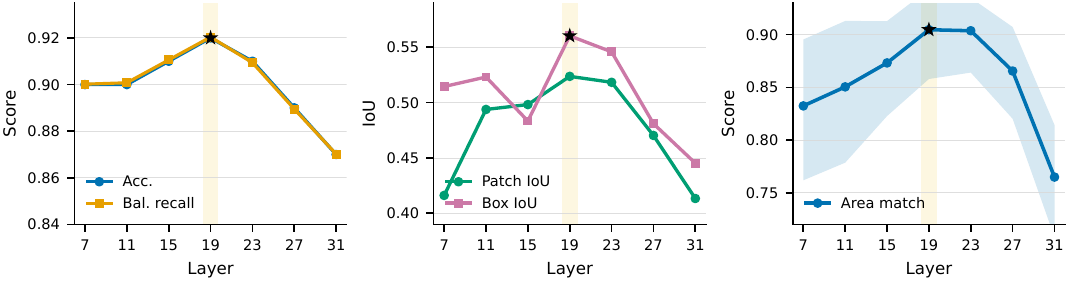}
\caption{\textbf{Layer selection diagnostics for the Forensic Evidence Provider.}
We compare frozen Qwen2.5-VL-7B visual features from different layers on the in-domain validation split using the same FETA readout.
Layer 19 provides the best joint behavior: it achieves the highest scope accuracy and balanced recall, the strongest tampered-support alignment, and a high topology area match.
We therefore use layer 19 as the provider feature source.}
\label{sup:figure5}
\end{figure}

\begin{figure*}[t]
\centerline{\includegraphics[width=\linewidth]{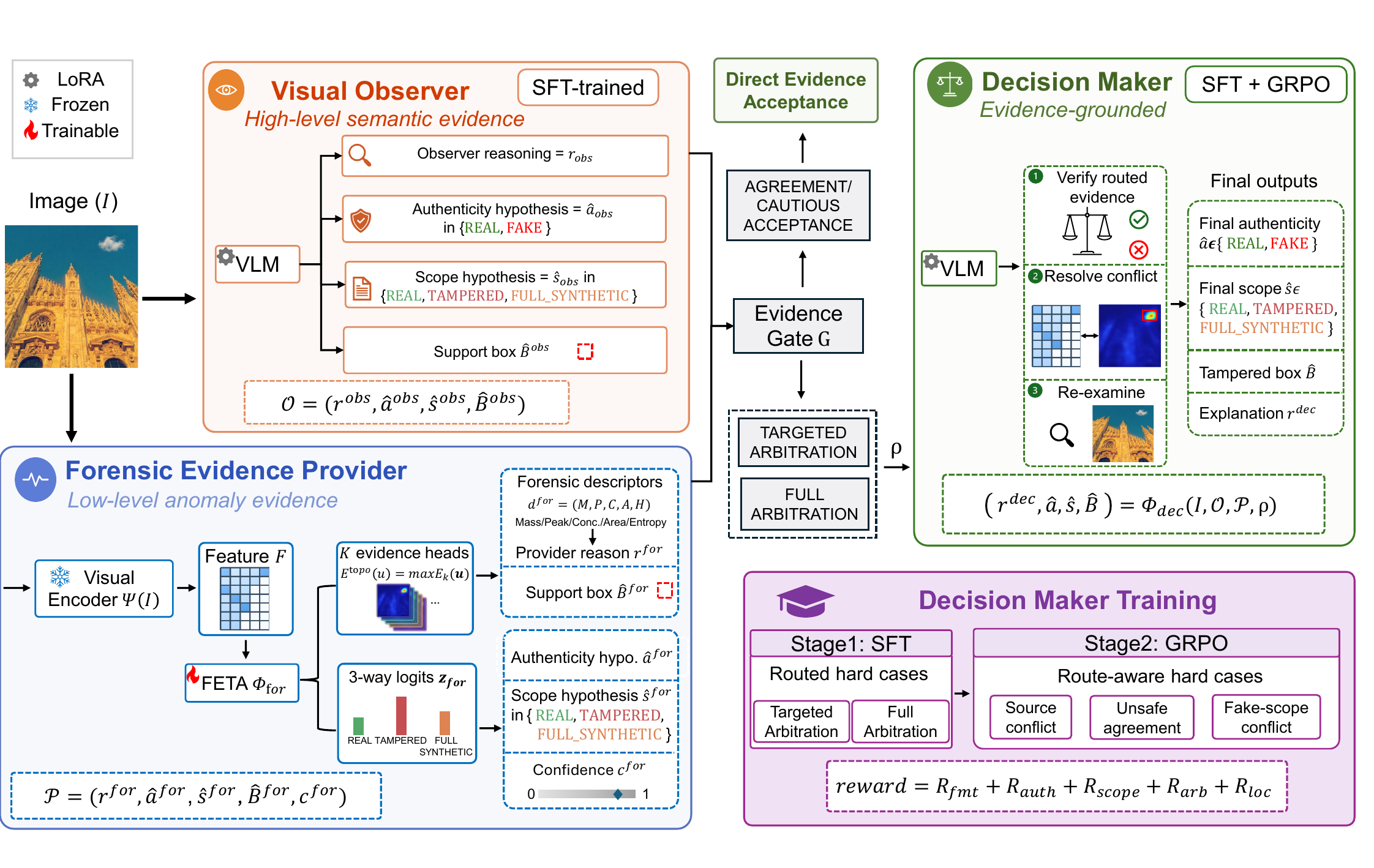}}

\caption{\textbf{Overview of So-Fake-R1.}
The Visual Observer extracts semantic evidence from the input image, and the Forensic Evidence Provider derives low-level anomaly evidence from frozen visual features through FETA.
A deterministic Evidence Gate checks evidence agreement, support, and scope consistency: clean cases are directly accepted, while unsafe or conflicting cases are routed to the Decision Maker.
The Decision Maker performs image-conditioned arbitration using the original image and both evidence records, and outputs authenticity, fake scope, an optional tampered box, and an explanation.
It is trained by SFT on routed hard cases and refined by GRPO with authenticity, scope, arbitration, localization, and format rewards.}

\label{Figure}
\end{figure*}

\begin{table*}[t]
\centering
\caption{\textbf{OOD routed-arbitration example in So-Fake-R1.}
A \texttt{REAL} observer prediction conflicts with a localized
\texttt{TAMPERED} provider prediction, triggering
\texttt{FULL-ARBITRATION}; the Decision Maker resolves the case as
\texttt{TAMPERED} with a support box.}
\label{table5}

\begin{tcolorbox}[exbox]
\footnotesize
\setlength{\parskip}{1.5pt}
\noindent
\begin{minipage}[t]{0.74\linewidth}
\Obs{Visual Observer} ($\mathcal{O}$)
\vspace{1pt}

\Obs{Reasoning:} The image shows a person in a formal setting, possibly an office or a professional environment, with a blue door and a warning sign in the background. The individual is wearing a white shirt and a blue tie, suggesting professional attire. The lighting is even, and there are no visible signs of tampering or manipulation. The overall composition and details are consistent with a real photograph.

\Obs{Authenticity $\hat{a}^{\mathrm{obs}}$:} \textbf{REAL} \qquad
\Obs{Answer $\hat{s}^{\mathrm{obs}}$:} \textbf{REAL} \qquad
\Obs{Bbox $\hat{B}^{\mathrm{obs}}$:} \textbf{NONE}

\vspace{6pt}

\Evi{Forensic Evidence Provider} ($\mathcal{P}$)

\Evi{Descriptors $\mathbf{d}^{\mathrm{for}}$:} mass $14.92$: anomaly evidence is substantial; peak $1.000$: at least one region has a saturated response; concentration $0.984$: the response is concentrated in high-activation patches; support area $0.385$: the candidate region covers a large but not full-frame part of the image; entropy $3.110$: the response remains spatially organized.

\Evi{Basis:} The provider marks a large subject-side candidate region with strong descriptor-confidence support, so it favors a localized manipulation rather than accepting the image as fully authentic.

\Evi{Answer $\hat{s}^{\mathrm{for}}$:} \textbf{TAMPERED} \qquad
\Evi{Confidence $c^{\mathrm{for}}$:} \textbf{0.979} \qquad
\Evi{Bbox $\hat{B}^{\mathrm{for}}$:} \textbf{(0, 22, 167, 224)}

\end{minipage}
\hfill
\begin{minipage}[t]{0.20\linewidth}
\centering
\vspace{0pt}
\vspace{1cm}
\includegraphics[height=3.5cm]{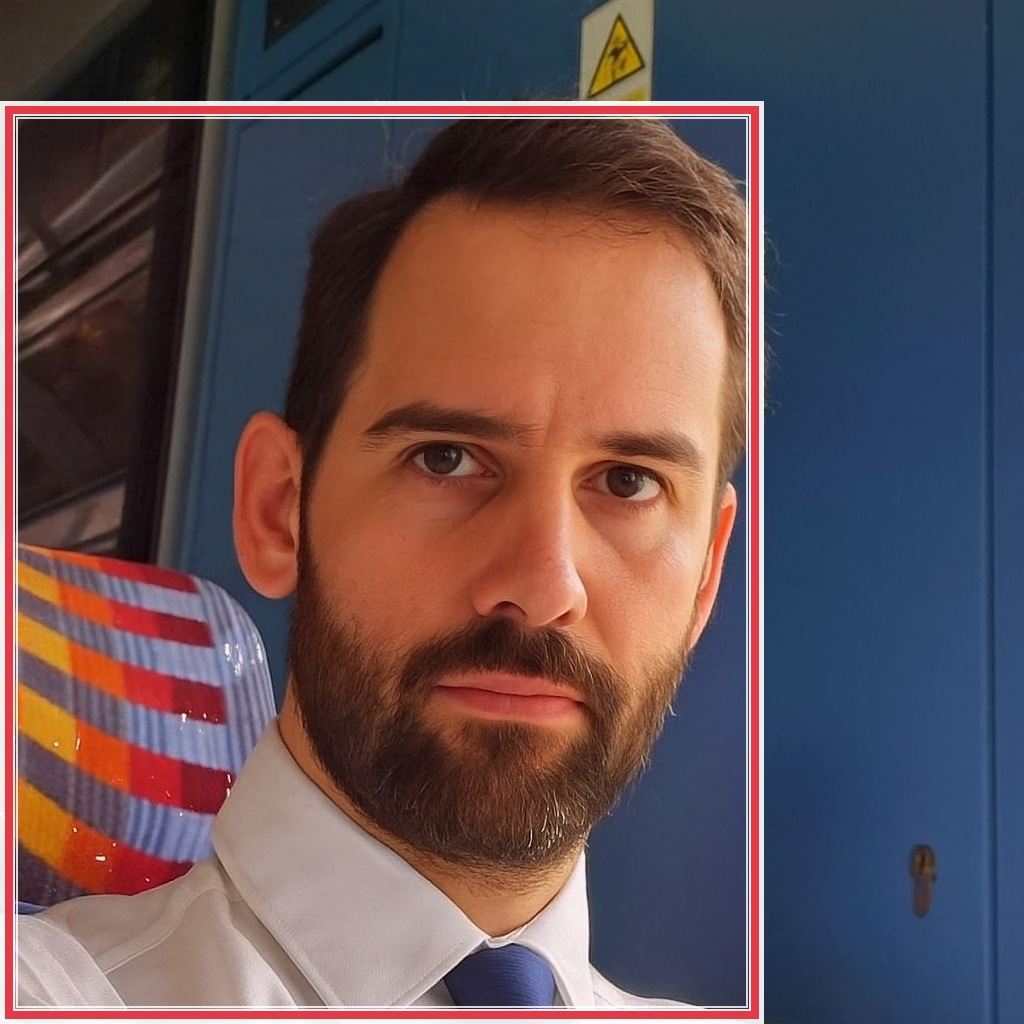}
\end{minipage}
\vspace{6pt}

\Dec{Decision Maker Output} ($\mathcal{D}$)

\Dec{Route:} \textbf{FULL-ARBITRATION}

\Dec{Final explanation $r^{\mathrm{dec}}$:} The observer's \texttt{REAL} claim is plausible from scene layout alone, but it only argues from ordinary appearance and absence of obvious seams. The provider gives a concrete fake-side hypothesis with a large subject-region support box. Rechecking the image shows that the suspicious evidence is concentrated around the person rather than distributed across the entire frame, so the case is better explained as localized tampering than as either a fully authentic photograph or whole-image synthesis.

\Dec{Decision:} The provider's tampering hypothesis survives the image-side check, while the observer's authenticity claim misses the localized fake evidence.

\Dec{Final authenticity $\hat{a}$:} \textbf{FAKE} \qquad
\Dec{Final answer $\hat{s}$:} \textbf{TAMPERED} \qquad
\Dec{Final bbox $\hat{B}$:} \textbf{(0, 22, 167, 224)}

\end{tcolorbox}
\end{table*}

\subsection{Forensic Evidence Provider}
Rather than using the final visual representation, the Forensic Evidence Provider reads an intermediate transformer layer as its feature source.
This choice is motivated by the observation that the most useful visual embeddings for downstream perception are not necessarily located at the output of the network~\cite{DBLP:journals/corr/abs-2504-13181}.
For forensic evidence, this distinction is especially relevant: final-layer VLM representations are increasingly shaped by semantic abstraction and language alignment, while local manipulation traces depend on patch-level texture, boundary, and spatial irregularity cues that may be better preserved before the final layers.
In our implementation, we use the 19th layer:
\begin{equation}
    \mathbf{F}^{(19)}=\Psi_{19}(I)\in\mathbb{R}^{h\times w\times d}.
\end{equation}
The layer is selected by an in-domain validation sweep over
$\{7,11,15,19,23,27,31\}$ using the same FETA readout.
As shown in Fig.~\ref{sup:figure5}, layer 19 gives the best joint behavior among the tested single-layer features: it achieves the highest three-way prediction accuracy, the strongest tampered-support alignment, and a high topology area match.
This suggests that layer 19 preserves sufficient local forensic structure while still providing enough semantic abstraction for stable scope discrimination over $\mathcal{Y}$(Eq.~\ref{eq:space}).
The VLM backbone is kept frozen throughout provider training.

\noindent\textbf{FETA Evidence Heads.} The feature grid is processed by a Forensic Evidence Topology Adapter (FETA), consisting of a projection layer, residual spatial blocks, and $K=2$ parallel evidence heads:
\begin{equation}
\{\mathbf{E}_k\}_{k=1}^{2}
=\Phi_{\mathrm{FETA}}(\mathbf{F}^{(19)}),
\qquad
\mathbf{E}_k\in[0,1]^{h\times w}.
\end{equation}

The heads are not assigned fixed semantics, but provide complementary anomaly-response patterns.

\noindent\textbf{Topology Descriptors.} For each evidence map $\mathbf{E}_k$, we compute five descriptors: mass, peak, concentration, activated area, and entropy.
The mass and peak are:
\begin{equation}
M_k = \sum_{u\in\Omega}\mathbf{E}_k(u),
\qquad
P_k = \max_{u\in\Omega}\mathbf{E}_k(u),
\end{equation}
where $\Omega$ is the patch grid.
Mass measures total anomaly activation, while peak measures the strongest local response.

We normalize the evidence map into a spatial distribution:
\begin{equation}
\widetilde{\mathbf{E}}_k(u)
=
\frac{\mathbf{E}_k(u)}
{\sum_{v\in\Omega}\mathbf{E}_k(v)+\epsilon}.
\end{equation}
The entropy descriptor is:
\begin{equation}
H_k
=
-\sum_{u\in\Omega}
\widetilde{\mathbf{E}}_k(u)
\log\left(\widetilde{\mathbf{E}}_k(u)+\epsilon\right),
\end{equation}
where larger values indicate more spatially dispersed responses.
The concentration descriptor measures how much response is carried by the highest-activation patches:
\begin{equation}
C_k =
\frac{
\sum_{u\in \mathcal{T}_{25}(\mathbf{E}_k)}
\mathbf{E}_k(u)
}{
\sum_{v\in\Omega}\mathbf{E}_k(v)+\epsilon
},
\end{equation}
where $\mathcal{T}_{25}(\mathbf{E}_k)$ denotes the top quartile of patch locations ranked by response.
The activated area is:
\begin{equation}
A_k =
\frac{1}{|\Omega|}
\sum_{u\in\Omega}
\sigma\left(10(\mathbf{E}_k(u)-0.5)\right).
\end{equation}
Let $\mathbf d_k=(M_k,P_k,C_k,A_k,H_k)$ denote the topology descriptor of the $k$-th evidence head, and concatenate the two heads as $\mathbf d^{\mathrm{cls}}=[\mathbf d_1;\mathbf d_2]\in\mathbb{R}^{10}$.

\noindent\textbf{Forensic Classification and Calibration.} A compact topology classifier $g_{\mathrm{cls}}:\mathbb{R}^{10}\rightarrow\mathbb{R}^{3}$ maps $\mathbf d^{\mathrm{cls}}$ to uncalibrated scope logits:
\begin{equation}
\mathbf z^{\mathrm{for}}
=
g_{\mathrm{cls}}\left(\mathbf d^{\mathrm{cls}}\right)
=
\left(
z_{\mathrm{REAL}}^{\mathrm{for}},
z_{\mathrm{FULL}}^{\mathrm{for}},
z_{\mathrm{TAMP}}^{\mathrm{for}}
\right),
\end{equation}
In addition to the topology classifier, the provider uses a residual branch as a one-sided fake-suspiciousness correction.
The residual branch estimates a scalar suspiciousness score $s$ from the provider features.
The score $s$ is not used as an independent classifier; instead, it monotonically shifts the topology-classifier logits away from \textsc{Real} and toward the two fake classes:
\begin{equation}
\begin{aligned}
z_{\mathrm{REAL}}^{\prime\,\mathrm{for}}
&=
z_{\mathrm{REAL}}^{\mathrm{for}}-f(s),\\
z_{\mathrm{FULL}}^{\prime\,\mathrm{for}}
&=
z_{\mathrm{FULL}}^{\mathrm{for}}+\tfrac{1}{2}f(s),\\
z_{\mathrm{TAMP}}^{\prime\,\mathrm{for}}
&=
z_{\mathrm{TAMP}}^{\mathrm{for}}+\tfrac{1}{2}f(s).
\end{aligned}
\end{equation}
In the locked provider used in the main experiments, we use the log-scaled calibration:
\begin{equation}
f(s)
=
s_{\max}
\frac{\log(1+\beta s)}
{\log(1+\beta s_{\max})},
\qquad
s_{\max}=4,\ \beta=2.
\end{equation}
Because the same positive correction is applied to both fake logits, the residual branch strengthens fake suspiciousness without changing the relative preference between \textsc{Full\_Synthetic} and \textsc{Tampered}.
The calibrated logits yield the provider scope prediction and confidence:
\begin{equation}
\begin{gathered}
\mathbf p^{\mathrm{for}}
=
\operatorname{softmax}
\left(\mathbf z^{\prime\,\mathrm{for}}\right),\\
\hat{s}^{\mathrm{for}}
=
\operatorname*{arg\,max}_{y\in\mathcal{Y}}
p_y^{\mathrm{for}},
\qquad
c^{\mathrm{for}}
=
\max_{y\in\mathcal{Y}}
p_y^{\mathrm{for}}.
\end{gathered}
\end{equation}
The corresponding authenticity hypothesis is $\hat{a}^{\mathrm{for}}=\textsc{Real}$ if $\hat{s}^{\mathrm{for}}=\textsc{Real}$, and $\hat{a}^{\mathrm{for}}=\textsc{Fake}$ otherwise.

\noindent\textbf{Topology Map and Support Box.}
For spatial evidence, the evidence heads are aggregated into a topology map:
\begin{equation}
\mathbf E^{\mathrm{topo}}(u)
=
\max_{1\leq k\leq K}
\mathbf E_k(u).
\end{equation}
The topology map is resized to the image resolution and used as an anomaly visualization rather than as an independent segmentation output or final class prediction.
A localization branch predicts a coarse support heatmap from the same frozen feature grid, and the resulting local support is resized and converted to the normalized bounding-box coordinate system used by the Decision Maker.
A support box $\hat{B}^{\mathrm{for}}$ is emitted only when the provider predicts \textsc{Tampered} and a valid local support region is available.

\noindent\textbf{Evidence Provider Output.}
The final provider record is:
\begin{equation}
\mathcal{P}
=
\bigl(
r^{\mathrm{for}},
\hat{a}^{\mathrm{for}},
\hat{s}^{\mathrm{for}},
\hat{B}^{\mathrm{for}},
c^{\mathrm{for}}
\bigr).
\end{equation}
Here, $r^{\mathrm{for}}$ is the serialized forensic report, $\hat{a}^{\mathrm{for}}$ and $\hat{s}^{\mathrm{for}}$ are the authenticity and scope hypotheses, $\hat{B}^{\mathrm{for}}$ is the optional tampered support box, and $c^{\mathrm{for}}$ is the provider confidence.

\subsection{Evidence Gate and Decision Maker}
\label{subsec:decision_maker}

Given the observer record $\mathcal{O}$ and the provider record $\mathcal{P}$, the deterministic evidence gate first checks whether each record is parseable and sufficiently supported.
For the Visual Observer, support is read from the serialized record or inferred using a fixed keyword rule over label-aligned, image-checkable phrases.
For the Forensic Evidence Provider, support combines the classifier logit margin with scope-specific topology-descriptor strength.
A \textsc{Tampered} record is considered sufficiently supported only when it also contains a valid bounding box.

\begin{algorithm}[t]
\caption{Deterministic Evidence Gate
$G(\mathcal{O},\mathcal{P})$}
\label{alg:evidence-gate}
\footnotesize
\begin{algorithmic}[1]
\Require observer record $\mathcal{O}$, provider record $\mathcal{P}$

\If{\text{either record is structurally invalid}}
    \State \Return \textsc{Full-Arbitration}

\ElsIf{$\hat{s}^{\mathrm{obs}}=\hat{s}^{\mathrm{for}}$}
    \If{$\hat{s}^{\mathrm{obs}}=\textsc{Tampered}$ and neither
        record contains a valid box}
        \State \Return \textsc{Full-Arbitration}
    \ElsIf{\text{at least one record is sufficiently supported}}
        \State \Return \textsc{Direct-Accept}
    \Else
        \State \Return \textsc{Full-Arbitration}
    \EndIf

\ElsIf{$
\{\hat{s}^{\mathrm{obs}},\hat{s}^{\mathrm{for}}\}
=
\{\textsc{Real},\textsc{Full\_Synthetic}\}$}
    \State \Return \textsc{Targeted-Arbitration}

\Else
    \State \Return \textsc{Full-Arbitration}
\EndIf
\end{algorithmic}
\end{algorithm}

Let
$\rho=G(\mathcal{O},\mathcal{P})$
denote the assigned route.
For \textsc{Direct-Accept} cases, the Decision Maker is not invoked.
The common scope label is copied to the final output, and the authenticity label is derived using the scope-to-authenticity mapping defined in~\ref{subsec:visual_observer}.
When both evidence records satisfy their support criteria, the case is marked as clean acceptance; when only one record is sufficiently supported, it is marked as cautious acceptance.
For a directly accepted \textsc{Tampered} case, a valid provider box is preferred, with the observer box used as fallback.

Under \textsc{Targeted-Arbitration}, the admissible scope labels are restricted to $\{\textsc{Real},\textsc{Full\_Synthetic}\}$.
Under \textsc{Full-Arbitration}, the Decision Maker may predict any scope in $\mathcal{Y}$.
Only these two arbitration routes invoke the image-conditioned VLM. 
The complete algorithm is shown in Alg.~\ref{alg:evidence-gate}.

On the So-Fake-OOD evaluation split, \textsc{Direct-Accept}, \textsc{Targeted-Arbitration}, and \textsc{Full-Arbitration} account for 75.2\%, 3.6\%, and 21.2\% of the samples, respectively.

For a routed case, the Decision Maker receives the original image, the two evidence records, and the assigned route:
\begin{equation}
(r^{\mathrm{dec}},\hat{a},\hat{s},\hat{B})
=
\Phi_{\mathrm{dec}}
(I,\mathcal{O},\mathcal{P},\rho),
\end{equation}
where $\hat{a}$ is the final \textsc{Real}/\textsc{Fake} decision, $\hat{s}\in\mathcal{Y}$ is the final scope label, $\hat{B}$ is the predicted tampered region when applicable, and $r^{\mathrm{dec}}$ is the final evidence-grounded assessment.
A complete routed-arbitration example is provided in Tab.~\ref{table5}.

\noindent\textbf{Two-stage Training.}
The evidence gate is deterministic, so training focuses on the image-conditioned Decision Maker used for routed hard cases.
Accordingly, SFT and GRPO are constructed only from the two routed categories, targeted arbitration and full arbitration.
We first apply SFT to teach the model the routed decision format: it must read the two evidence records, follow the route, output the final authenticity and scope, provide a tamper box when needed, and explain the arbitration.
However, evidence arbitration is not only a format-following problem.
For the same hard case, a model may produce several plausible explanations while following different evidence sources or rejecting both.
We therefore apply GRPO~\cite{DBLP:journals/corr/abs-2501-12948} to optimize the preference structure of the Decision Maker.

Given a generated response, we parse the final label, authenticity tag, scope tag, route tag, and optional tamper box.
The route-aware reward is:
\begin{equation}
    R =
     R_{\mathrm{fmt}}
    + R_{\mathrm{auth}}
    + R_{\mathrm{scope}}
    + R_{\mathrm{arb}}
    + R_{\mathrm{loc}},
\end{equation}
Here $R_{\mathrm{fmt}}$ checks output validity, $R_{\mathrm{auth}}$ rewards the binary REAL/FAKE decision, $R_{\mathrm{scope}}$ rewards the three-way scope decision, $R_{\mathrm{arb}}$ rewards correct evidence arbitration, and $R_{\mathrm{loc}}$ rewards tamper-box localization when applicable.
Full coefficients are provided in Appendix Sec.~\ref{suppl:reproducibility}.

\begin{table}[t]
\centering
\small
\setlength{\tabcolsep}{4pt}
\caption{\textbf{Parameter and inference efficiency of VLM-based methods.} Trainable parameters exclude frozen backbones; Rel. cost is per-image compute normalized to one 13B forward pass.}
\label{sup:table13}
\begin{tabular}{@{}lcccc@{}}
\toprule
\cellcolor{lightgray!30}Method &\cellcolor{lightgray!30} \cellcolor{lightgray!30}Backbone & \cellcolor{lightgray!30}Trainable &\cellcolor{lightgray!30} Fwd./img & \cellcolor{lightgray!30}Rel. cost \\
\midrule
FakeShield & 13B & $\sim$1.1B & 1.0 & 1.00 \\
SIDA       & 13B & 166M & 1.0 & 1.00 \\
\textbf{So-Fake-R1} & 7B  & \textbf{81M} & 1.25 & \textbf{0.67} \\
\bottomrule
\end{tabular}
\end{table}

\noindent\textbf{Parameter and Inference Efficiency.}
Tab.~\ref{sup:table13} compares trainable parameters and inference cost among VLM-based methods.
So-Fake-R1 trains only 81M parameters (two LoRA adapters of 40.4M each and the 0.26M FETA adapter) on a frozen 7B backbone, roughly half of SIDA~\cite{DBLP:journals/corr/abs-2412-04292} and an order of magnitude fewer than FakeShield~\cite{DBLP:journals/corr/abs-2410-02761}, both of which build on 13B backbones.
This expected cost remains below a single forward pass through the 13B backbones used by several explanation-oriented baselines.

\subsection{Training Data Construction}
\label{subsec:training_data_construction}

We construct separate So-Fake-Set training data for the two evidence sources and for routed Decision Maker arbitration.

\noindent\textbf{Visual Observer Data.}
For the Visual Observer, we sample 100{,}000 image--response pairs from the So-Fake-Set training split.
The samples are class-balanced across the three scope labels, with 33{,}334 REAL, 33{,}333 FULL\_SYNTHETIC, and 33{,}333 TAMPERED images.
Each target contains a semantic visual report and a preliminary scope label; for TAMPERED samples with available annotations, the target also includes the manipulated region.

\noindent\textbf{Forensic Evidence Provider Data.}
The Forensic Evidence Provider uses the same image training pool, preserving the same class-balanced distribution.
All samples provide image-level three-way labels for forensic scope prediction.
For TAMPERED samples with masks or boxes, we additionally use the spatial annotation to supervise localization.

\noindent\textbf{Decision Maker SFT Data.}
After training the two evidence sources, we run both on So-Fake-Set and serialize their outputs into paired Visual Observer and Forensic Evidence Provider records.
We then apply the evidence gate to these pairs.
Cases that can be directly accepted by the gate are excluded from image-conditioned Decision Maker training; only routed hard cases are retained.
Each example contains the original image, the two evidence records, and the assigned route.

\noindent\textbf{Decision Maker GRPO Data.}
For GRPO, we construct a larger routed hard case set of 5{,}000 examples.
Unlike the SFT set, this set is explicitly re-balanced across final labels: 1{,}738 REAL, 1{,}606 FULL\_SYNTHETIC, and 1{,}656 TAMPERED.
It contains 894 targeted conflict cases and 4{,}106 full arbitration cases.
The retained cases cover source conflicts, fake-scope conflicts, weak or unclean agreement, and cases where neither evidence record directly matches the ground truth.
This sampling reduces the risk that the Decision Maker learns a fixed preference for either evidence source and focuses GRPO on hard cases where arbitration is required.
Detailed data construction is provided in Appendix Sec.~\ref{suppl:training-data-optimization}.

\section{Experiments}
\label{sec:Experiments}
\begin{table*}[th]
\centering
\scriptsize
\setlength{\tabcolsep}{1.5pt}
\caption{\textbf{Detection performance on So-Fake-Set and So-Fake-OOD.}
We report per-class accuracy and F1, overall balanced accuracy and macro F1, and the performance drop from So-Fake-Set to So-Fake-OOD.
``*'' indicates released-weight results; FakeShield uses its released checkpoint because its domain-tag-guided design is not directly adaptable to the three-way protocol.}
\begin{adjustbox}{width=1.0\textwidth}
\begin{tabular}{lcc cc c cc c cc c cc c cc c cc c cc c cc c cc}
\toprule
\cellcolor{lightgray!30} &
\cellcolor{lightgray!30} &
\cellcolor{lightgray!30} &
\multicolumn{11}{c}{\cellcolor{lightgray!30} So-Fake-Set} &&
\multicolumn{11}{c}{\cellcolor{lightgray!30} So-Fake-OOD} &&
\multicolumn{2}{c}{\cellcolor{lightgray!30} Drop $\downarrow$} \\
\cline{4-14} \cline{16-26} \cline{28-29}

\cellcolor{lightgray!30} Method &
\cellcolor{lightgray!30} Year &
\cellcolor{lightgray!30} Type &
\multicolumn{2}{c}{\cellcolor{lightgray!30} REAL} & \cellcolor{lightgray!30} &
\multicolumn{2}{c}{\cellcolor{lightgray!30} FULL} & \cellcolor{lightgray!30} &
\multicolumn{2}{c}{\cellcolor{lightgray!30} TAMP.} & \cellcolor{lightgray!30} &
\multicolumn{2}{c}{\cellcolor{lightgray!30} Overall} &&
\multicolumn{2}{c}{\cellcolor{lightgray!30} REAL} & \cellcolor{lightgray!30} &
\multicolumn{2}{c}{\cellcolor{lightgray!30} FULL} & \cellcolor{lightgray!30} &
\multicolumn{2}{c}{\cellcolor{lightgray!30} TAMP.} & \cellcolor{lightgray!30} &
\multicolumn{2}{c}{\cellcolor{lightgray!30} Overall} &&
\multicolumn{2}{c}{\cellcolor{lightgray!30} Overall} \\
\cline{4-5} \cline{7-8} \cline{10-11} \cline{13-14}
\cline{16-17} \cline{19-20} \cline{22-23} \cline{25-26}
\cline{28-29}

\cellcolor{lightgray!30} &
\cellcolor{lightgray!30} &
\cellcolor{lightgray!30} &
\cellcolor{lightgray!30} Acc. &
\cellcolor{lightgray!30} F1 & \cellcolor{lightgray!30} &
\cellcolor{lightgray!30} Acc. &
\cellcolor{lightgray!30} F1 & \cellcolor{lightgray!30} &
\cellcolor{lightgray!30} Acc. &
\cellcolor{lightgray!30} F1 & \cellcolor{lightgray!30} &
\cellcolor{lightgray!30} Acc. &
\cellcolor{lightgray!30} F1 &&
\cellcolor{lightgray!30} Acc. &
\cellcolor{lightgray!30} F1 & \cellcolor{lightgray!30} &
\cellcolor{lightgray!30} Acc. &
\cellcolor{lightgray!30} F1 & \cellcolor{lightgray!30} &
\cellcolor{lightgray!30} Acc. &
\cellcolor{lightgray!30} F1 & \cellcolor{lightgray!30} &
\cellcolor{lightgray!30} Acc. &
\cellcolor{lightgray!30} F1 &&
\cellcolor{lightgray!30} Acc. &
\cellcolor{lightgray!30} F1 \\

\midrule

CnnSpot~\cite{wang2019cnngenerated} & 2021 & Detection & 93.1 & 89.3 && 84.4 & 90.1 && 90.7 & 85.6  &  & 89.4 & 88.3 && 80.0 & 64.7  && 67.8 & 60.9  && 14.7 & 23.1 & & 54.2 & 49.5 && 35.2 & 38.8 \\
FatFormer~\cite{DBLP:conf/cvpr/LiuTTW0Z24} & 2024 & Detection  & 94.1 & 93.8 && 85.5 & 91.7 && 93.8 & 86.9 & & 91.1 & 90.8 && 83.6 & 83.5 && 78.8 & 66.1 && 17.1 & 22.2 & & 59.8 & 57.2 && 31.3 & 33.6 \\
UnivFD~\cite{DBLP:conf/cvpr/OjhaLL23} & 2023 & Detection & 81.5 & 81.3 && 80.0 & 85.2 && 85.8 & 78.1  &  & 82.4 & 81.5 && 37.2 & 45.4  && \textbf{82.6} & 63.6  && 26.1 & 31.6 & & 48.6 & 46.9 && 33.8 & 34.7 \\
FreAware~\cite{DBLP:conf/aaai/Tan0WGLW24} & 2024 & Detection & 83.9 & 78.7 && 81.6 & 86.7 && 83.4 & 81.1  &  & 83.0 & 82.2 && 87.9 & 73.5  && 49.6 & 44.3  && 10.9 & 16.4 & & 49.5 & 44.7 && 33.5 & 37.5 \\
EFFORT~\cite{DBLP:conf/icml/0002WJZLCYDW025} & 2025 & Detection  & \textbf{99.2} & \textbf{98.8} && \textbf{87.9} & \textbf{93.5} && \textbf{99.2} & 90.8  &  & \textbf{95.5} & \textbf{94.3} && 94.7 & 81.8  && 73.6 & 66.1  && 28.3 & 39.6 & & 65.5 & 62.5 && 29.9 & 31.8 \\
DDA~\cite{DBLP:journals/corr/abs-2505-14359} & 2025 & Detection & 96.6 & 93.4 && 85.2 & 91.7 && 95.0 & 88.0  &  & 92.3 & 91.0 && 87.2 & 67.1  && 65.3 & 63.5  && 25.2 & 37.3 & & 59.2 & 56.0 && 33.0 & 35.0 \\
C2P-CLIP~\cite{DBLP:conf/aaai/TanTLGWZW25} & 2025 & Detection & 77.3 & 75.1 && 78.6 & 82.7 && 74.0 & 70.4  &  & 76.7 & 76.1 && 36.1 & 46.1  && 79.0 & 60.7  && 22.3 & 26.1 & & 45.8 & 44.3 && 30.8 & 31.8 \\
\midrule
TruFor~\cite{DBLP:conf/cvpr/GuillaroCSDV23} & 2023 & IFDL & 98.9 & 72.3 && 63.5 & 59.4 && 82.1 & 57.0  &  & 81.5 & 62.9 && 96.0 & 47.9  && 10.6 & 17.8  && 6.5 & 10.6 & & 37.7 & 25.4 && 43.8 & 37.5 \\
PSCC-Net~\cite{DBLP:journals/tcsv/LiuLCL22} & 2022 & IFDL& 97.5 & 80.4 && 84.7 & 72.5 && 78.2 & 48.8  &  & 86.8 & 67.2 && \textbf{98.2} & 48.9  && 15.2 & 25.6  && 2.1 & 3.5 & & 38.5 & 26.0 && 48.3 & 41.2 \\
FakeShield*~\cite{DBLP:journals/corr/abs-2410-02761} & 2025 & IFDL & 78.0 & 72.0 && 61.5 & 61.0 && 61.5 & 59.3  &  & 67.0 & 64.1 && 56.0 & 48.0 && 42.0 & 42.0 && 25.3 & 32.1 & & 41.1 & 40.7 && 25.9 & 23.4 \\
SIDA~\cite{DBLP:journals/corr/abs-2412-04292} & 2025 & IFDL & 94.5 & 92.0 && 83.0 & 86.5 && 87.1 & 83.4  &  & 88.2 & 87.3 && 83.0 & 77.5 && 68.0 & 66.0 && 45.8 & 49.1 & & 65.6 & 64.2 && 22.6 & 23.1 \\
\midrule
LLaVA-1.5-13B~\cite{DBLP:conf/nips/LiuLWL23a} & 2023 & VLM & 91.0 & 88.5 && 77.0 & 82.0 && 82.5 & 78.2  &  & 83.5 & 82.9 && 80.0 & 75.0 && 68.0 & 66.0 && 40.5 & 43.9 & & 62.8 & 61.6 && 20.7 & 21.3 \\
LISA~\cite{DBLP:conf/cvpr/LaiTCLY0J24} & 2024 & VLM & 94.0 & 91.0 && 82.2 & 85.0 && 86.0 & 81.7  &  & 87.4 & 85.9 && 82.0 & 76.0 && 68.0 & 66.5 && 43.1 & 47.9 & & 64.4 & 63.5 && 23.0 & 22.4 \\
DeepSeek-VL-7B~\cite{DBLP:journals/corr/abs-2403-05525} & 2025 & VLM & 94.0 & 89.0 && 76.0 & 80.0 && 81.1 & 74.3  &  & 83.7 & 81.1 && 80.5 & 74.0 && 66.5 & 64.0 && 41.0 & 43.0 & & 62.7 & 60.3 && 21.0 & 20.8 \\
DeepSeek-VL2-tiny~\cite{DBLP:journals/corr/abs-2412-10302} & 2025 & VLM & 86.0 & 82.0 && 68.5 & 74.0 && 72.0 & 65.4  &  & 75.5 & 73.8 && 71.0 & 67.0 && 57.0 & 55.0 && 37.0 & 38.9 & & 55.0 & 53.6 && 20.5 & 20.2 \\
Qwen2.5-Omni-3B~\cite{DBLP:journals/corr/abs-2503-20215} & 2025 & VLM & 95.0 & 92.5 && 83.0 & 87.0 && 87.2 & 82.7  &  & 88.4 & 87.4 && 83.0 & 78.0 && 69.0 & 67.0 && 43.4 & 47.2 & & 65.1 & 64.1 && 23.3 & 23.3 \\
Qwen2.5-VL-7B~\cite{DBLP:journals/corr/abs-2502-13923} & 2025 & VLM  & 95.3 & 92.0 && 73.5 & 82.1 && 96.1 & 89.9  &  & 88.3 & 88.0 && 86.5 & 82.0  && 71.0 & 65.0  && 38.5 & 41.0 & & 65.3 & 62.7 && 23.0 & 25.3 \\

\cellcolor{softblue}\textbf{So-Fake-R1} \cellcolor{softblue}& -- \cellcolor{softblue}& VLM \cellcolor{softblue}& 98.0 \cellcolor{softblue}& 94.8 \cellcolor{softblue}&\cellcolor{softblue}& 76.5 \cellcolor{softblue}& \cellcolor{softblue}85.8 &\cellcolor{softblue}& 99.0 \cellcolor{softblue}& \textbf{92.2}  \cellcolor{softblue}&  \cellcolor{softblue}& 91.2 \cellcolor{softblue}& 90.9 \cellcolor{softblue}&& 94.0 \cellcolor{softblue}& \textbf{86.2}  \cellcolor{softblue}&\cellcolor{softblue}& 71.2 \cellcolor{softblue}& \textbf{68.3}  \cellcolor{softblue}&\cellcolor{softblue}& \textbf{50.8} \cellcolor{softblue}& \textbf{58.2} \cellcolor{softblue}& \cellcolor{softblue}& \textbf{72.0} \cellcolor{softblue}& \textbf{70.9} \cellcolor{softblue}& & \textbf{19.2} \cellcolor{softblue}& \textbf{20.0} \cellcolor{softblue}\\

\bottomrule
\end{tabular}
\end{adjustbox}
\label{table6}
\end{table*}
\subsection{Experimental Settings}
\label{subsec:experimentalsettings}

\textbf{Methods.}
To compare So-Fake-R1 across detection, localization, and explanation tasks, we evaluate it against three groups of baselines under a unified protocol:
\textbf{(1) Detection-only methods}, including CnnSpot~\cite{wang2019cnngenerated}, UnivFD~\cite{DBLP:conf/cvpr/OjhaLL23}, FreAware ~\cite{DBLP:conf/aaai/Tan0WGLW24},
EFFORT~\cite{DBLP:conf/icml/0002WJZLCYDW025},
DDA~\cite{DBLP:journals/corr/abs-2505-14359}, FatFormer~\cite{DBLP:conf/cvpr/LiuTTW0Z24} and 
C2P-CLIP~\cite{DBLP:conf/aaai/TanTLGWZW25}.
\textbf{(2) Image Forgery Detection and Localization (IFDL) methods}, including  TruFor~\cite{DBLP:conf/cvpr/GuillaroCSDV23}, PSCC-Net~\cite{DBLP:journals/tcsv/LiuLCL22}, SIDA~\cite{DBLP:journals/corr/abs-2412-04292}, and FakeShield~\cite{DBLP:journals/corr/abs-2410-02761}.
\textbf{(3) Explanation-oriented methods,} including LISA~\cite{DBLP:conf/cvpr/LaiTCLY0J24}, InternVL3-8B~\cite{zhu2025internvl3exploringadvancedtraining}, LLaVA-1.5-13B~\cite{DBLP:conf/nips/LiuLWL23a}, Qwen2.5-VL-7B~\cite{DBLP:journals/corr/abs-2502-13923}, Qwen2.5-Omni-3B~\cite{DBLP:journals/corr/abs-2503-20215}, DeepSeek-VL2-tiny~\cite{DBLP:journals/corr/abs-2412-10302}, and DeepSeek-VL-7B~\cite{DBLP:journals/corr/abs-2403-05525}. Unless otherwise noted, all baselines are fine-tuned on the So-Fake-Set training split and tested on both the So-Fake-Set validation split and So-Fake-OOD. 
Non-VLM baselines are trained on the full training split, while VLM-based baselines use the same 100K class-balanced subset as our Visual Observer.
Per-baseline three-way adaptations and training configurations are detailed in Sec.~\ref{suppl:baseline-config} of the Appendix.

\noindent\textbf{Metrics.} 
We evaluate models across the three tasks defined by So-Fake. 
For \textbf{detection}, we report image-level accuracy (Acc.) and F1.  
For \textbf{localization}, we report mask-level Intersection over Union (IoU) and mask-level F1.
For methods that output bounding boxes (e.g., VLM-based baselines and our method), we use a fixed SAM2~\cite{DBLP:journals/corr/abs-2408-00714} configuration to convert predicted boxes into segmentation masks before evaluation; for methods that directly output masks, we evaluate their predicted masks.
For \textbf{explanation}, we employ two complementary metrics: 
(1) Cosine Semantic Similarity (CSS) and 
(2) ROUGE-L, which quantifies textual overlap through longest common subsequence matching. Reference rationales were first generated with Claude Opus 4.8~\cite{claude} and then carefully revised and validated by human experts. 

\subsection{Performance and Analysis of So-Fake-R1}
\label{subsec:sofake_r1_performance}

\subsubsection{Comparison Results on So-Fake}
\label{subsubsec:sofake_results}

\noindent\textbf{Detection Results.}
Tab.~\ref{table6} reports detection performance on both So-Fake-Set and So-Fake-OOD under the three-way authenticity protocol.
On So-Fake-Set, EFFORT~\cite{DBLP:conf/icml/0002WJZLCYDW025} and DDA~\cite{DBLP:journals/corr/abs-2505-14359} achieve strong source-domain performance, while So-Fake-R1 remains competitive overall and obtains the highest TAMPERED F1.
However, the difference between methods becomes clearer under So-Fake-OOD, where models must transfer to native social-media imagery and held-out commercial generators.
Most baselines suffer substantial performance drops in this setting, especially on the TAMPERED class, suggesting that their predictions rely heavily on source-specific or generator-specific artifacts.
In contrast, So-Fake-R1 achieves the best OOD overall performance, with \textbf{72.0} balanced accuracy and \textbf{70.9} F1, and also obtains the strongest TAMPERED detection results.
These results suggest that evidence-grounded reasoning is particularly useful under distribution shift, where localized manipulations must be distinguished from both authentic content and full synthetic images.

\begin{table}[t]
\centering
\scriptsize
\setlength{\tabcolsep}{1.5pt}
\caption{\textbf{Tamper localization performance on So-Fake.}}
\begin{adjustbox}{width=1.0\columnwidth}
\begin{tabular}{l cc c cc c cc}
\toprule
\cellcolor{lightgray!30} &
\multicolumn{2}{c}{\cellcolor{lightgray!30} So-Fake-Set} &&
\multicolumn{2}{c}{\cellcolor{lightgray!30} So-Fake-OOD} &&
\multicolumn{2}{c}{\cellcolor{lightgray!30} Drop $\downarrow$} \\
\cline{2-3} \cline{5-6} \cline{8-9}
\multirow{-2}{*}{\cellcolor{lightgray!30} \makecell[c]{Method}} &
\cellcolor{lightgray!30} IoU &
\cellcolor{lightgray!30} F1 &&
\cellcolor{lightgray!30} IoU &
\cellcolor{lightgray!30} F1 &&
\cellcolor{lightgray!30} IoU &
\cellcolor{lightgray!30} F1 \\
\midrule
TruFor~\cite{DBLP:conf/cvpr/GuillaroCSDV23} &
47.5 & 55.6 &&
26.4 & 37.1 &&
21.1 & 18.5 \\
PSCC-Net~\cite{DBLP:journals/tcsv/LiuLCL22} &
29.4 & 35.1 &&
26.3 & 34.9 &&
3.1 & 0.2 \\
FakeShield*~\cite{DBLP:journals/corr/abs-2410-02761} &
33.7 & 46.1 &&
24.9 & 30.2 &&
8.8 & 15.9 \\
SIDA~\cite{DBLP:journals/corr/abs-2412-04292} &
44.1 & 51.4 &&
40.1 & 49.3 &&
4.0 & 9.6 \\
\midrule
LLaVA-1.5-13B~\cite{DBLP:conf/nips/LiuLWL23a} &
29.8 & 38.1 &&
25.6 & 35.1 &&
4.2 & 3.0 \\
LISA~\cite{DBLP:conf/cvpr/LaiTCLY0J24} &
40.5 & 47.6 &&
38.2 & 47.5 &&
2.3 & \textbf{0.1} \\
DeepSeek-VL-7B~\cite{DBLP:journals/corr/abs-2403-05525} &
27.8 & 35.4 &&
25.4 & 34.6 &&
2.4 & 0.8 \\
DeepSeek-VL2-tiny~\cite{DBLP:journals/corr/abs-2412-10302} &
37.5 & 43.0 &&
27.5 & 33.9 &&
10.0 & 9.1 \\
Qwen2.5-Omni-3B~\cite{DBLP:journals/corr/abs-2503-20215} &
43.1 & 52.2 &&
41.5 & 47.0 &&
1.6 & 8.2 \\
Qwen2.5-VL-7B~\cite{DBLP:journals/corr/abs-2502-13923} &
42.7 & 50.1 &&
41.2 & 47.9 &&
1.5 & 2.2 \\

\cellcolor{softblue}\textbf{So-Fake-R1} &
\cellcolor{softblue}\textbf{48.6} &\cellcolor{softblue} \textbf{56.4} &&\cellcolor{softblue}
\textbf{47.8} & \cellcolor{softblue}\textbf{52.6} &&\cellcolor{softblue}
\textbf{0.8} & \cellcolor{softblue}3.8 \\
\bottomrule
\end{tabular}
\end{adjustbox}
\label{table7}
\end{table}

\noindent\textbf{Localization Results.}
Tab.~\ref{table7} reports tamper localization performance on TAMPERED samples.
On So-Fake-Set, So-Fake-R1 achieves the best localization performance, with \textbf{48.6} IoU and \textbf{56.4} localization F1.
This indicates that the proposed evidence-grounded decision process not only identifies manipulated images, but also produces more accurate spatial grounding for the manipulated region.
The advantage is more evident under the OOD setting.
While most baselines degrade when evaluated on unseen social-media platforms and generators, So-Fake-R1 remains the top method with \textbf{47.8} IoU and \textbf{52.6} localization F1.
It outperforms the strongest OOD baseline by \textbf{6.3} points in IoU and \textbf{3.3} points in localization F1, while
showing only a small IoU drop of \textbf{0.8} from So-Fake-Set to So-Fake-OOD.

\begin{figure*}[ht]
\begin{center}
\centerline{\includegraphics[width=1\linewidth]{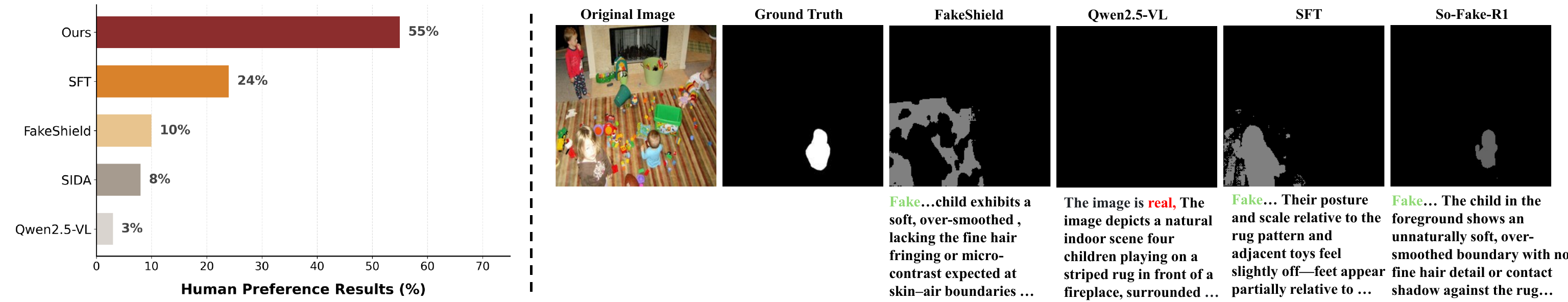}}
\caption{\textbf{Human evaluation of explanation and localization quality.}
The left panel reports human preference scores across competing methods, while the right panel shows representative examples comparing predicted labels, localization results, and explanations. }
\label{Figure7}
\end{center}
\end{figure*}

\noindent\textbf{Explanation Results.}
Tab.~\ref{table8} reports explanation quality on So-Fake-Set and So-Fake-OOD for explanation-capable methods.
So-Fake-R1 achieves the best performance on both splits, with \textbf{0.47} ROUGE-L and \textbf{0.88} CSS on So-Fake-Set, and \textbf{0.47} ROUGE-L and \textbf{0.84} CSS on So-Fake-OOD.
Compared with strong MLLM baselines, So-Fake-R1 produces explanations that are more semantically aligned with the reference rationales.
Moreover, So-Fake-R1 maintains strong explanation performance under the OOD setting, indicating that the evidence-grounded decision process helps produce reliable rationales.

\noindent\textbf{Human Evaluation.}
To further assess whether the generated explanations are faithful to image evidence, we conduct a human preference study with five
independent evaluators over 500 So-Fake-OOD samples spanning the three authenticity scopes.
Each sample is assessed by one evaluator, yielding 500 selections in total.
For each sample, the evaluator is shown the original image and the
anonymized, randomly ordered outputs of the five competing methods,
including the predicted label, explanation, and localization result
when available.
They are asked to select the single output that best aligns with their judgment, considering whether the explanation reflects the actual manipulation cues and whether the localization is semantically coherent.
As shown in Fig.~\ref{Figure7} (left), So-Fake-R1 is selected as the best output in $55\%$ of all selections, substantially outperforming SFT ($24\%$), FakeShield ($10\%$), SIDA ($8\%$), and Qwen2.5-VL ($3\%$).
The qualitative example in Fig.~\ref{Figure7} (right) further illustrates this advantage: competing methods either over-localize irrelevant regions, miss the tampered area, or produce less grounded explanations.
In contrast, So-Fake-R1 localizes the manipulated region more accurately and provides an explanation consistent with the visual evidence.

\begin{table}[t]
\centering
\scriptsize
\setlength{\tabcolsep}{1.5pt}
\caption{\textbf{Explanation performance on So-Fake-Set and So-Fake-OOD using ROUGE-L and CSS.}}
\begin{adjustbox}{width=1.0\columnwidth}
\begin{tabular}{l cc c cc}
\toprule
\cellcolor{lightgray!30} &
\multicolumn{2}{c}{\cellcolor{lightgray!30} So-Fake-Set} &&
\multicolumn{2}{c}{\cellcolor{lightgray!30} So-Fake-OOD} \\
\cline{2-3} \cline{5-6}
\multirow{-2}{*}{\cellcolor{lightgray!30} \makecell[c]{Method}} &
\cellcolor{lightgray!30} ROUGE-L &
\cellcolor{lightgray!30} CSS &&
\cellcolor{lightgray!30} ROUGE-L &
\cellcolor{lightgray!30} CSS \\
\midrule

FakeShield*~\cite{DBLP:journals/corr/abs-2410-02761} &
0.24 & 0.51 &&
0.24 & 0.51 \\
SIDA~\cite{DBLP:journals/corr/abs-2412-04292} &
0.43 & 0.80 &&
0.41 & 0.78 \\
\midrule
LLaVA-1.5-13B~\cite{DBLP:conf/nips/LiuLWL23a} &
0.42 & 0.79 &&
0.42 & 0.76 \\
LISA~\cite{DBLP:conf/cvpr/LaiTCLY0J24} &
0.42 & 0.79 &&
0.41 & 0.78 \\
DeepSeek-VL-7B~\cite{DBLP:journals/corr/abs-2403-05525} &
0.44 & 0.82 &&
0.42 & 0.77 \\
DeepSeek-VL2-tiny~\cite{DBLP:journals/corr/abs-2412-10302} &
0.44 & 0.83 &&
0.41 & 0.76 \\
Qwen2.5-Omni-3B~\cite{DBLP:journals/corr/abs-2503-20215} &
0.45 & 0.83 &&
0.43 & 0.80 \\
Qwen2.5-VL-7B~\cite{DBLP:journals/corr/abs-2502-13923} &
0.45 & \underline{0.84} &&
0.43 & 0.81 \\

\cellcolor{softblue}
\textbf{So-Fake-R1} &\cellcolor{softblue}
\textbf{0.47} &\cellcolor{softblue}
\textbf{0.88} &&\cellcolor{softblue}
\textbf{0.47} &\cellcolor{softblue}
\textbf{0.84} \\

\bottomrule
\end{tabular}
\end{adjustbox}
\label{table8}
\end{table}

\begin{table*}[t]
\centering
\scriptsize
\setlength{\tabcolsep}{1.5pt}
\caption{\textbf{External evaluation.}
Released models are evaluated on independent social-media and in-the-wild benchmarks.
Because these datasets provide only binary real/fake labels, we report authenticity accuracy.}
\begin{adjustbox}{width=1.0\textwidth}
\begin{tabular}{l c c cc c ccc c cc c cc c c c c c}
\toprule
\cellcolor{lightgray!30} &
\cellcolor{lightgray!30} &&
\multicolumn{2}{c}{\cellcolor{lightgray!30} AIGI-Now} &&
\multicolumn{3}{c}{\cellcolor{lightgray!30} WildRF} &&
\multicolumn{2}{c}{\cellcolor{lightgray!30} AIGIBench} &&
\multicolumn{2}{c}{\cellcolor{lightgray!30} CO-SPY-Bench/in-the-wild} &&
\cellcolor{lightgray!30} &&
\cellcolor{lightgray!30} &
\cellcolor{lightgray!30} \\
\cline{4-5} \cline{7-9} \cline{11-12} \cline{14-15}
\multirow{-2}{*}{\cellcolor{lightgray!30} \makecell[c]{Method}} &
\multirow{-2}{*}{\cellcolor{lightgray!30} \makecell[c]{Chameleon}} &&
\cellcolor{lightgray!30} Pixel &
\cellcolor{lightgray!30} Semantic &&
\cellcolor{lightgray!30} FB &
\cellcolor{lightgray!30} Reddit &
\cellcolor{lightgray!30} Twitter &&
\cellcolor{lightgray!30} SocRF &
\cellcolor{lightgray!30} ComAI &&
\cellcolor{lightgray!30} DALLE3 &
\cellcolor{lightgray!30} Lexica &&
\multirow{-2}{*}{\cellcolor{lightgray!30} \makecell[c]{RR-\\Dataset}} &&
\multirow{-2}{*}{\cellcolor{lightgray!30} \makecell[c]{BFree-\\Online}} &
\multirow{-2}{*}{\cellcolor{lightgray!30} \makecell[c]{avg\\Acc}} \\
\midrule
NPR~\cite{DBLP:conf/cvpr/TanLZWGLW24} & 55.2 && 69.3 & 50.0 && 53.8 & 53.7 & 57.7 && 55.8 & 52.0 && 74.4 & 3.9 && 48.3 && 40.5 & 51.2 \\
UnivFD~\cite{DBLP:conf/cvpr/OjhaLL23} & 57.2 && 52.5 & 52.4 && 52.7 & 55.1 & 66.5 && 51.6 & 45.4 && 76.6 & 63.1 && 51.1 && 57.2 & 56.8 \\
FatFormer~\cite{DBLP:conf/cvpr/LiuTTW0Z24} & 57.8 && 52.6 & 49.9 && 52.5 & 65.4 & 39.7 && 55.7 & 50.4 && 2.2 & 0.3 && 50.4 && 32.7 & 42.5 \\
SAFE~\cite{DBLP:conf/kdd/LiCHJHF25} & 56.8 && 77.5 & 56.7 && 49.7 & 49.2 & 34.7 && 49.2 & 49.3 && 0.9 & 0.3 && 49.3 && 32.6 & 42.2 \\
C2P-CLIP~\cite{DBLP:conf/aaai/TanTLGWZW25} & 57.6 && 51.7 & 50.0 && 51.9 & 67.6 & 40.4 && 58.1 & 50.4 && 4.5 & 0.2 && 50.0 && 32.7 & 42.9 \\
B-Free~\cite{DBLP:conf/cvpr/GuillaroZUSCV25} & 75.1 && 70.0 & 54.3 &&94.7 & 85.5 & 97.1 && 84.3 & 78.9 &&96.3 & 50.8 && 69.5 && 90.3 &  78.9 \\
AIDE~\cite{DBLP:conf/iclr/YanLCHJ0X25} & 58.4 && 77.1 & 57.4 && 61.6 & 66.3 & 52.5 && 59.2 & 62.2 && 34.0 & 19.5 && 57.6 && 52.1 & 54.8 \\
DRCT~\cite{DBLP:conf/icml/ChenZYY24} & 56.6 && 80.9 & 57.9 && 90.3 & 66.8 & 79.6 && 71.3 & 84.6 && 82.1 & 55.8 && 58.2 && 77.1 & 71.8 \\

Aligned~\cite{DBLP:conf/iclr/RajanOSL25} & 65.3 && 68.4 & 50.2 && 48.4 & 54.0 & 40.6 && 51.0 & 60.2 && 8.4 & 11.5 && 47.7 && 38.1 & 45.3 \\
\cellcolor{softblue}\textbf{So-Fake-R1} & 
\cellcolor{softblue}73.4 &&
\cellcolor{softblue}89.4 &
\cellcolor{softblue}96.4 &&
\cellcolor{softblue}74.7 &
\cellcolor{softblue}89.3 &
\cellcolor{softblue}85.0 &&
\cellcolor{softblue}79.8 &
\cellcolor{softblue}68.9 &&
\cellcolor{softblue}91.2 &
\cellcolor{softblue}72.4 &&
\cellcolor{softblue}74.8 &&
\cellcolor{softblue}90.7 &
\cellcolor{softblue}\textbf{82.2} \\
\bottomrule
\end{tabular}
\end{adjustbox}
\label{table9}
\end{table*}

\subsubsection{External Evaluation}
\label{subsubsec}

To further examine the method-level generalizability of So-Fake-R1, we evaluate it on a collection of independent social-media and in-the-wild forgery-detection benchmarks, as shown in Tab.~\ref{table9}.
These datasets are collected independently from So-Fake and cover diverse online sources and social platforms.

Overall, So-Fake-R1 achieves the highest average accuracy across the external benchmarks, reaching \textbf{82.2}.
A notable observation is that many released detectors perform strongly on certain subsets but fail to transfer consistently across data sources.
For example, NPR reaches $74.4$ accuracy on the DALLE3 subset of CO-SPY-Bench, but drops to $3.9$ on Lexica.
The large cross-dataset variation highlights sensitivity to dataset composition, label convention, thresholding, and source shift.
In contrast, So-Fake-R1 maintains more balanced performance across the evaluated sources, suggesting that its evidence-grounded decision process transfers more reliably to independent social-media scenarios.

\begin{table}[t]
\centering
\scriptsize
\setlength{\tabcolsep}{1.5pt}
\caption{\textbf{Robustness to social-media propagation chains.}
Detection accuracy on clean and transformed So-Fake-OOD;
Avg. Drop denotes the decrease from Clean to the transformed average.}
\begin{adjustbox}{width=1.0\columnwidth}
\begin{tabular}{l c c c c c c}
\toprule
\cellcolor{lightgray!30} Method &
\cellcolor{lightgray!30} Clean &
\cellcolor{lightgray!30} Re-enc. &
\cellcolor{lightgray!30} Web Dist. &
\cellcolor{lightgray!30} Repost &
\cellcolor{lightgray!30} Cross-Plat. &
\cellcolor{lightgray!30} Avg. Drop $\downarrow$ \\
\midrule

CnnSpot~\cite{wang2019cnngenerated} & 54.2 & 49.6 & 47.8 & 45.4 & 43.2 & 7.7 \\
UnivFD~\cite{DBLP:conf/cvpr/OjhaLL23} & 48.6 & 45.5 & 43.8 & 42.0 & 39.8 & 5.8 \\
FreAware~\cite{DBLP:conf/aaai/Tan0WGLW24} & 49.5 & 45.6 & 43.7 & 41.5 & 39.8 & 6.9 \\
EFFORT~\cite{DBLP:conf/icml/0002WJZLCYDW025} & 65.5 & 60.1 & 57.6 & 54.8 & 51.9 & 9.4 \\
DDA~\cite{DBLP:journals/corr/abs-2505-14359}  & 59.2 & 54.0 & 51.5 & 49.0 & 46.7 & 8.9 \\
C2P-CLIP~\cite{DBLP:conf/aaai/TanTLGWZW25} & 45.8 & 42.1 & 40.3 & 38.5 & 36.7 & 6.4 \\

\rowcolor{softblue}
\textbf{So-Fake-R1} & \textbf{72.0} & \textbf{69.5} & \textbf{68.2} & \textbf{67.0} & \textbf{65.9} & \textbf{4.4} \\

\bottomrule
\end{tabular}
\end{adjustbox}
\label{table10}
\end{table}

\subsubsection{Robustness to Social-Media Propagation}
\label{subsubsec:propagation_robustness}

Although So-Fake-OOD evaluates detector generalization to native social-media imagery and unseen generative models, real online images are often redistributed after their initial upload through downloading, reposting, screenshotting, resizing, and platform-side re-encoding.
Such secondary propagation can weaken forensic artifacts and alter image statistics, posing additional challenges for practical deployment.
To approximate secondary propagation, we apply four controlled propagation chains to So-Fake-OOD:
\begin{itemize}
\item \textbf{Re-enc.}: resize the long side to $1024$ pixels; JPEG compression at quality $75$.
\item \textbf{Web Dist.}: resize the long side to $1024$ pixels; convert to WebP at quality $80$; re-encode as JPEG at quality $85$.
\item \textbf{Repost.}: place the image on a $1080{\times}1920$ canvas with mild padding; JPEG compression at quality $80$; resize back to
the evaluation resolution.
\item \textbf{Cross-Plat.}: resize the long side to $1280$ pixels; JPEG compression at quality $85$; convert to WebP at quality $80$; resize the long side to $1024$ pixels; re-encode at JPEG $75$.
\end{itemize}

We use the same fine-tuned detection models as in the So-Fake-OOD comparison and report image-level detection accuracy in Tab.~\ref{table10}.
The results show that social-media propagation substantially affects many detection-only baselines, especially those relying on fragile compression traces or generator-specific artifacts.
These methods often suffer clear performance drops after re-encoding or reposting, indicating limited stability under realistic redistribution.
In contrast, So-Fake-R1 maintains the most stable performance across the propagation chains and achieves the smallest average degradation relative to the clean So-Fake-OOD setting. 
This suggests that its evidence-grounded decision process is less dependent on a single low-level artifact pattern and remains more reliable when social-media images undergo additional platform processing or cross-platform reposting.

\subsubsection{Ablation Studies}
\label{subsubsec:ablation_studies}

\begin{table}[t]

\centering

\scriptsize

\setlength{\tabcolsep}{1.5pt}

\caption{\textbf{Ablation on evidence composition and arbitration.} }

\begin{adjustbox}{width=1.0\columnwidth}

\begin{tabular}{l cccc c cccc}

\toprule

\cellcolor{lightgray!30} &

\multicolumn{4}{c}{\cellcolor{lightgray!30} So-Fake-Set} &&

\multicolumn{4}{c}{\cellcolor{lightgray!30} So-Fake-OOD} \\

\cline{2-5} \cline{7-10}

\multirow{-2}{*}{\cellcolor{lightgray!30} \makecell[c]{Variant}} &

\cellcolor{lightgray!30} REAL &

\cellcolor{lightgray!30} FULL &

\cellcolor{lightgray!30} TAMP. &

\cellcolor{lightgray!30} Overall &&

\cellcolor{lightgray!30} REAL &

\cellcolor{lightgray!30} FULL &

\cellcolor{lightgray!30} TAMP. &

\cellcolor{lightgray!30} Overall \\

\midrule

Observer-only & 95.3 & 73.8 & 78.4 & 82.5 && 89.2 & 66.4 & 37.5 & 64.4 \\

Provider-only & 84.0 & 62.0 & 91.5 & 79.2 && 75.0 & 57.2 & 45.6 & 59.3 \\

Voting        & 95.5 & 73.5 & 92.0 & 87.0 && 89.0 & 67.5 & 46.0 & 67.5 \\

\cellcolor{softblue}\textbf{Full (So-Fake-R1)} &

\cellcolor{softblue}\textbf{98.0} &

\cellcolor{softblue}\textbf{76.5} &

\cellcolor{softblue}\textbf{99.0} &

\cellcolor{softblue}\textbf{91.2} &&

\cellcolor{softblue}\textbf{94.0} &

\cellcolor{softblue}\textbf{71.2} &

\cellcolor{softblue}\textbf{50.8} &

\cellcolor{softblue}\textbf{72.0} \\

\bottomrule

\end{tabular}

\end{adjustbox}

\label{table11}

\end{table}

\noindent\textbf{Evidence Composition and Arbitration.}
We ablate the evidence design by comparing the full So-Fake-R1 model against three variants: \textit{Observer-only}, \textit{Provider-only}, and \textit{Voting}, in which the Decision Maker is replaced by a confidence-weighted fusion rule.
All variants use the same backbone and training data.
As shown in Tab.~\ref{table11}, neither single evidence stream is sufficient: the Observer-only variant is less reliable for localized tampering, while the Provider-only variant tends to confuse global synthesis with local edits.
Voting improves over either stream, confirming their complementarity, but still falls behind the learned Decision Maker, especially on \textsc{Full\_Synthetic} and \textsc{Tampered} cases where the two reviewers disagree.
By re-examining the image and reasoning over semantic observations, forensic evidence, and their conflicts, the full So-Fake-R1 achieves the best performance on both So-Fake-Set and So-Fake-OOD.

\noindent\textbf{Decision-Maker Input Ablation.}
We further ablate the inputs available to the Decision Maker.
All variants use the same backbone and training data, differing only in whether the original image, observer report, and provider report are provided.
As shown in Tab.~\ref{table12}, the image-only Decision Maker falls below the full model, and removing the input image and relying only on textual reports is insufficient.
Removing the observer weakens semantic discrimination, while removing the provider reduces sensitivity to localized tampering.
The full model achieves the best overall performance on both splits.

\begin{figure*}[ht]
\centerline{\includegraphics[width=\linewidth]{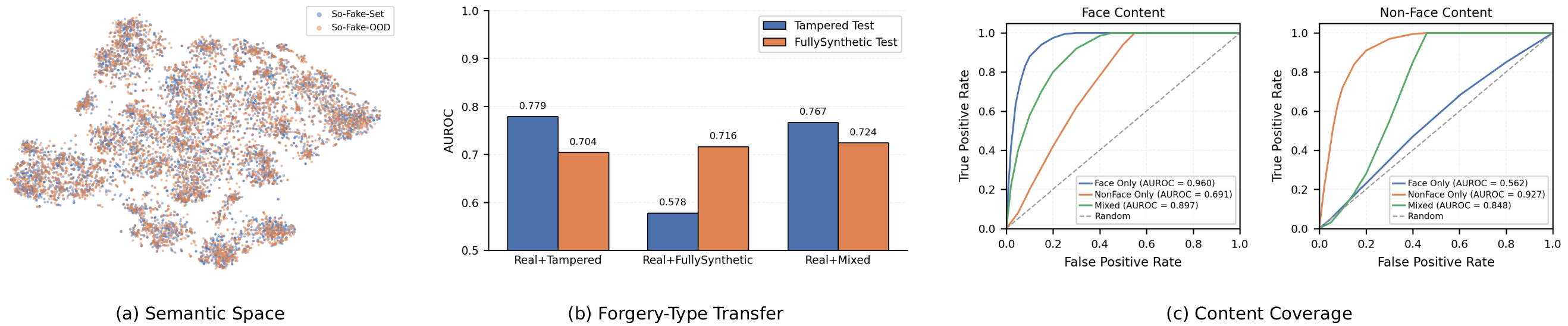}}

\caption{
\textbf{Benchmark analysis of So-Fake-OOD.}
(a) CLIP semantic-space visualization of randomly sampled images from the two splits.
(b) Cross-type transfer between tampered and full synthetic forgeries under different training settings.
(c) Face and non-face ROC curves showing the content-dependent generalization gap.}
\label{Figure4}
\end{figure*}

\begin{table}[t]
\centering
\scriptsize
\setlength{\tabcolsep}{1.5pt}
\caption{\textbf{Decision-maker input ablation.}}
\begin{adjustbox}{width=1.0\columnwidth}
\begin{tabular}{l cccc c cccc}
\toprule
\cellcolor{lightgray!30} &
\multicolumn{4}{c}{\cellcolor{lightgray!30} So-Fake-Set} &&
\multicolumn{4}{c}{\cellcolor{lightgray!30} So-Fake-OOD} \\
\cline{2-5} \cline{7-10}
\multirow{-2}{*}{\cellcolor{lightgray!30} \makecell[c]{Variant}} &
\cellcolor{lightgray!30} REAL &
\cellcolor{lightgray!30} FULL &
\cellcolor{lightgray!30} TAMP. &
\cellcolor{lightgray!30} Overall &&
\cellcolor{lightgray!30} REAL &
\cellcolor{lightgray!30} FULL &
\cellcolor{lightgray!30} TAMP. &
\cellcolor{lightgray!30} Overall \\
\midrule
Image-only DM &
94.8 & 71.8 & 90.2 & 85.6 &&
87.6 & 63.4 & 41.5 & 64.2 \\
w/o Input Image &
93.1 & 73.0 & 88.4 & 84.8 &&
88.0 & 66.2 & 43.7 & 66.0 \\
w/o Observer Report &
91.0 & 65.2 & 95.5 & 83.9 &&
85.5 & 58.5 & 48.9 & 64.3 \\
w/o Provider Report &
96.2 & 77.1 & 89.5 & 87.6 &&
91.0 & 69.8 & 43.2 & 68.0 \\
\cellcolor{softblue}
\textbf{Full (So-Fake-R1)} &
\cellcolor{softblue}\textbf{98.0} &\cellcolor{softblue} \textbf{76.5} & \cellcolor{softblue}\textbf{99.0} & \cellcolor{softblue}\textbf{91.2} &&\cellcolor{softblue}
\textbf{94.0} & \cellcolor{softblue}\textbf{71.2} & \cellcolor{softblue}\textbf{50.8} & \cellcolor{softblue}\textbf{72.0} \\

\bottomrule
\end{tabular}
\end{adjustbox}
\label{table12}
\end{table}

\noindent\textbf{Training Strategy Ablation.}
\begin{table}[t]

\centering

\scriptsize

\setlength{\tabcolsep}{1.5pt}

\caption{\textbf{Ablation on training strategy.}
}

\begin{adjustbox}{width=1.0\columnwidth}

\begin{tabular}{l cccc c cccc}

\toprule

\cellcolor{lightgray!30} &

\multicolumn{4}{c}{\cellcolor{lightgray!30} So-Fake-Set} &&

\multicolumn{4}{c}{\cellcolor{lightgray!30} So-Fake-OOD} \\

\cline{2-5} \cline{7-10}

\multirow{-2}{*}{\cellcolor{lightgray!30} \makecell[c]{Variant}} &

\cellcolor{lightgray!30} REAL &

\cellcolor{lightgray!30} FULL &

\cellcolor{lightgray!30} TAMP. &

\cellcolor{lightgray!30} Overall &&

\cellcolor{lightgray!30} REAL &

\cellcolor{lightgray!30} FULL &

\cellcolor{lightgray!30} TAMP. &

\cellcolor{lightgray!30} Overall \\

\midrule

Base VLM &

86.5 & 68.0 & 70.2 & 74.9 &&

80.5 & 60.3 & 38.0 & 59.6 \\

SFT only &

95.0 & 74.5 & 90.8 & 86.8 &&

88.7 & 67.0 & 45.0 & 66.9 \\

GRPO only &

90.2 & 70.0 & 80.5 & 80.2 &&

82.5 & 61.8 & 42.3 & 62.2 \\

\cellcolor{softblue}\textbf{SFT + GRPO} &

\cellcolor{softblue}\textbf{98.0} &

\cellcolor{softblue}\textbf{76.5} &

\cellcolor{softblue}\textbf{99.0} &

\cellcolor{softblue}\textbf{91.2} &&

\cellcolor{softblue}\textbf{94.0} &

\cellcolor{softblue}\textbf{71.2} &

\cellcolor{softblue}\textbf{50.8} &

\cellcolor{softblue}\textbf{72.0} \\

\bottomrule

\end{tabular}

\end{adjustbox}

\label{table13}

\end{table}
We evaluate the Decision Maker's training strategy.
As shown in Tab.~\ref{table13}, the base VLM performs poorly without task-specific adaptation, especially on TAMPERED and FULL \_SYNTHETIC cases.
SFT substantially improves the structured three-way decision ability, while GRPO-only training provides limited gains because of unstable output formatting and weak initialization.
Combining SFT and GRPO yields the best performance on both So-Fake-Set and So-Fake-OOD.
This indicates that SFT provides a reliable evidence-grounded response format, whereas GRPO further improves decision quality.

\noindent\textbf{Reward Function.}
Tab.~\ref{table14}(a) ablates the GRPO reward terms.
The authenticity reward alone is insufficient because it only supervises the binary REAL/FAKE decision.
Adding the scope reward improves fake-scope discrimination, while the arbitration reward improves conflict resolution between the two evidence sources.
The localization reward further strengthens spatial grounding for TAMPERED cases, and the format reward stabilizes structured outputs.
The full reward achieves the best overall performance.

\noindent\textbf{Policy Model.}
Tab.~\ref{table14}(b) compares different Decision Maker backbones under the same evidence inputs and training protocol.
Qwen2.5-VL-7B performs best across accuracy, IoU, and CSS, indicating that stronger visual reasoning and instruction-following ability better support our task.
\begin{table}[t!]
\vspace{-2mm}
\centering
\caption{\textbf{Ablation study of (a) reward functions and (b) policy models on the So-Fake-Set validation split.}}
\vspace{-2mm}
\label{table14}
\begin{minipage}{\columnwidth}
\begin{minipage}[t]{0.51\columnwidth}
    \centering
    \scriptsize
    \setlength{\tabcolsep}{1.5pt}
    \begin{adjustbox}{width=0.98\linewidth}
    \begin{tabular}{ccccc ccc}
    \toprule
    \cellcolor{lightgray!30} $R_{\mathrm{auth}}$ &
    \cellcolor{lightgray!30} $R_{\mathrm{scope}}$ &
    \cellcolor{lightgray!30} $R_{\mathrm{arb}}$ &
    \cellcolor{lightgray!30} $R_{\mathrm{loc}}$ &
    \cellcolor{lightgray!30} $R_{\mathrm{fmt}}$ &
    \cellcolor{lightgray!30} Acc. &
    \cellcolor{lightgray!30} IoU &
    \cellcolor{lightgray!30} CSS \\
    \midrule
    $\checkmark$ &  &  &  &  & 84.7 & 31.6 & 0.78 \\
    $\checkmark$ & $\checkmark$ &  &  &  & 88.9 & 41.7 & 0.81 \\
    $\checkmark$ & $\checkmark$ & $\checkmark$ &  &  & 89.8 & 43.5 & 0.84 \\
    $\checkmark$ & $\checkmark$ &  & $\checkmark$ &  & 90.4 & 47.2 & 0.83 \\
    $\checkmark$ & $\checkmark$ & $\checkmark$ & $\checkmark$ &  & 90.7 & 47.8 & 0.85 \\
    \rowcolor{softblue}
    $\checkmark$ & $\checkmark$ & $\checkmark$ & $\checkmark$ & $\checkmark$ &
    \textbf{91.2} & \textbf{48.6} & \textbf{0.88} \\
    \bottomrule
    \end{tabular}
    \end{adjustbox}
    
    (a)
\end{minipage}
\hfill
\begin{minipage}[t]{0.48\columnwidth}
    \centering
    \scriptsize
    \setlength{\tabcolsep}{1.5pt}
    \begin{adjustbox}{width=0.98\linewidth}
    \begin{tabular}{lccc}
    \toprule
    \cellcolor{lightgray!30} Policy Model &
    \cellcolor{lightgray!30} Acc. &
    \cellcolor{lightgray!30} IoU &
    \cellcolor{lightgray!30} CSS \\
    \midrule
    InternVL3-8B~\cite{zhu2025internvl3exploringadvancedtraining} & 89.0 & 41.5 & 0.83 \\
    DeepSeek-VL-7B~\cite{DBLP:journals/corr/abs-2403-05525} & 86.7 & 39.2 & 0.82 \\
    \rowcolor{softblue}
    Qwen2.5-VL-7B~\cite{DBLP:journals/corr/abs-2502-13923} & \textbf{91.2} & \textbf{48.6} & \textbf{0.88} \\
    \bottomrule
    \end{tabular}
    \end{adjustbox}
    
    (b)
\end{minipage}
\end{minipage}
\vspace{-2mm}
\end{table}

\begin{table}[t]
\centering
\scriptsize
\setlength{\tabcolsep}{1.5pt}
\caption{\textbf{Evidence reliability analysis on So-Fake-OOD.} }
\begin{adjustbox}{width=1.0\columnwidth}
\begin{tabular}{lccccc}
\toprule
\cellcolor{lightgray!30} Case &
\cellcolor{lightgray!30} Ratio &
\cellcolor{lightgray!30} Observer &
\cellcolor{lightgray!30} Evidence Provider &
\cellcolor{lightgray!30} Voting &
\cellcolor{lightgray!30} Final Acc \\
\midrule
Both correct   & 61.5 & 100.0 & 100.0 & 100.0 & \cellcolor{softblue}\textbf{98.5} \\
Observer only  & 10.2 & 100.0 & 0.0   & 28.5  & \cellcolor{softblue}\textbf{54.6} \\
Provider only  & 11.4 & 0.0   & 100.0 & 27.1  & \cellcolor{softblue}\textbf{52.3} \\
Both wrong     & 16.9 & 0.0   & 0.0   & 0.0   & \cellcolor{softblue}\textbf{10.2} \\
\midrule
Any-source error & 38.5 & 27.3 & 29.6 & 27.5 & \cellcolor{softblue}\textbf{34.5} \\
\bottomrule
\end{tabular}
\end{adjustbox}
\label{table15}
\end{table}
\noindent\textbf{Evidence Reliability Analysis.}
We analyze how the Decision Maker behaves when the evidence is imperfect.
We partition So-Fake-OOD samples according to whether the Visual Observer and the Forensic Evidence Provider each predict the correct scope, and compare the final Decision Maker with a confidence-weighted voting rule.
As shown in Tab.~\ref{table15}, both evidence streams are useful but imperfect.
Voting improves over a single stream in easy cases, but remains unreliable when the two sources disagree.
In contrast, the Decision Maker achieves higher accuracy in both observer-only and provider-only cases, showing that it does not simply copy one source.
More importantly, it recovers a portion of the both-wrong cases, where fixed voting cannot produce the correct answer.
This confirms that So-Fake-R1 benefits from learned evidence
arbitration and input-image re-examination.

\subsection{Analysis of So-Fake}
\label{subsec:analysis_dataset}

\begin{figure*}[t]
\centering
\includegraphics[width=1\textwidth]{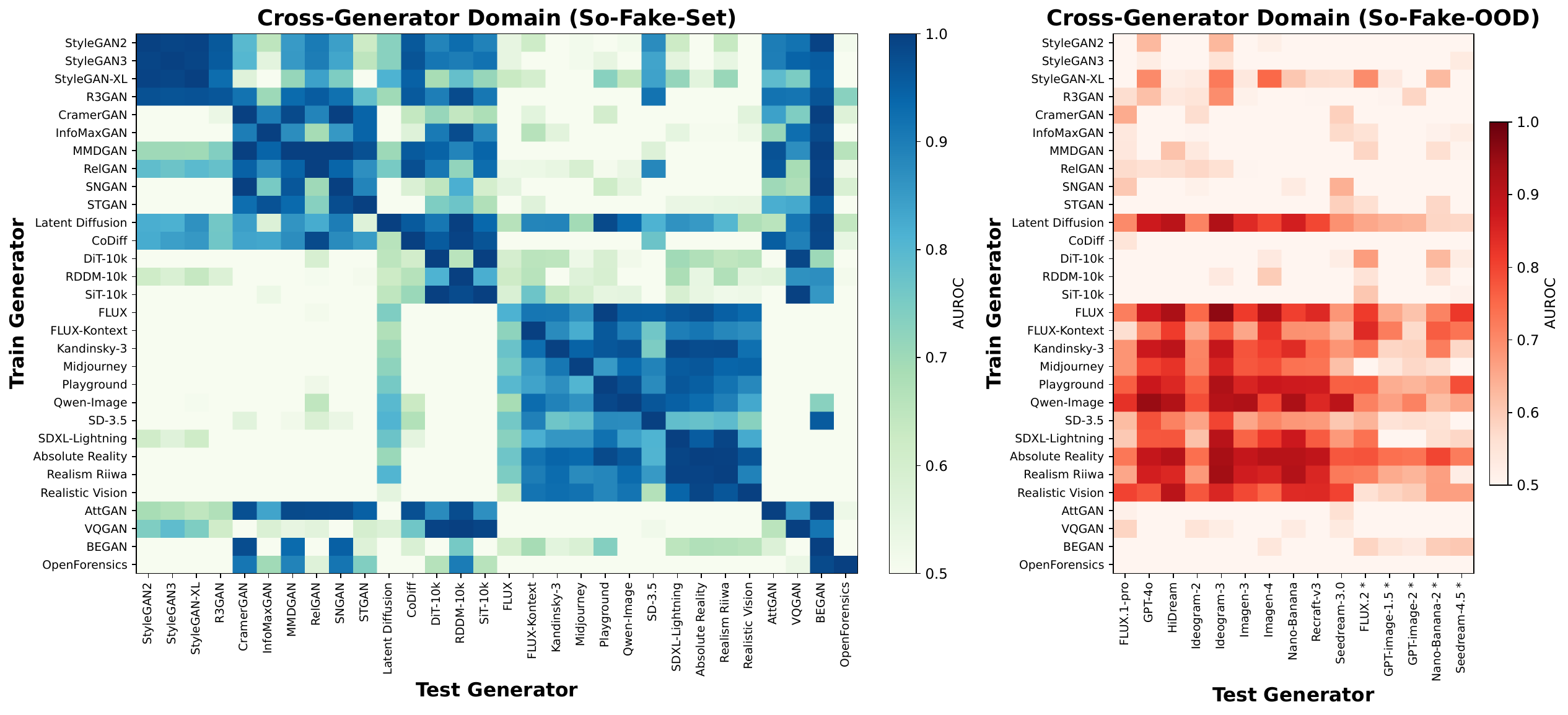}
\caption{\textbf{Cross-domain generalization across generator families.}
Each matrix reports cross-generator transfer performance, where CNNSpot~\cite{wang2019cnngenerated} is trained on forgeries from one generator and evaluated on each target generator; each cell reports AUROC against a fixed real subset. The block structure shows that detectors transfer better within related generator families but degrade across distinct generation paradigms, especially when moving from training generators in So-Fake-Set to commercial generators in So-Fake-OOD.}
\label{Figure6}
\end{figure*}

\subsubsection{Benchmark Analysis}
\label{subsec:BenchmarkAnalysis}

We analyze the composition and confounds of the compound OOD split used for robustness evaluation.
A useful OOD benchmark should preserve semantic coverage comparable to the training corpus while introducing distributional shifts that are absent from curated public datasets.
The platform-native image-statistics and tamper-geometry diagnostics have been reported in Sec.~\ref{subsec:dataset_statistics}. 
Here we further analyze semantic alignment, feature-level domain shift, content coverage, and forgery-type transfer.

\noindent\textbf{Semantic Alignment and Feature-Level Domain Shift.}
We first characterize the nature of the distribution shift introduced by So-Fake-OOD.
Under the unified 12-class taxonomy, the Jensen--Shannon divergence (JSD) between the real image subsets of So-Fake-Set and So-Fake-OOD is low (JSD $\approx 0.08$), indicating closely matched category-level coverage across the two splits; we compare real images so that the measurement reflects semantic coverage.
At the feature level, CLIP~\cite{DBLP:conf/icml/RadfordKHRGASAM21} embeddings from 10K images per split show substantial overlap, indicating that the two splits share semantic content while differing in image-level statistics (Fig.~\ref{Figure4}(a)).

\noindent\textbf{Content and Forgery-Type Coverage.}
Using UnivFD~\cite{DBLP:conf/cvpr/OjhaLL23} as a representative generator-agnostic detector, our analysis reveals two properties of social-media forgery detection, each with direct implications for how the task should be evaluated.
\textbf{First}, localized tampering and full-image synthesis expose partially distinct forensic cues.
A detector trained on tampered forgeries performs better on tampered test samples than on full synthetic ones; a detector trained on full synthetic forgeries shows the reverse pattern; and mixed training narrows this asymmetric gap (Fig.~\ref{Figure4}(b)).
Treating the two as separate scopes under a three-way REAL/FULL\_ SYNTHETIC/TAMPERED protocol therefore captures forensic structure that a single binary fake class would discard.
\textbf{Second}, generalization is strongly content-dependent.
A detector trained only on facial forgeries reaches $0.960$ AUROC on face content but decreases to $0.562$ on non-face content; combining face images with the remaining 11 categories rebalances performance to $0.897$ and $0.848$ AUROC respectively (Fig.~\ref{Figure4}(c)).
A face-centric benchmark would thus substantially overstate real-world generalization, whereas a broad 12-class taxonomy mitigates this bias.
Together with the preceding analysis, these findings show that both forgery type and content category affect generalization, and they directly inform the three-way protocol and 12-class taxonomy settings.

\subsubsection{Cross-Domain Generalization Analysis}
\label{subsubsec:cross_domain_generalization}

We examine whether detectors trained on So-Fake-Set learn transferable forgery cues or overfit to generator-specific artifacts.
Fig.~\ref{Figure6} presents cross-generator transfer matrices, where CNNSpot~\cite{wang2019cnngenerated} is trained on one generator group and evaluated on another.
The results reveal a clear block structure: transfer is stronger within related generator families, but degrades substantially across different generation paradigms.
In particular, detectors trained on GAN-based forgeries transfer poorly to diffusion-based samples, and models trained on open-source generators remain unreliable on held-out commercial generators.
This suggests that generator-specific cues remain an important component of the learned decision function.

To systematically analyze cross-domain generalization, we group the training generators into three families according to architecture and deployment context: GAN-based models, open-source diffusion models, and commercial models.
The full generator list is provided in Appendix Sec.~\ref{suppl:generative-methods}.
Tab.~\ref{table16} summarizes the same trend at the generator-family level.
Our cross-domain evaluation reveals distinct generalization patterns that reflect the evolving technological landscape of social-media forgery.
As shown in Tab.~\ref{table16} and Fig.~\ref{Figure6}, GAN-based methods achieve the strongest performance on So-Fake-Set but exhibit substantial degradation on So-Fake-OOD. Detectors trained on commercial-generator forgeries transfer best to So-Fake-OOD, which is expected since all OOD generators are commercial.
\begin{table}[t]
\centering
\scriptsize
\setlength{\tabcolsep}{1.5pt}
\caption{\textbf{Generator-family transfer analysis.}
Mean AUROC is reported over representative generator groups.
$\Delta_{\mathrm{OOD-Set}}$ denotes So-Fake-OOD minus So-Fake-Set AUROC.}
\begin{adjustbox}{width=1.0\columnwidth}
\begin{tabular}{lcccc}
\toprule
\cellcolor{lightgray!30} Generator Family &
\cellcolor{lightgray!30} \#Models &
\cellcolor{lightgray!30} Set AUROC &
\cellcolor{lightgray!30} OOD AUROC &
\cellcolor{lightgray!30} $\Delta_{\mathrm{OOD-Set}}$ \\
\midrule
GAN-based             & 14 & 0.59 & 0.39 & $-0.20$ \\
Open-source Diffusion & 13  & 0.58 & 0.47 & $-0.11$ \\
Commercial Diffusion  & 3  & 0.48 & 0.73 & $\phantom{-}0.25$ \\
\bottomrule
\end{tabular}
\end{adjustbox}
\label{table16}
\end{table}

\begin{table}[t!]
\vspace{-2mm}
\centering
\caption{\textbf{Analysis of generator diversity and training scale.}
(a) Effect of generator diversity using UnivFD. (b) Effect of training scale using CNNSpot.}
\vspace{-2mm}
\label{table17}
\begin{minipage}{\columnwidth}
\begin{minipage}[t]{0.49\columnwidth}
    \centering
    \scriptsize
    \setlength{\tabcolsep}{1.5pt}
    \begin{adjustbox}{width=0.98\linewidth}
    \begin{tabular}{lcc}
    \toprule
    \cellcolor{lightgray!30} Training Subset &
    \cellcolor{lightgray!30} Set &
    \cellcolor{lightgray!30} OOD \\
    \midrule
    StyleGAN3 & 46.8 & 34.2 \\
    LDM & 49.3 & 34.6 \\
    StyleGAN3 + LDM & 49.2 & 34.9 \\
    \rowcolor{softblue}
    \textbf{Full So-Fake-Set} & \textbf{82.4} & \textbf{48.6} \\
    \bottomrule
    \end{tabular}
    \end{adjustbox}
    
    (a)
\end{minipage}
\hfill
\begin{minipage}[t]{0.35\columnwidth}
    \centering
    \scriptsize
    \setlength{\tabcolsep}{1.5pt}
    \begin{adjustbox}{width=0.98\linewidth}
    \begin{tabular}{lc}
    \toprule
    \cellcolor{lightgray!30} Training Scale &
    \cellcolor{lightgray!30} OOD \\
    \midrule
    10\%  & 38.3 \\
    30\%  & 49.4 \\
    50\%  & 49.9 \\
    \rowcolor{softblue}
    100\%  & \textbf{54.2} \\
    \bottomrule
    \end{tabular}
    \end{adjustbox}
    
    (b)
\end{minipage}
\end{minipage}
\vspace{-2mm}
\end{table}

\subsubsection{Generator Diversity and Training Scale}
\label{subsubsec:generator_scale}

We analyze whether the scale and generator coverage of So-Fake-Set are necessary for cross-domain generalization.
Tab.~\ref{table17}(a) studies generator diversity using UnivFD~\cite{DBLP:conf/cvpr/OjhaLL23} as a representative detector.
Training on a single generator family, such as StyleGAN3~\cite{DBLP:conf/nips/KarrasALHHLA21} or LDM~\cite{DBLP:conf/cvpr/RombachBLEO22}, leads to poor performance on both So-Fake-Set and So-Fake-OOD.
Even combining the two generators yields performance far below that obtained by training on the full multi-family So-Fake-Set.
This suggests that small generator subsets are insufficient for learning transferable forensic cues, and that broad generator coverage is essential for reducing overfitting to generator-specific artifacts.
Tab.~\ref{table17}(b) studies the effect of training scale using CNNSpot~\cite{wang2019cnngenerated}.
When training subsets are sampled from So-Fake-Set while preserving class and generator coverage, OOD accuracy steadily improves as the training scale increases.
This shows that scale further strengthens cross-domain generalization once diversity is maintained.
Together, these results support the design of So-Fake-Set as a large-scale, multi-generator training corpus for social-media image forgery detection.


\section{Conclusion}
\label{sec:Conclusion}

We introduced \textbf{So-Fake}, a social-media benchmark for image forgery detection, consisting of \textbf{So-Fake-Set} for training and validation and \textbf{So-Fake-OOD} for held-out evaluation on native social-media sources and unseen commercial generators.
The benchmark covers three authenticity scopes and 12 diverse
semantic categories, enabling evaluation beyond binary and
face-centric forgery detection.
Our analyses show that performance varies across platform, generator, content category, and forgery scope, highlighting the need for evidence-grounded evaluation.
We further proposed \textbf{So-Fake-R1}, which builds complementary semantic and forensic evidence views, applies an Evidence Gate for direct acceptance or routed arbitration, and invokes a Decision Maker only for weakly supported cases.
Together, So-Fake and So-Fake-R1 provide a benchmark and an evidence-structured reference framework for future social-media forgery research.

{\small
\vspace{-3mm}
\bibliographystyle{IEEEtran}
\bibliography{IEEEabrv,main}
}
\clearpage

\appendices

\renewcommand{\thesubsectiondis}{\arabic{subsection}.}
\renewcommand{\thesubsection}
{\mbox{\thesection-\arabic{subsection}}}

\setcounter{section}{0}
\setcounter{subsection}{0}
\setcounter{table}{0}
\setcounter{figure}{0}
\setcounter{equation}{0}

\makeatletter
\renewcommand*{\theHfigure}
  {appendix.\Alph{section}.\arabic{figure}}
\renewcommand*{\theHtable}
  {appendix.\Alph{section}.\arabic{table}}
\renewcommand*{\theHequation}
  {appendix.\Alph{section}.\arabic{equation}}
\renewcommand*{\theHalgorithm}
  {appendix.\Alph{section}.\arabic{algorithm}}
\makeatother

\section{Data Construction}
\label{suppl:data-construction}

In this section, we provide additional details on the construction of So-Fake, complementing the description in the main paper.
We first present the complete list of generative and editing methods used to produce full synthetic and tampered images in So-Fake-Set and So-Fake-OOD (Sec.~\ref{suppl:generative-methods}).
We then describe the platform-native collection and release protocol for authentic social-media imagery in So-Fake-OOD, including the data sources, source-data filtering, and platform-specific release forms (Sec.~\ref{suppl:platform-collection}).
To assess potential overlap in the OOD evaluation, we further conduct a sampled duplicate-detection analysis between So-Fake-Set and So-Fake-OOD (Sec.~\ref{suppl:duplicate}).
Next, we detail the quality-control procedure applied to generated samples, including automatic aesthetic filtering and additional expert review for So-Fake-OOD (Sec.~\ref{suppl:quality-control}).
Finally, we describe the construction of human-verified reference rationales used for explanation evaluation (Sec.~\ref{suppl:ground-truth}).

\subsection{Complete Generative and Editing Methods}
\label{suppl:generative-methods}
We summarize the generative and editing methods used to synthesize full synthetic and tampered images in So-Fake-Set and So-Fake-OOD, as shown in Tab.~\ref{table2}.
In total, we employ 45 generative or editing methods to improve the coverage and diversity of the benchmark.
To improve architectural coverage, we include both GAN-based and diffusion-based models.
While some included methods are relatively outdated, we deliberately retain a small subset to reflect historical generation artifacts and to test robustness to older synthesis pipelines.
For recent high-fidelity and openly accessible models such as FLUX~\cite{flux2024}, SD-3.5~\cite{DBLP:conf/icml/EsserKBEMSLLSBP24}, SD-XL~\cite{DBLP:conf/iclr/PodellELBDMPR24}, and Latent Diffusion~\cite{DBLP:conf/cvpr/RombachBLEO22}, we generate more samples but cap their contribution within each evaluation split to avoid dominance by any single model family.

\begin{table*}[th]
\caption{
\textbf{Details of generative methods used in constructing So-Fake-Set and So-Fake-OOD.}
Column abbreviations: Set = So-Fake-Set; OOD = So-Fake-OOD; F = fully synthetic images; T = tampered images.
Real data source abbreviations: F30k = Flickr30k, OI = OpenImages,  OF = OpenForensics.
}

\resizebox{\textwidth}{!}{
\begin{tabular}{cccccccccc}
\toprule[1.5pt]
\label{table2}
\cellcolor{lightgray!30}ID-Number                  &  \cellcolor{lightgray!30}Method  &\cellcolor{lightgray!30} Model Type & \cellcolor{lightgray!30}Data Used                     & \cellcolor{lightgray!30}Generation Target                  & \cellcolor{lightgray!30}Year &\cellcolor{lightgray!30} Venue         &\cellcolor{lightgray!30} Real Data Source                                                 & \cellcolor{lightgray!30}Data Scale & \cellcolor{lightgray!30}Code Link  \\ \midrule
 1         &  Absolute\_Reality~\cite{AbsoluteReality}  &  Diffusion      &  Set                &  F             &  2023 &  None          &  F30k \& WIDER                                               &  50{,}000      &  \href{https://huggingface.co/Lykon/AbsoluteReality}{URL}  \\
2         & AttGAN~\cite{DBLP:journals/corr/abs-1711-10678}         & GAN          & Set                & F             & 2017 & PAMI          & FFHQ \& CelebA                                                   & 2{,}000       & \href{https://github.com/elvisyjlin/AttGAN-PyTorch}{URL}  \\
3         & BEGAN~\cite{DBLP:journals/corr/BerthelotSM17}            & GAN        & Set                & F             & 2017 & NeurIPS       & FFHQ \& CelebA                                                   & 2{,}000       & \href{https://github.com/artcg/BEGAN}{URL}  \\

4         & Collaborative Diffusion~\cite{DBLP:conf/cvpr/HuangC0023}   & Diffusion & Set                & F             & 2023 & CVPR          & FFHQ \& CelebA                                                   & 30{,}000      &\href{https://github.com/ziqihuangg/Collaborative-Diffusion}{URL} \\
5         & CramerGAN~\cite{DBLP:journals/corr/BellemareDDMLHM17}   & GAN             & Set                & F             & 2017 & NeurIPS       & FFHQ \& CelebA                                                   & 2{,}000       &\href{https://github.com/jiamings/cramer-gan}{URL} \\
6         & DiT-XL~\cite{DBLP:conf/iccv/PeeblesX23}                 & Diffusion  & Set                & F             & 2023 & ICCV          & FFHQ \& CelebA                                                   & 10{,}000      &\href{https://github.com/facebookresearch/DiT}{URL} \\

7         & FLUX.1 Full [Dev]~\cite{flux2024}       & Diffusion              & Set & F \& T & 2024 & None          & FFHQ \& CelebA \& OI \& COCO  \& OF  & 350{,}000+   &\href{https://huggingface.co/black-forest-labs/FLUX.1-dev}{URL} \\

8         & InfoMaxGAN~\cite{DBLP:conf/wacv/LeeTC21}    & GAN           & Set                & F             & 2021 & WACV          & FFHQ \& CelebA                                                   & 2{,}000       &\href{https://github.com/kwotsin/mimicry}{URL} \\
9         & Kandinsky3~\cite{arkhipkin2023kandinsky}     & Diffusion          & Set                & F             & 2023 & None          & F30k \& WIDER                                               & 80{,}000      &\href{https://github.com/ai-forever/Kandinsky-3}{URL} \\
10        & Latent Diffusion~\cite{DBLP:conf/cvpr/RombachBLEO22}   & Diffusion     & Set                & F \& T & 2022 & CVPR          & F30k \& WIDER \&COCO \& OI                          & 250{,}000+    &\href{https://github.com/CompVis/latent-diffusion}{URL} \\
11        & MMDGAN~\cite{DBLP:conf/nips/LiCCYP17}     & GAN              & Set                & F             & 2017 & NeurIPS       & FFHQ \& CelebA                                                   & 2{,}000       &\href{https://github.com/OctoberChang/MMD-GAN}{URL} \\
12        & Playground~\cite{DBLP:journals/corr/abs-2402-17245}     & Commercial           & Set                & F             & 2024 & None          & F30k \& WIDER                                               & 50{,}000      &\href{https://huggingface.co/playgroundai/playground-v2.5-1024px-aesthetic}{URL} \\
13        & R3GAN~\cite{DBLP:conf/nips/HuangGK024}   & GAN                 & Set                & F             & 2024 & NeurIPS       & FFHQ \& CelebA                                                   & 30{,}000      &\href{https://github.com/brownvc/R3GAN}{URL} \\
14        & RDDM~\cite{DBLP:conf/cvpr/LiuWFWTQ24}      & Diffusion               & Set                & F             & 2024 & CVPR          & FFHQ \& CelebA                                                   & 10{,}000      &\href{https://github.com/nachifur/RDDM}{URL} \\
15        & rRealism\_riiwa~\cite{rRealism_riiwa}    & Diffusion       & Set                & F             & 2024 & None          & F30k \& WIDER                                               & 50{,}000      &\href{https://huggingface.co/digiplay/rRealism_v1.0_riiwa}{URL} \\
16        & RelGAN~\cite{DBLP:conf/iclr/NieNP19}     & GAN              & Set                & F             & 2019 & ICLR          & FFHQ \& CelebA                                                   & 2{,}000       &\href{https://github.com/elvisyjlin/RelGAN-PyTorch}{URL} \\
17        & SD-3.5~\cite{DBLP:conf/icml/EsserKBEMSLLSBP24}    & Diffusion                & Set                & F             & 2024 & ICML          & FFHQ \& CelebA \& OI \& COCO  \& F30k                & 100{,}000+    &\href{https://github.com/Stability-AI/sd3.5}{URL} \\
18        & SD-XL~\cite{DBLP:conf/iclr/PodellELBDMPR24}    & Diffusion              & Set  & F \& T & 2024 & ICLR          & F30k \& WIDER \& COCO  \&OF             & 300{,}000+    &\href{https://huggingface.co/stabilityai/stable-diffusion-xl-base-1.0}{URL} \\
19        & StyleGAN-2~\cite{DBLP:conf/cvpr/KarrasLAHLA20}   & GAN            & Set                & F             & 2020 & CVPR          & FFHQ \& CelebA                                                   & 30{,}000      &\href{https://github.com/NVlabs/stylegan2}{URL} \\
20        & StyleGAN-3~\cite{DBLP:conf/nips/KarrasALHHLA21}    & GAN           & Set                & F             & 2021 & NeurIPS       & FFHQ \& CelebA                                                   & 30{,}000      &\href{https://github.com/NVlabs/stylegan3}{URL} \\
21        & StyleGAN-XL~\cite{DBLP:conf/siggraph/SauerS022}   & GAN           & Set                & F             & 2022 & SIGGRAPH & FFHQ \& CelebA                                                   & 30{,}000      &\href{https://github.com/autonomousvision/stylegan-xl}{URL} \\
22        & SiT-XL~\cite{DBLP:conf/eccv/MaGABVX24}       & Diffusion            & Set                & F             & 2024 & ECCV          & FFHQ \& CelebA                                                   & 10{,}000      &\href{https://github.com/willisma/SiT}{URL} \\
23        & SNGAN~\cite{DBLP:conf/iclr/MiyatoKKY18}     & GAN               & Set                & F             & 2018 & ICLR          & FFHQ \& CelebA                                                   & 2{,}000       &\href{https://github.com/christiancosgrove/pytorch-spectral-normalization-gan}{URL} \\
24        & STGAN~\cite{DBLP:conf/cvpr/0018DXLDZW19}     & GAN               & Set                & F             & 2019 & CVPR          & FFHQ \& CelebA                                                   & 2{,}000       &\href{https://github.com/csmliu/STGAN}{URL} \\
25        & VQGAN~\cite{DBLP:conf/iclr/YuLKZPQKXBW22}     & GAN               & Set                & F             & 2022 & ICLR          & FFHQ \& CelebA                                                   & 2{,}000       &\href{https://github.com/CompVis/taming-transformers}{URL} \\
26        & OpenForensics~\cite{DBLP:conf/iccv/LeNYE21}    & GAN             & Set                & F             & 2021 & ICCV          & Google Open Images                                                           & 10{,}000      &\href{https://github.com/ltnghia/openforensics?tab=readme-ov-file}{URL} \\
27        & Realistic Vision~\cite{RealisticVision}    &Diffusion            & Set                & F             & 2023 & None          & F30k \& WIDER \& COCO  \&OF                                                          & 50{,}000       &\href{https://huggingface.co/stablediffusionapi/realistic-vision-v51}{URL} \\

28         & FLUX.1-Kontext-dev ~\cite{flux2024}       & Diffusion             & Set & F  & 2025 & None          &  OpenImages   & 8{,}000+ &\href{https://docs.bfl.ai/kontext/kontext_text_to_image}{URL} \\

29         & Qwen-image ~\cite{DBLP:journals/corr/abs-2508-02324}       & Commercial             & Set & F  & 2025 & arXiv          &  OpenImages   & 3{,}000+ &\href{https://github.com/QwenLM/Qwen-Image}{URL} \\

30         & Midjourney ~\cite{midjourney2023}       & Commercial             & Set & F  & 2023 & None          &  OpenImages   & 8{,}000+ &\href{https://www.midjourney.com/home}{URL} \\

31        & Recraft-v3~\cite{recraft}    & Commercial            & OOD                & F  & 2024 & None          & Reddit                                                           & 5{,}000+      &\href{https://www.recraft.ai/}{URL} \\
32        & GPT-4o\cite{DBLP:journals/corr/abs-2410-21276}  & Commercial   & OOD                & F \& T & 2025 & arXiv         & Reddit                                                           & 8{,}000+      &\href{https://openai.com/index/hello-gpt-4o/}{URL} \\
33       & Imagen 3~\cite{DBLP:journals/corr/abs-2408-07009}   & Commercial                 & OOD                & F             & 2025 & arXiv         & Reddit                                                           & 3{,}000+      &\href{https://deepmind.google/technologies/imagen-3/}{URL} \\

34       & Imagen 4~\cite{Imagen4.0}   &Commercial                 & OOD                & F             & 2025 & arXiv         & Reddit                                                           & 3{,}000+      &\href{https://image4.com/}{URL} \\

35       & Nano Banana~\cite{Nano-banana}   & Commercial                 & OOD                & F             & 2025 & None        & Reddit                                                           & 2{,}000+      &\href{https://aistudio.google.com/models/gemini-2-5-flash-image}{URL} \\

36       & Seedream 3.0~\cite{DBLP:journals/corr/abs-2504-11346}   &Commercial            & OOD                & F             & 2025 & arXiv          & Reddit                                                           & 5{,}000+       &\href{https://preview.reve.art/}{URL} \\
37        & Ideogram 3.0~\cite{Ideogram3.0}   & Commercial            & OOD                & F \& T & 2025 & None          & Reddit                                                           & 5{,}000+      &\href{https://about.ideogram.ai/3.0}{URL} \\

38       & Ideogram 2.0~\cite{Ideogram2.0}   & Commercial             & OOD                & F             & 2024 & None          & Reddit                                                           & 5{,}000+      &\href{https://about.ideogram.ai/2.0}{URL} \\
39        & FLUX 1.1\_pro~\cite{flux1.1pro}   & Commercial           & OOD                & F             & 2025 & None          & Reddit                                                           & 3{,}000+      &\href{https://replicate.com/black-forest-labs/flux-1.1-pro}{URL} \\
40       & HiDream3~\cite{HiDream}     & Commercial            & OOD                & F             & 2025 & None          & Reddit                                                           & 3{,}000+       &\href{https://github.com/HiDream-ai/HiDream-I1}{URL} \\
41       & GPT-Image-1.5~\cite{GPT-Image-1.5}     & Commercial             & OOD                & F             & 2026 & None          & Bluesky, Tumblr                                                          & 2{,}500+       &\href{https://community.openai.com/t/gpt-image-1-5-rolling-out-in-the-api-and-chatgpt/1369443}{URL} \\
42       & GPT-Image-2~\cite{GPT-Image-2}     &Commercial            & OOD                & F             & 2026 & None          & Bluesky, Tumblr                                                         & 1{,}500+       &\href{https://community.openai.com/t/introducing-gpt-image-2-available-today-in-the-api-and-codex/1379479}{URL} \\
43       & Nano Banana 2~\cite{Nano-banana-2}     & Commercial             & OOD                & F             & 2026 & None          & Bluesky, Tumblr                                                          & 1{,}000+       &\href{https://gemini.google/overview/image-generation/}{URL} \\
44       & FLUX 2~\cite{flux2}     & Commercial             & OOD                & F             & 2026 & None          & Bluesky, Tumblr                                                           & 1{,}500+       &\href{https://bfl.ai/models/flux-2}{URL} \\
45       & Seedream4.5~\cite{Seedream4.5}     & Commercial             & OOD                & F             & 2026 & None          & Bluesky, Tumblr                                                          & 1{,}000+       &\href{https://seed.bytedance.com/en/seedream4_5}{URL} \\

\bottomrule[1.5pt]
\vspace{-2mm}
\end{tabular}
}
\end{table*}

\subsection{Platform-Native Collection and Release}
\label{suppl:platform-collection}

So-Fake-OOD draws authentic imagery from five social-media platforms—Reddit, X, Instagram, Bluesky, and Tumblr—to capture platform-native characteristics that are underrepresented in the curated public datasets used to construct So-Fake-Set.

\noindent\textbf{Source Data Filtering.}
To make the benchmark suitable for public research release, we apply a dedicated source-data filtering stage to the collected real-image candidates.
Five reviewers inspect candidate images and remove samples containing sensitive, personally identifiable, harmful, or task-irrelevant content.
In addition, reviewers are instructed to remove images with obvious AI-generation or heavy-editing traces, such as distorted object structure, unnatural text, inconsistent shadows, synthetic-looking faces or hands, or other visible artifacts that make the image unreliable as a platform-native REAL sample.
After filtering, So-Fake-OOD retains approximately 33K real images.

\noindent\textbf{Release Policy.}
Consistent with platform-specific redistribution policies, we release the retained images differently across platforms.
For Reddit, Bluesky, and Tumblr, we release filtered images directly for non-commercial academic use when permitted by the applicable platform terms and dataset license.
For X and Instagram, whose terms restrict the third-party redistribution of image content, we release only links to the original sources, allowing users to retrieve the content in accordance with each platform’s terms, as summarized in Tab.~\ref{sup:table1}.

\noindent\textbf{Consent and Removal Requests.}
We did not seek individual informed consent from users depicted in the collected images, as this is infeasible at the scale required for a cross-platform benchmark.
Collection instead relies on content that users chose to post publicly under the hosting platform's terms.
To respect the interests of individuals who appear in the released images, we provide a removal channel: any depicted individual may request that their image, and any synthetic or tampered variant derived from it, be excluded from subsequent releases of So-Fake-OOD by contacting zhenglin@liverpool.ac.uk. 

\begin{figure}[t]
\begin{center}
\centerline{\includegraphics[width=1.0\columnwidth]{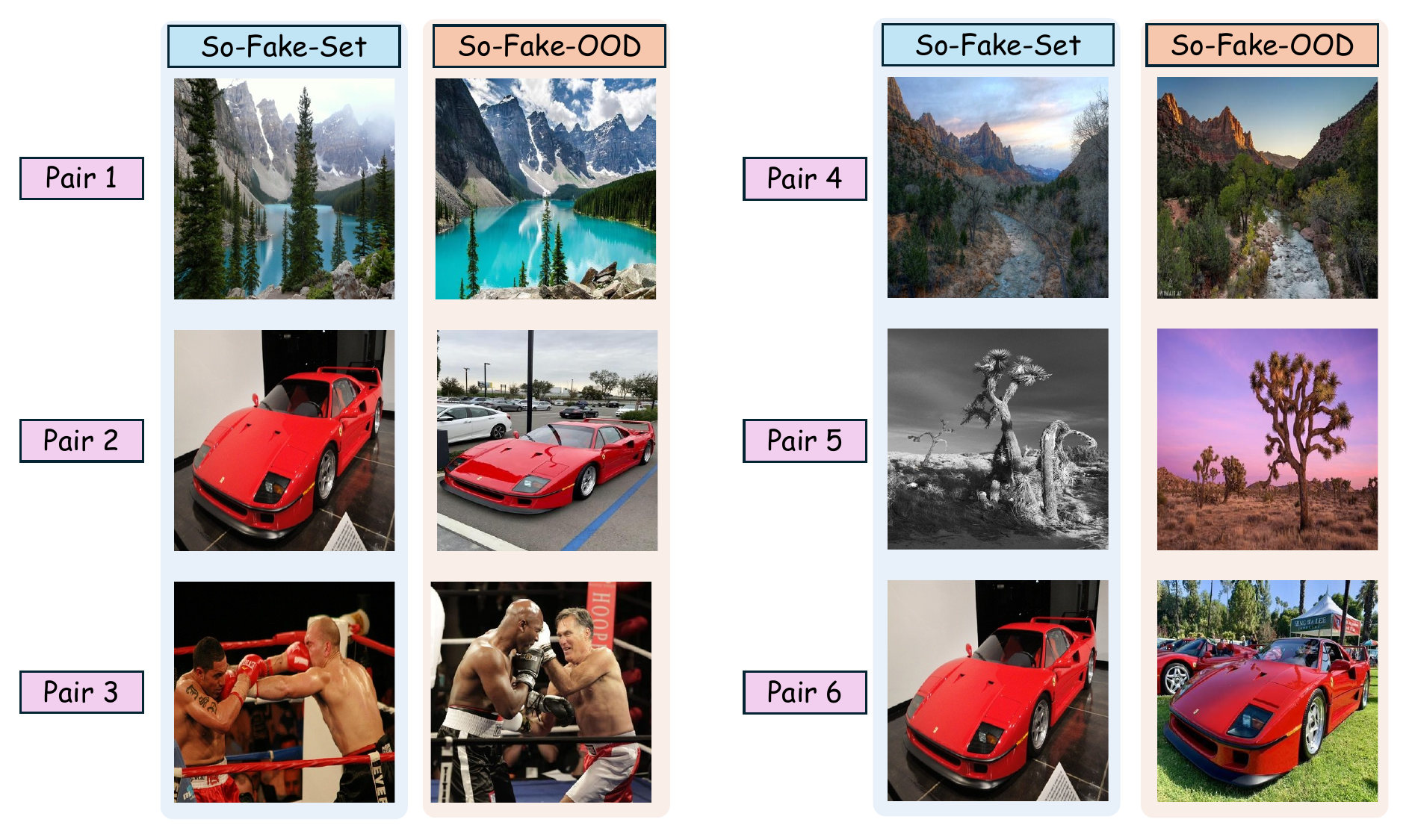}} 
\caption{\textbf{Representative high-similarity pairs identified by DINOv3 (cosine similarity $\geq$ 0.9) between So-Fake-Set and So-Fake-OOD.} Visual inspection confirms these are distinct images with topic-level similarity rather than actual duplicates, demonstrating minimal overlap between the two splits.}
\label{sup_figure1}
\end{center}
\end{figure}

\begin{table}[t]
\centering
\scriptsize
\setlength{\tabcolsep}{1.5pt}
\caption{\textbf{Platform-native real-image sources and release forms in So-Fake-OOD.}}
\begin{adjustbox}{width=1.0\columnwidth}
\begin{tabular}{lcccc}
\toprule
\cellcolor{lightgray!30} Platform &
\cellcolor{lightgray!30} Count &
\cellcolor{lightgray!30} Access Source &
\cellcolor{lightgray!30} Released Form &
\cellcolor{lightgray!30} Release Constraint \\
\midrule
Reddit & 20K & Official API & Filtered images & API redistribution \\
X & 7K & Platform links & Source hyperlinks & Link-only sharing \\
Instagram & 3K & Platform links & Source hyperlinks & Link-only sharing \\
Tumblr & 2K & Official API & Filtered images & API redistribution \\
Bluesky & 1.5K & Official API & Filtered images & API redistribution \\
\bottomrule
\end{tabular}
\end{adjustbox}
\label{sup:table1}
\end{table}

\begin{figure}[t]
    \centering
    \includegraphics[width=\columnwidth]{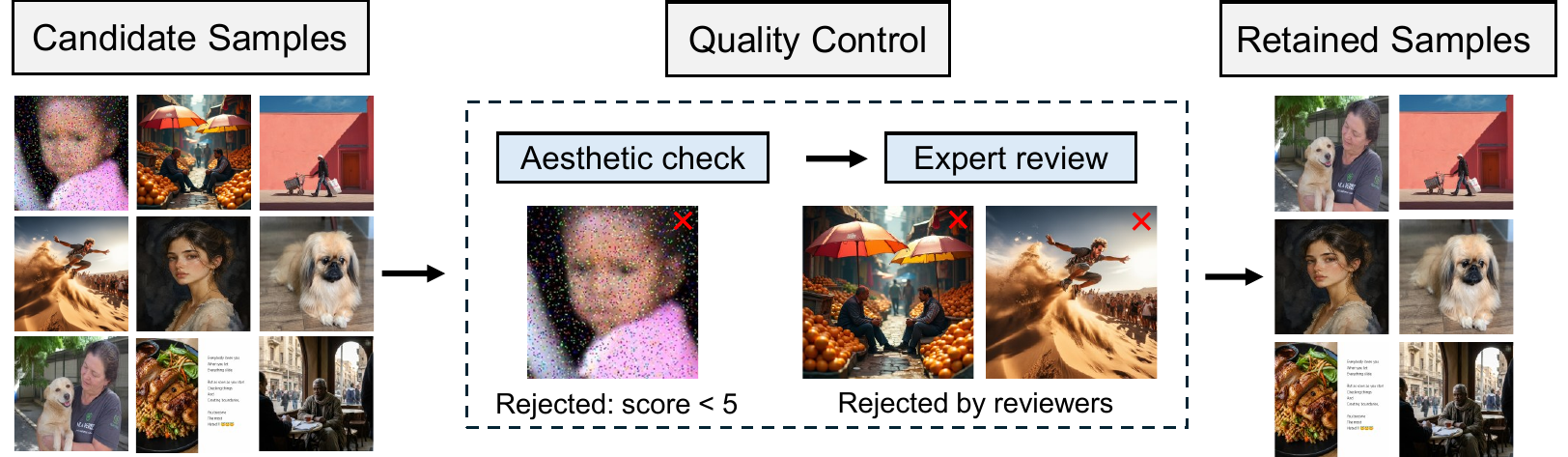}
    \caption{\textbf{Illustration of the generated-sample quality-control process.}
    Representative FULL\_SYNTHETIC and TAMPERED candidates rejected because of low perceptual quality, severe visual artifacts, semantic inconsistency, failed localized editing, or unsuitable content.}
    \label{figure:QD}
\end{figure}

\subsection{Duplicate Detection Analysis}
\label{suppl:duplicate}
To assess potential near-duplicate overlap between So-Fake-Set and So-Fake-OOD, we conducted a sampled similarity analysis using DINOv3 (ViT-L/16)~\cite{DBLP:journals/corr/abs-2508-10104}, a strong self-supervised visual representation model.
We evaluated duplicate similarity under the following configuration:

\noindent\textbf{Data Sampling.} Due to computational constraints, we randomly sample 10\% of the real images from each split (seed = 42), yielding 65{,}000 real images from So-Fake-Set and approximately 3{,}300 from So-Fake-OOD.

\noindent\textbf{Similarity Metric.} We compute cosine similarity between normalized DINOv3 embeddings and use a threshold of 0.9 to flag potential near duplicates.

\noindent\textbf{Results.} The experiment flagged 19 high-similarity candidate pairs across the two subsets.
We manually inspected all 19 pairs; representative examples are shown in Fig.~\ref{sup_figure1}.
Visual inspection suggests that these pairs are distinct images exhibiting semantic or compositional similarity rather than exact duplicates.
For instance, Pair 1 shows two photographs of Moraine Lake captured under different lighting and weather conditions, and Pair 2 depicts the same vehicle model photographed in different settings.

\subsection{Generated-Sample Quality Control}
\label{suppl:quality-control}

To reduce trivial generation failures and ensure that So-Fake-OOD provides a meaningful cross-domain evaluation, we apply quality control to generated samples at two levels.

\noindent\textbf{Automatic Aesthetic Filtering.}
All FULL\_SYNTHETIC and TAMPERED samples generated for both So-Fake-Set and So-Fake-OOD are evaluated with the LAION aesthetic predictor~\cite{DBLP:conf/nips/SchuhmannBVGWCC22}.
Samples with an aesthetic score below 5 are discarded to remove obvious generation failures, visual artifacts, and images of insufficient perceptual quality.

\noindent\textbf{Expert Review of So-Fake-OOD.}
Because So-Fake-OOD serves as the held-out evaluation benchmark, its FULL\_SYNTHETIC and TAMPERED candidates undergo an additional expert-review stage.
Five expert reviewers independently rate each candidate on a five-point scale along three dimensions: visual realism, semantic consistency, and appropriateness for benchmark inclusion.
A candidate is retained only if, for each dimension, at least three of the five reviewers assign a score of 3 or higher.
Borderline cases and samples with substantial reviewer disagreement are subjected to a second review round before the final decision.
This additional screening removes approximately 20\% of candidate So-Fake-OOD generations, reducing the likelihood that OOD performance is driven by trivial low-quality artifacts rather than genuine distribution shift.
Representative samples rejected during quality control are shown in Fig.~\ref{figure:QD}.

\subsection{Explanation Reference Construction}
\label{suppl:ground-truth}

For explanation evaluation, we construct two sets of human-verified reference rationales, each containing 3{,}000 samples: one drawn from the So-Fake-Set validation split and the other from the So-Fake-OOD evaluation set.
Each reference describes the visual evidence supporting the assigned authenticity annotation.
These rationales serve as reference texts for CSS and ROUGE-L evaluation, complementing the human preference study reported in the main paper.

\noindent\textbf{Evidence-conditioned Generation.}
We first generate an initial rationale with Claude Opus 4.8~\cite{claude}, conditioned on the image, its annotation context, and a fixed rationale-generation prompt.
For \textbf{TAMPERED} samples, the ground-truth mask is provided to constrain the description to the manipulated region.
For \textbf{FULL\_SYNTHETIC}, the rationale describes image-level generation traces; for \textbf{REAL} samples, it describes evidence of natural image acquisition.
The model is therefore used to verbalize evidence for a given annotation, rather than to infer the authenticity label or localization target.

\begin{figure*}[t]
\centering
\includegraphics[width=\textwidth]{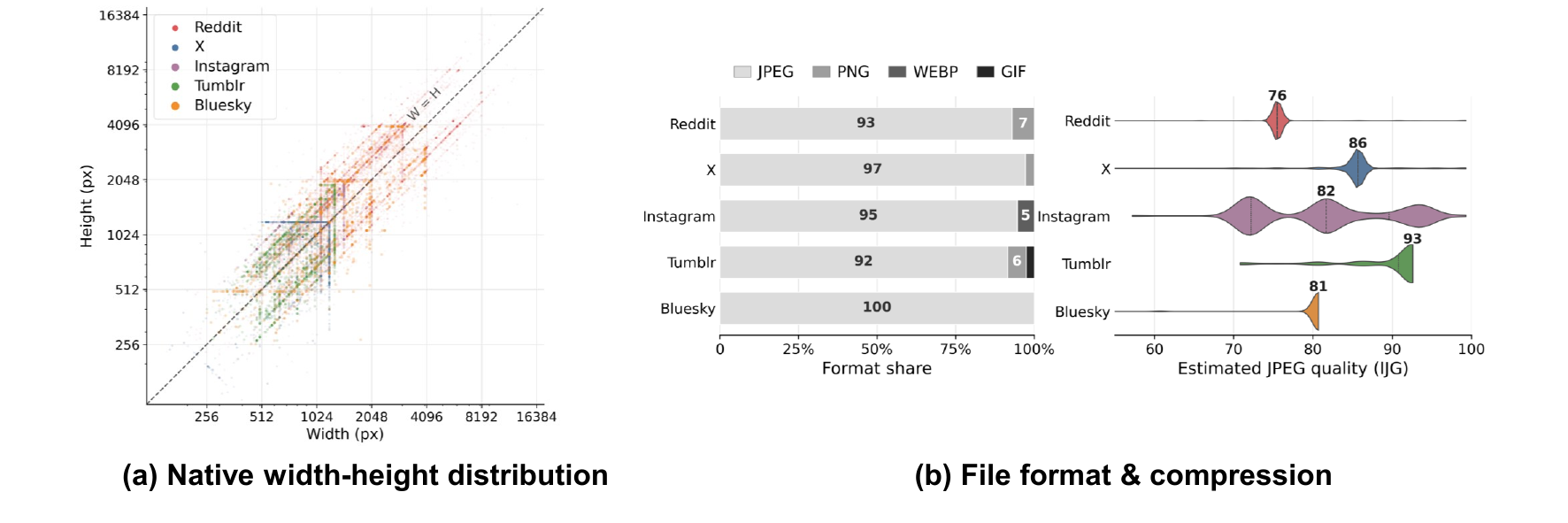}
\caption{
\textbf{Platform-native statistics of authentic images in So-Fake-OOD.}
(a) Native width--height distribution, with the dashed diagonal marking W=H; crosshairs show per-platform medians with interquartile ranges.
(b) File-format shares and estimated JPEG-quality distributions; numbers give per-platform median quality.
}
\label{sup:figure2}

\end{figure*}

\begin{figure*}[t]
\centering
\includegraphics[width=0.9\textwidth]{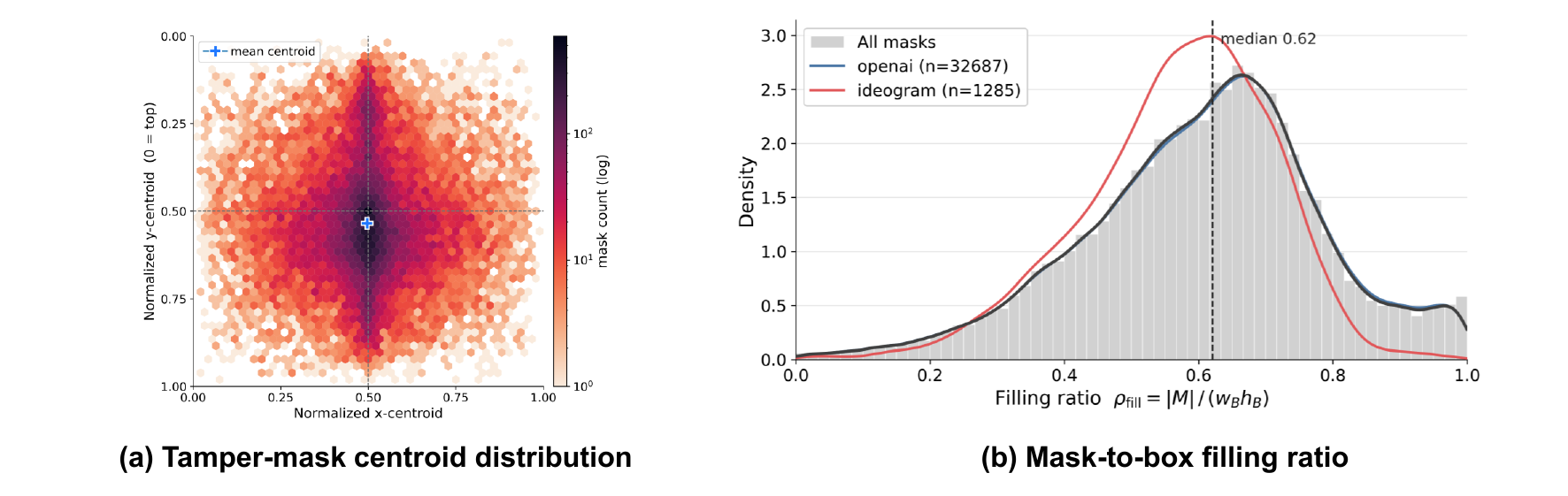}
\caption{
\textbf{Geometric statistics of tampered regions in So-Fake-OOD.}
(a) Tamper-mask centroid distribution in normalized image coordinates, where the y-axis follows image coordinates.
(b) Mask-to-box filling ratio $\rho_{\mathrm{fill}} = |M|/(w_B h_B)$, where lower values indicate sparser or more irregular masks and higher values indicate more compact regions.
}
\label{sup:figure3}
\end{figure*}

\noindent\textbf{Human Verification.}
Each generated rationale is reviewed by five trained annotators, who check whether the stated cues are visible in the image, consistent with the annotation context, and clearly expressed according to a shared annotation guideline.
Cues supported by a majority of annotators are kept, whereas unsupported, ambiguous, or annotation-inconsistent statements are revised or removed.
Rationales that cannot be reliably corrected are discarded.

\section{Benchmark Characteristics and Diagnostics}
\label{suppl:dataset-characteristics}

In this section, we provide additional benchmark-level diagnostics of So-Fake, with a focus on So-Fake-OOD.
We first characterize the platform-native statistics of authentic OOD images, including resolution, file format, and compression-related properties across different social-media platforms (Sec.~\ref{suppl:ood-real-statistics}).
We then examine the geometry of OOD tampered regions, going beyond
mask-area ratios to analyze their spatial locations and mask-to-box
occupancy patterns (Sec.~\ref{suppl:ood-tamper-geometry}).
Finally, we examine potential shortcut and confound factors,
including format and resolution normalization, editing-model
identity, platform--label association, and frequency-domain
statistics, to verify that such cues do not explain the benchmark
results (Sec.~\ref{suppl:low-level-statistics}).

\subsection{Platform-Native Statistics of Authentic OOD Images}
\label{suppl:ood-real-statistics}

We first analyze the platform-native statistics of authentic images in So-Fake-OOD.
The analysis uses the retained authentic OOD images before any model-specific resizing or preprocessing, and characterizes their native width–height distribution, file formats, and JPEG compression-related properties.

Fig.~\ref{sup:figure2}(a) visualizes the native width--height distribution of authentic OOD images.
The samples are not constrained to a fixed square resolution, but instead cover a wide range of image sizes and layouts.
In particular, Reddit and Bluesky contain images with broader resolution ranges, whereas X, Instagram, and Tumblr show more concentrated width–height distributions.

Fig.~\ref{sup:figure2}(b)  reports file-format shares and estimated JPEG quality. 
JPEG dominates across all platforms, but the estimated JPEG quality differs noticeably across sources.

Overall, these platform-level statistics complement the aggregate real-image (OOD-R) statistics in the main paper and show that So-Fake-OOD preserves heterogeneous native image conditions from real social-media sources.

\subsection{Geometry of OOD Tampered Regions}
\label{suppl:ood-tamper-geometry}

We next analyze the geometric properties of localized manipulations in So-Fake-OOD.
The main paper reports the mask-area ratio and groups tampered regions into small, medium, and large tiers.
However, mask area alone does not fully characterize localization difficulty: two manipulated regions with similar area may differ substantially in spatial location, enclosing-box geometry, and mask-to-box occupancy.
Therefore, we provide additional diagnostics based on the tamper masks and bounding boxes.

Fig.~\ref{sup:figure3}(a) shows the spatial distribution of tamper-mask centroids in normalized image coordinates.
The distribution is centered around the image middle, which is consistent with object-centric localized social-media edits, while still covering substantial off-center manipulations.
This indicates that OOD tampering is not restricted to a single fixed image location.

Fig.~\ref{sup:figure3}(b) further measures the mask-to-box filling ratio, defined as $\rho_{\mathrm{fill}} = |M|/(w_B h_B)$.
This statistic quantifies how densely the manipulated pixels occupy their enclosing bounding box.
Low filling ratios correspond to sparse, elongated, fragmented, or perforated masks, whereas values close to one correspond to compact regions that nearly fill their bounding boxes.
The broad distribution shows that OOD-Tampered images contain both compact and irregular localized manipulations. Therefore, box-level and mask-level localization pose different levels of difficulty.

Together, these diagnostics complement the area-based tamper statistics in the main paper and provide a more complete view of the spatial localization challenge posed by So-Fake-OOD.

\subsection{Shortcut and Confound Diagnostics}
\label{suppl:low-level-statistics}
We further examine potential shortcut factors in So-Fake-OOD from four perspectives: format and resolution normalization, editing-model identity, platform--label association, and frequency-domain
statistics.
These diagnostics are intended to quantify whether acquisition- or
construction-related cues can explain the benchmark results, rather
than to remove the platform-native characteristics of the data.

\begin{table}[t]
\centering
\small
\setlength{\tabcolsep}{6pt}
\caption{\textbf{Format- and resolution-normalization stress test on So-Fake-OOD.}}
\label{sup:table11}
\resizebox{\columnwidth}{!}{%
\begin{tabular}{@{}llccc@{}}
\toprule
\cellcolor{lightgray!30}Method &\cellcolor{lightgray!30} Type & \cellcolor{lightgray!30}Original & \cellcolor{lightgray!30}Normalized & \cellcolor{lightgray!30}Drop $\downarrow$ \\
\midrule
UnivFD~\cite{DBLP:conf/cvpr/OjhaLL23}        & Detection & 48.6 & 44.1 & 4.5 \\
EFFORT~\cite{DBLP:conf/icml/0002WJZLCYDW025} & Detection & 65.5 & 62.2 & 3.3 \\
SIDA~\cite{DBLP:journals/corr/abs-2412-04292}          & IFDL & 65.6 & 62.4 & 3.2 \\
Qwen2.5-VL-7B~\cite{DBLP:journals/corr/abs-2502-13923} & VLM  & 65.3 & 61.8 & 3.5 \\
\textbf{So-Fake-R1} & VLM & \textbf{72.0} & \textbf{70.3} & \textbf{1.7} \\
\bottomrule
\end{tabular}}
\end{table}

\noindent\textbf{Format and Resolution Normalization.}
Real and forged images may differ in native file format and resolution because they originate from different acquisition, generation, and editing pipelines (Tab.~III(a) of the main paper).
In particular, authentic OOD images are predominantly JPEG, whereas generated FULL\_SYNTHETIC and TAMPERED samples are mostly PNG.
A detector could in principle exploit such storage-pipeline cues rather than genuine forgery evidence. 
To quantify the extent to which such cues explain detection performance on So-Fake-OOD, we construct a \emph{format- and resolution-normalized} copy of the benchmark.
Every image, regardless of authenticity scope, is processed by an identical pipeline: it is decoded to RGB, resized with bicubic interpolation to a long side of $1024$ pixels, and re-encoded as JPEG at quality $85$. The JPEG quality is chosen to be
close to the median estimated quality of authentic OOD images
(Fig.~\ref{sup:figure2}(b)).
This normalization removes all class-correlated format information and unifies the resolution statistics across the three scopes at evaluation time.

We apply this stress test to representative methods from each family in the main comparison:
UnivFD~\cite{DBLP:conf/cvpr/OjhaLL23} and EFFORT~\cite{DBLP:conf/icml/0002WJZLCYDW025} (detection-only),
SIDA~\cite{DBLP:journals/corr/abs-2412-04292} (IFDL),
Qwen2.5-VL-7B~\cite{DBLP:journals/corr/abs-2502-13923}
(VLM), and So-Fake-R1. 
All models are the same fine-tuned checkpoints used in the main OOD comparison; only the evaluation images are normalized.
We report image-level balanced accuracy, since mask-level
localization is not comparable after resizing.

As shown in Tab.~\ref{sup:table11}, normalization leads to only a
moderate drop for all evaluated methods, and their relative ranking is preserved.
Thus, file format and resolution carry some label-correlated
information, but no method relies on them as a primary shortcut. 
This is further corroborated by the propagation-chain study in Tab.~X of the main paper, where all images are jointly resized and re-encoded and the method ordering likewise remains stable.
Since normalization is applied only at evaluation time, accuracy that is stable under normalization indicates a decision function that does not depend on these cues.

\begin{table}[t]
\centering
\small
\setlength{\tabcolsep}{4pt}
\caption{\textbf{Per-editing model diagnostics on So-Fake-OOD.} Left:
scope discrimination accuracy within
each editing-model. Right: per-editing model TAMPERED detection accuracy and
localization IoU of So-Fake-R1.}
\label{sup:table12}
\begin{tabular}{@{}lccc@{}}
\toprule
\cellcolor{lightgray!30} Editing-Model & \cellcolor{lightgray!30} Scope Acc. (F vs. T) & \cellcolor{lightgray!30}TAMP. Acc. &\cellcolor{lightgray!30}IoU \\
\midrule
GPT-4o       & 88.5 & 51.6 & 48.3 \\
Ideogram 3.0 & 86.2 & 48.9 & 45.7 \\
\bottomrule
\end{tabular}
\end{table}

\noindent\textbf{Editing-Model Identity Analysis for TAMPERED Samples.}
The TAMPERED samples in So-Fake-OOD are produced by two held-out
commercial editing models with an imbalanced sample ratio
(Fig.~\ref{sup:figure3}(b)).
A potential concern is that a detector could first recognize the
editing-model identity and then map it to the TAMPERED scope, rather than detecting the manipulation itself.

Two observations argue against this shortcut.
First, both OOD editing models also produce FULL\_SYNTHETIC samples, so model identity alone cannot determine the scope label.
We therefore evaluate scope discrimination within each editing model:
as shown in Tab.~\ref{sup:table12}, So-Fake-R1 separates
FULL\_SYNTHETIC from TAMPERED samples of the same editing model with high accuracy, which is incompatible with an identity-to-scope mapping.
Second, per-model TAMPERED detection and localization results in
Tab.~\ref{sup:table12} are comparable across the two editing models, indicating that the aggregate TAMPERED results are not driven by the
fingerprint of the dominant model.
Finally, the OOD editing models are disjoint from those used to
construct So-Fake-Set, so memorized training-editor fingerprints cannot transfer to So-Fake-OOD by construction.

\begin{table}[t]
\centering
\scriptsize
\setlength{\tabcolsep}{1.5pt}
\caption{\textbf{Per-source-platform detection results of
So-Fake-R1 on So-Fake-OOD.} REAL recall is reported per source platform; FULL and
TAMP.\ are reported for the platforms whose images serve as
generation sources. X and Instagram contribute REAL samples only.}
\label{sup:table15}
\begin{tabular}{lccccc}
\toprule
\cellcolor{lightgray!30}Platform & \cellcolor{lightgray!30}\#REAL & \cellcolor{lightgray!30}REAL &\cellcolor{lightgray!30} FULL & \cellcolor{lightgray!30}TAMP. &\cellcolor{lightgray!30} Bal. Acc. \\
\midrule
Reddit    & 20K   & 93.6 & 71.5 & 51.2 & 72.1 \\
Bluesky   & 1.5K  & 94.9 & 70.4 & 49.6 & 71.6 \\
Tumblr    & 2K    & 94.3 & 70.8 & 50.1 & 71.7 \\
X         & 7K    & 94.7 & --   & --   & --   \\
Instagram & 3K    & 94.4 & --   & --   & --   \\
\midrule
Common-platform subset
           & 23.5K & 93.7 & 71.2 & 50.8 & 71.9 \\
All        & 33.5K & 94.0 & 71.2 & 50.8 & 72.0 \\
\bottomrule
\end{tabular}
\end{table}
\noindent\textbf{Platform--Label Association.}
To avoid constructing redistributed derivatives from platforms whose terms restrict content redistribution, X and Instagram contribute REAL samples only, whereas forged samples are constructed from Reddit, Bluesky, and Tumblr sources.
This creates a partial association between source platform and the
REAL label, which a detector could in principle exploit through
platform-specific layout or content characteristics.
To quantify its effect, we additionally evaluate on a source-matched subset restricted to Reddit, Bluesky, and Tumblr, for which all three authenticity scopes are represented.

As shown in Tab.~\ref{sup:table15}, So-Fake-R1 obtains 71.9 balanced accuracy on this subset, compared with 72.0 on the full So-Fake-OOD benchmark.
Its per-platform balanced accuracy is also stable across the three
source-matched platforms.
These results indicate that the REAL-only X and Instagram samples do not materially inflate the aggregate benchmark result, although
platform characteristics may still provide weak auxiliary cues.

\begin{figure*}[t]
\centering
\includegraphics[width=\textwidth]{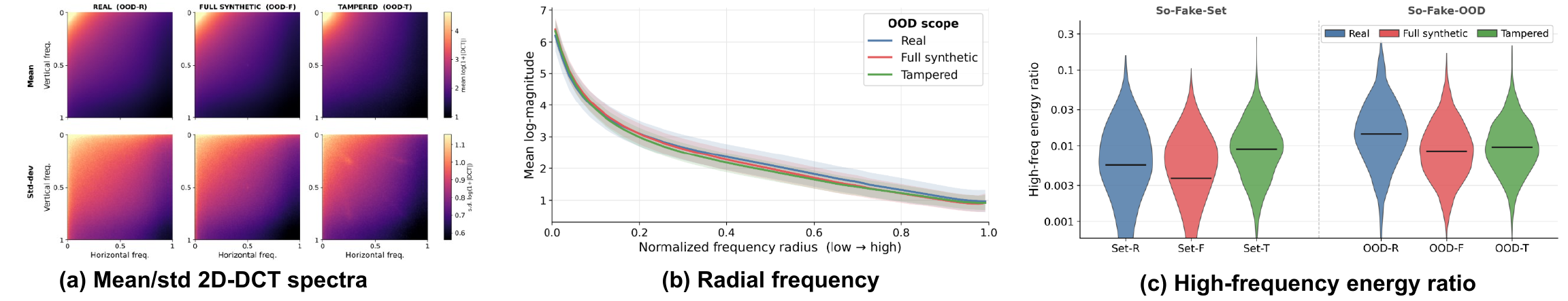}
\caption{
\textbf{Low-level forensic statistics of So-Fake.}
(a) Mean and per-pixel standard-deviation log-magnitude 2D-DCT spectra of OOD images under the REAL, FULL\_SYNTHETIC, and TAMPERED scopes.
(b) Azimuthally averaged radial frequency profiles of OOD images, where shaded regions denote $\pm 1$ standard deviation.
(c) High-frequency energy ratio $E_{\mathrm{high}}$ across the six split-scope groups, computed as the share of AC  (non-DC) DCT energy at normalized radius larger than 0.5.
These diagnostics visualize frequency-domain differences across authenticity scopes and benchmark splits, and help assess potential low-level shortcut factors.
}
\label{sup:figure4}
\end{figure*}

\noindent\textbf{Frequency-domain Statistics.}
We next examine low-level forensic statistics from a frequency-domain perspective.
We convert each image to its luminance channel, resize it to $256 \times 256$, subtract the per-image mean, and compute the log-magnitude spectrum of the 2D discrete cosine transform (DCT).
We then aggregate the resulting spectra by authenticity label and benchmark split.

Fig.~\ref{sup:figure4}(a) shows the mean and per-frequency-bin standard-deviation spectra of OOD images under three scopes.
All three scopes share a similar coarse spectral envelope dominated by low-frequency energy.
The standard-deviation maps also show broadly similar variability patterns, with only moderate visual differences between real, full synthetic, and tampered images.

Fig.~\ref{sup:figure4}(b) summarizes the same frequency statistics using averaged radial profiles.
The horizontal axis denotes the normalized frequency radius, ranging from low to high frequencies.
The three curves largely overlap across the spectrum, although REAL images retain slightly more energy on average in the high-frequency tail than FULL\_SYNTHETIC images.
The substantial overlap of the $\pm 1$ standard-deviation bands suggests that frequency magnitude alone does not provide an obvious visual separation among the three OOD scopes.

Fig.~\ref{sup:figure4}(c) further reports the high-frequency energy ratio $E_{\mathrm{high}}$ across So-Fake-Set and So-Fake-OOD.
We define $E_{\mathrm{high}}$ as the proportion of non-DC raw DCT energy whose normalized frequency radius exceeds 0.5:
\begin{equation}
E_{\mathrm{high}} =
\frac{
\sum_{r(u,v) > 0.5} |D(u,v)|^2
}{
\sum_{r(u,v) > 0} |D(u,v)|^2
},
\end{equation}
where $D(u,v)$ denotes the raw DCT coefficient and $r(u,v)$ denotes the normalized frequency radius.
REAL images tend to carry slightly higher high-frequency energy on average than FULL\_SYNTHETIC images, especially in So-Fake-OOD, but the distributions overlap heavily across scopes.
This suggests that low-level frequency statistics may act as weak spurious cues, but the visualized statistics do not reveal a simple frequency-based shortcut that cleanly separates real from forged images.

Overall, these diagnostics show that low-level acquisition, storage, and frequency cues can contribute to detection, but they do not provide a sufficient shortcut for solving So-Fake-OOD.
The format/resolution normalization test shows only a moderate drop, and the DCT statistics show substantial overlap across scopes.
Reliable detection therefore still requires semantic and forensic evidence beyond simple metadata- or spectrum-level statistics.

\section{Method and Training Details}
\label{suppl:method-details}

In this section, we provide additional  training details for So-Fake-R1, complementing the framework description in the main paper.
We first report the training data construction for the evidence-producing modules and the Decision Maker, including Visual Observer SFT, Forensic Evidence Provider training, routed Decision Maker SFT, and GRPO data construction (Sec.~\ref{suppl:training-data-optimization}).
Next, we specify how the deterministic evidence gate routes each case and how outputs are parsed and finalized during inference (Sec.~\ref{suppl:evidence-routing}).
We then describe the adaptation and training configurations of all baseline methods (Sec.~\ref{suppl:baseline-config}).
Finally, we report the hyperparameters, preprocessing settings, compute resources, random seeds, and other reproducibility details needed to reproduce our experiments (Sec.~\ref{suppl:reproducibility}).

\subsection{Training Data Construction}
\label{suppl:training-data-optimization}

This section reports the data used to train each module. All samples are drawn from the So-Fake-Set training split unless otherwise stated.

\noindent\textbf{Visual Observer Data.}
The Visual Observer is trained on 100{,}000 image--response pairs from So-Fake-Set.
We construct this set by stratified proportional sampling.
We first set approximately balanced target counts for the three scope labels:
33{,}334 REAL, 33{,}333 FULL\_SYNTHETIC, and 33{,}333 TAMPERED.
Within each fake class, samples are further drawn according to the generator proportions in the So-Fake-Set training split, so that the balanced subset does not over-represent a particular generator.
For REAL images, we preserve the source/category composition of the training split.

\noindent\textbf{Forensic Evidence Provider Data.}
The provider is trained on the same sampled image pool as the Visual Observer, rather than on an independently drawn subset.
Therefore, it follows the same three-way label balance and the same generator-proportional sampling within FULL\_SYNTHETIC and TAMPERED images.
The difference lies in the supervision used for each sample: the provider uses cached frozen features, image-level labels, and patch-level spatial targets derived from tamper masks when available.

\noindent\textbf{Decision Maker SFT.}
To construct Decision Maker SFT data, we first run the trained observer and provider on the training split and cache their serialized records, and then apply the evidence gate to each record pair.
Only routed cases are kept: \textsc{Direct-Accept} cases are excluded because they never trigger a Decision Maker call.
The resulting SFT set contains 2{,}465 examples:
564 \textsc{Targeted-Arbitration} and 1{,}901 \textsc{Full-Arbitration}.
The final-label distribution is 1{,}109 \textsc{Real}, 1{,}005 \textsc{Full\_Synthetic}, and 351 \textsc{Tampered}.

\noindent\textbf{Decision Maker GRPO.}
The GRPO set contains 5{,}000 routed hard cases.
It includes the 2{,}465 SFT routed cases and additional mined hard cases from the same cached observer--provider outputs.
The route distribution is 894 \textsc{Targeted-Arbitration} and 4{,}106 \textsc{Full-Arbitration}.
The final-label distribution is 1{,}738 \textsc{Real}, 1{,}606 \textsc{Full\_Synthetic}, and 1{,}656 \textsc{Tampered}.
All \textsc{Tampered} GRPO examples have valid target boxes.
The selected hard cases cover four main patterns:
source conflict, fake-scope conflict, unsafe agreement, and both-source failure.

\subsection{Evidence Routing and Inference Finalization}
\label{suppl:evidence-routing}
At inference time, So-Fake-R1 does not invoke the Decision Maker on every image.
Instead, the two cached evidence records $\mathcal{O}$ and $\mathcal{P}$ are first passed through the deterministic evidence gate $G$, which reads only the serialized records.

\noindent\textbf{Support Levels and Sufficiency.}
The gate uses a deterministic support predicate rather than raw model confidence alone.
For the Visual Observer, the support level is read from the serialized record or inferred via a fixed keyword-matching rule over label-aligned, image-checkable phrases.
For the Forensic Evidence Provider, the support level combines a
confidence level derived from the classifier logit margin with a
rule-based descriptor-strength assessment specific to the hypothesized scope.
A record is \emph{sufficiently supported} if its support level meets the per-source acceptance bar, and a \textsc{Tampered} claim
additionally requires a valid bounding box.
Records with insufficient support, unparseable labels, or a
\textsc{Tampered} claim without a usable box are treated as unsafe for direct acceptance; the complete rules, keyword lexicon, and thresholds are released with our code.

\begin{algorithm}[t]
\caption{Evidence Gate $G(\mathcal{O},\mathcal{P})$}
\label{supalg:evidence-gate}
\begin{algorithmic}[1]
\Require observer record $\mathcal{O}$, provider record $\mathcal{P}$
\State $u_\mathcal{O} \gets \mathrm{Supported}(\mathcal{O})$;\quad
       $u_\mathcal{P} \gets \mathrm{Supported}(\mathcal{P})$
       \Comment{support predicate defined above}
\If{$\mathrm{invalid}(\mathcal{O}) \vee \mathrm{invalid}(\mathcal{P})$}
    \State \Return \textsc{Full-Arbitration}
\EndIf
\If{$\hat{s}^{\mathrm{obs}} = \hat{s}^{\mathrm{for}}$}
    \Comment{same-scope agreement}
    \If{$\hat{s}^{\mathrm{obs}} = \textsc{Tampered} \wedge
         \neg\,\mathrm{validBox}(\hat{B}^{\mathrm{obs}}) \wedge
         \neg\,\mathrm{validBox}(\hat{B}^{\mathrm{for}})$}
        \State \Return \textsc{Full-Arbitration}
        \Comment{no usable tamper box}
    \EndIf
    \If{$u_\mathcal{O} \wedge u_\mathcal{P}$}
        \State \Return \textsc{Direct-Accept}
        \Comment{clean agreement}
    \ElsIf{$u_\mathcal{O} \vee u_\mathcal{P}$}
        \State \Return \textsc{Direct-Accept}
        \Comment{flagged as cautious acceptance}
    \Else
        \State \Return \textsc{Full-Arbitration}
        \Comment{agreement with weak/mixed support}
    \EndIf
\ElsIf{$\{\hat{s}^{\mathrm{obs}}, \hat{s}^{\mathrm{for}}\} =
        \{\textsc{Real}, \textsc{Full\_Synthetic}\}$}
    \State \Return \textsc{Targeted-Arbitration}
    \Comment{narrow authenticity dispute}
\Else
    \State \Return \textsc{Full-Arbitration}
    \Comment{disagreement involving \textsc{Tampered}}
\EndIf
\end{algorithmic}
\end{algorithm}

\noindent\textbf{Evidence Gate.}
The gate assigns each case one of three inference routes,
$\rho = G(\mathcal{O},\mathcal{P})$, following the
procedure in Alg.~\ref{supalg:evidence-gate}: clean, well-supported
same-scope agreement is accepted directly, a narrow
\textsc{Real}-versus-\textsc{Full\_Synthetic} dispute is routed to
targeted arbitration, and all remaining cases, including any
disagreement involving \textsc{Tampered} and same-scope cases with
insufficient support, are routed to full arbitration.
On the So-Fake-OOD evaluation split, \textsc{Direct-Accept},
\textsc{Targeted-Arbitration}, and \textsc{Full-Arbitration} account
for 75.2\%, 3.6\%, and 21.2\% of cases, respectively.

\noindent\textbf{Direct Acceptance.}
For these cases, no image-conditioned Decision Maker call is made.
The final scope is copied from the agreed evidence label, and the final authenticity is derived accordingly.
If the accepted scope is \textsc{Tampered}, the final bounding box is taken from the provider if its box is valid, and otherwise from the observer; the gate guarantees that at least one valid box exists for directly accepted \textsc{Tampered} cases.
Non-\textsc{Tampered} outputs use \textsc{None} as the bounding box.

\noindent\textbf{Routed Arbitration.}
For routed cases, the Decision Maker receives the original image together with the two evidence records and the assigned route.
The model-visible prompt states the route, but reveals neither the gate's internal routing criteria nor the ground-truth label.
Under \textsc{Targeted-Arbitration}, the prompt restricts the task to a binary decision between \textsc{Real} and \textsc{Full\_Synthetic}; \textsc{Tampered} is not allowed.
Under \textsc{Full-Arbitration}, the Decision Maker may output any scope in $\mathcal{Y}$, and a bounding box is required only when the final scope is \textsc{Tampered}.

\noindent\textbf{Decision Maker Output.}
The Decision Maker response contains five structured fields:
\texttt{authenticity\_\allowbreak decision},
\texttt{scope\_\allowbreak decision},
\texttt{routing\_\allowbreak decision},
\texttt{final\_\allowbreak assessment}, and
\texttt{final\_\allowbreak answer}, among which
\texttt{final\_answer} serves as the primary prediction.
Given a parseable final answer, the remaining fields are normalized to
satisfy the label consistency defined by the scope-to-authenticity
mapping: a \textsc{Real} answer implies authenticity \textsc{Real} and
no bounding box, a \textsc{Full\_Synthetic} answer implies authenticity
\textsc{Fake} and no bounding box, and a \textsc{Tampered} answer
implies authenticity \textsc{Fake} together with a valid bounding box.
Responses whose final answer cannot be parsed, or whose required
bounding box is missing, are counted as invalid predictions during
evaluation.
Finally, the direct-acceptance outputs and the routed Decision Maker
outputs are merged into a single prediction set for evaluation.

\begin{figure}[t]
\centering
\begin{tcolorbox}[title=Visual Observer Prompt,
fontupper=\ttfamily\scriptsize, boxsep=2pt, left=3pt, right=3pt]
<image>\\
Please carefully describe the content of this image and classify its
\textbf{REAL}, \textbf{TAMPERED}, or \textbf{FULL\_SYNTHETIC}.\\[2pt]
\textbf{DEFINITIONS:}\\
- \textbf{REAL}: The image is authentic with no AI generation or
manipulation.\\
- \textbf{TAMPERED}: Parts of the image have been manipulated,
edited, or altered.\\
- \textbf{FULL\_SYNTHETIC}: The entire image was generated by AI.\\[2pt]
Output the thinking process in \textbf{<think> </think>} and final
answer in \textbf{<answer> </answer>} tags. Additionally, if the
image is TAMPERED, output the one bbox to depict the tampered area.\\
i.e., <think> thinking process here </think>\\
<answer>\{Answer\}\}</answer>
\end{tcolorbox}
\caption{\textbf{Prompt template of the Visual Observer, used for both SFT and inference.} The SFT target consists of the corresponding \texttt{<think>} and \texttt{<answer>} contents, plus a bounding box for \textsc{Tampered} samples.}
\label{sup:prompt_obs}
\end{figure}

\begin{figure*}[t]
\centering
\begin{tcolorbox}[title=Decision Maker Prompt
(\texttt{FULL\_RECHECK}),
fontupper=\ttfamily\scriptsize, boxsep=2pt, left=3pt, right=3pt]
<image>\\
You are a forensic decision maker. You are given one image and two
evidence reports. Your task is to follow the recheck mode, compare
the two reports against the image, and produce the final authenticity
label. Do not perform unrelated open-ended detection.\\[2pt]
\textbf{Class definitions:}\\
- \textbf{REAL}: authentic image with no AI generation or
manipulation.\\
- \textbf{FULL\_SYNTHETIC}: the whole image is AI-generated.\\
- \textbf{TAMPERED}: part of the image is edited or manipulated.\\[2pt]
\textbf{{[Recheck Mode]}}\\
route\_action: FULL\_RECHECK\\[2pt]
\textbf{Task rules:}\\
- The two reports cannot be accepted directly.\\
- Use the image to verify both report claims.\\
- First decide authenticity: REAL or FAKE.\\
- If authenticity is REAL, set fake\_scope to N/A.\\
- If authenticity is FAKE, decide fake\_scope as FULL\_SYNTHETIC or
TAMPERED.\\
- Output a bbox only when the final answer is TAMPERED.\\
- You may reject both reports if both fail the image-side check.\\
- Explain why direct acceptance is unsafe, which claim survives or
fails, and why the final label follows.\\[2pt]
\textbf{Required output format:}\\
\textbf{<authenticity\_decision>} authenticity: REAL|FAKE
\textbf{</authenticity\_decision>}\\
\textbf{<scope\_decision>} fake\_scope: N/A|FULL\_SYNTHETIC|TAMPERED
\textbf{</scope\_decision>}\\
\textbf{<routing\_decision>} route\_action: FULL\_RECHECK
\textbf{</routing\_decision>}\\
\textbf{<final\_assessment>} One paragraph. Explain what the
image-side check finds, which report claim survives or fails, and why
the final label follows. \textbf{</final\_assessment>}\\
\textbf{<final\_answer>}REAL|FULL\_SYNTHETIC|TAMPERED[,
<|box\_start|>(x1,y1),(x2,y2)<|box\_end|>]\textbf{</final\_answer>}\\[2pt]
Here are the two evidence reports:\\[2pt]
\textbf{{[Visual Observer Report]}}\\
observer\_answer: \{observer\_answer\}\\
observer\_bbox: \{observer\_bbox\}\\
observer\_report: \{observer\_report\}\\[2pt]
\textbf{{[Forensic Evidence Provider Report]}}\\
evidence\_provider\_answer: \{provider\_answer\}\\
evidence\_provider\_bbox: \{provider\_bbox\}\\
evidence\_provider\_assessment: \{provider\_assessment\}\\[2pt]
Now produce the final decision.
\end{tcolorbox}
\caption{\textbf{Prompt template of the Decision Maker under
\texttt{FULL\_RECHECK} (\textsc{Full-Arbitration}).} The bracketed
placeholders are filled with the serialized evidence records and the
route assigned by the gate. Boldface is added for readability and is
not part of the prompt string.}
\label{sup:prompt_dm}
\end{figure*}

\begin{figure*}[t]
\centering
\begin{tcolorbox}[title=Decision Maker Prompt
(\texttt{TARGETED\_RECHECK}),
fontupper=\ttfamily\scriptsize, boxsep=2pt, left=3pt, right=3pt]
<image>\\
You are a forensic decision maker. You are given one image and two
evidence reports. Your task is to follow the recheck mode, compare
the two reports against the image, and produce the final authenticity
label. Do not perform unrelated open-ended detection.\\[2pt]
\textbf{Class definitions:}\\
- \textbf{REAL}: authentic image with no AI generation or
manipulation.\\
- \textbf{FULL\_SYNTHETIC}: the whole image is AI-generated.\\[2pt]
\textbf{{[Recheck Mode]}}\\
route\_action: TARGETED\_RECHECK\\[2pt]
\textbf{Task rules:}\\
- This is a narrow REAL-vs-FULL\_SYNTHETIC dispute.\\
- First decide authenticity: REAL or FAKE.\\
- If authenticity is REAL, set fake\_scope to N/A and final\_answer
to REAL.\\
- If authenticity is FAKE, set fake\_scope to FULL\_SYNTHETIC and
final\_answer to FULL\_SYNTHETIC.\\
- Use the image to verify which of the two report claims survives.\\
- Explain what the rejected claim gets wrong.\\[2pt]
\textbf{Required output format:}\\
\textbf{<authenticity\_decision>}\\
authenticity: REAL|FAKE\\
\textbf{</authenticity\_decision>}\\[2pt]
\textbf{<scope\_decision>}\\
fake\_scope: N/A|FULL\_SYNTHETIC\\
\textbf{</scope\_decision>}\\[2pt]
\textbf{<routing\_decision>}\\
route\_action: TARGETED\_RECHECK\\
\textbf{</routing\_decision>}\\[2pt]
\textbf{<final\_assessment>}\\
One paragraph. Decide between the REAL claim and the FULL\_SYNTHETIC
claim, explain which survives, and state what the rejected claim gets
wrong.\\
\textbf{</final\_assessment>}\\[2pt]
\textbf{<final\_answer>}REAL|FULL\_SYNTHETIC\textbf{</final\_answer>}\\[2pt]
Here are the two evidence reports:\\[2pt]
\textbf{{[Visual Observer Report]}}\\
observer\_answer: \{observer\_answer\}\\
observer\_bbox: \{observer\_bbox\}\\
observer\_report: \{observer\_report\}\\[2pt]
\textbf{{[Forensic Evidence Provider Report]}}\\
evidence\_provider\_answer: \{provider\_answer\}\\
evidence\_provider\_bbox: \{provider\_bbox\}\\
evidence\_provider\_assessment: \{provider\_assessment\}\\[2pt]
Now produce the final decision.
\end{tcolorbox}
\caption{\textbf{Prompt template of the Decision Maker under
\texttt{TARGETED\_RECHECK} (\textsc{Targeted-Arbitration}).} Boldface
is added for readability and is not part of the prompt string.}
\label{sup:prompt_dm_targeted}
\end{figure*}

\noindent\textbf{Prompt Templates.}
Figs.~\ref{sup:prompt_obs},~\ref{sup:prompt_dm}, and~\ref{sup:prompt_dm_targeted} list the prompt templates used by the Visual Observer and the Decision Maker.
The same templates are used for SFT, GRPO, and inference.
The templates are reproduced verbatim from our implementation, where \textsc{Targeted-Arbitration} and \textsc{Full-Arbitration} appear as \texttt{TARGETED\_RECHECK} and \texttt{FULL\_RECHECK}, and REAL cases use \texttt{fake\_scope: N/A} in the raw output format, which is normalized to scope \textsc{Real} during parsing.
The observer's \texttt{<think>} and \texttt{<answer>} outputs are
serialized into the \texttt{observer\_report} and
\texttt{observer\_scope} fields,
with the bounding box parsed into \texttt{observer\_bbox}.

\begin{table}[t]
\centering
\scriptsize
\setlength{\tabcolsep}{2pt}
\caption{\textbf{Per-baseline training configurations on So-Fake-Set.}}
\label{sup:table14}
\begin{adjustbox}{width=1.0\columnwidth}
\begin{tabular}{lccc}
\toprule
\cellcolor{lightgray!30} Method &
\cellcolor{lightgray!30} New head &
\cellcolor{lightgray!30} Trainable components &
\cellcolor{lightgray!30} Training data \\
\midrule
CnnSpot~\cite{wang2019cnngenerated}  & Linear(2048,\,3) & full model & full split \\
UnivFD~\cite{DBLP:conf/cvpr/OjhaLL23}   & Linear(768,\,3)  & linear head  & full split \\
C2P-CLIP~\cite{DBLP:conf/aaai/TanTLGWZW25} & Linear(768,\,3)  & FC head & full split \\
DDA~\cite{DBLP:journals/corr/abs-2505-14359}      & Linear(1024,\,3) & LoRA + FC head & full split \\
EFFORT~\cite{DBLP:conf/icml/0002WJZLCYDW025}   & Linear(1024,\,3) & SVD residuals + head & full split \\
FatFormer~\cite{DBLP:conf/cvpr/LiuTTW0Z24} & 3-class prompts & FAA + LGA & full split \\
FreAware~\cite{DBLP:conf/aaai/Tan0WGLW24} & Linear(512,\,3) & full model & full split \\
\midrule
TruFor~\cite{DBLP:conf/cvpr/GuillaroCSDV23}   & Linear(128,\,3) & original protocol & full split \\
PSCC-Net~\cite{DBLP:journals/tcsv/LiuLCL22}  & Linear(16,\,3) & original protocol & full split \\
SIDA~\cite{DBLP:journals/corr/abs-2412-04292} & text target & LoRA & 100K subset \\
\midrule
VLM-based (LLaVA~\cite{DBLP:conf/nips/LiuLWL23a}, \emph{etc.}) & text target & LoRA & 100K subset \\
\bottomrule
\end{tabular}
\end{adjustbox}
\end{table}

\subsection{Baseline Adaptation and Training Configurations}
\label{suppl:baseline-config}
This section details how each baseline is adapted to the three-way
protocol of So-Fake, with the complete per-baseline configurations
summarized in Tab.~\ref{sup:table14}.
All baselines are initialized from their official checkpoints and
trained with the official default hyperparameters, including
optimizer, learning rate, input resolution, and epochs.

\noindent\textbf{Detection-only Baselines.}
For binary detectors, the original real/fake classification head is
discarded and replaced with a randomly initialized three-way linear
head, trained with the standard cross-entropy loss over
\{\textsc{Real}, \textsc{Full\_Synthetic}, \textsc{Tampered}\}.
The remaining weights are loaded from the official checkpoints, and
the set of trainable components follows each method's original
protocol, e.g., full-model fine-tuning for CnnSpot~\cite{wang2019cnngenerated} and FreAware~\cite{DBLP:conf/aaai/Tan0WGLW24}, and
head-only training for CLIP-based detectors such as UnivFD~\cite{DBLP:conf/cvpr/OjhaLL23} and
C2P-CLIP~\cite{DBLP:conf/aaai/TanTLGWZW25}.
All detection-only baselines are trained on the full So-Fake-Set
training split.

\noindent\textbf{IFDL Baselines.}
For TruFor~\cite{DBLP:conf/cvpr/GuillaroCSDV23} and PSCC-Net~\cite{DBLP:journals/tcsv/LiuLCL22}, the original binary detection head is replaced with a randomly initialized three-way classification head trained with cross-entropy, while the localization branch follows the original protocol.
FULL\_SYNTHETIC masks are ignored in the localization loss, and
localization metrics are computed only on ground-truth
\textsc{Tampered} images with available masks.
SIDA~\cite{DBLP:journals/corr/abs-2412-04292} is fine-tuned on the same 100K class-balanced subset as our Visual Observer, following its official training protocol.
\noindent\textbf{VLM-based Baselines.}
VLM-based baselines are fine-tuned with LoRA on the same 100K
class-balanced subset used by our Visual Observer (Sec.~C-1), with
three-way scope labels as text targets.
This ensures a directly comparable training protocol with So-Fake-R1.

\begin{table*}[t]
\centering
\caption{\textbf{Per-stage training configuration of So-Fake-R1.}
All stages use a fixed random seed of 42 for data sampling and model training.
For VLM stages, batch size is reported as per-device batch size $\times$ gradient accumulation.}
\label{sup:table5}
\footnotesize
\setlength{\tabcolsep}{4pt}
\begin{tabular}{@{}lllll@{}}
\toprule
 \cellcolor{lightgray!30} &\cellcolor{lightgray!30}  Observer SFT & \cellcolor{lightgray!30} Provider (FETA) &\cellcolor{lightgray!30}  Decision Maker SFT & \cellcolor{lightgray!30} Decision Maker GRPO \\
\midrule
Initialization & Qwen2.5-VL-7B & frozen Qwen2.5-VL-7B & Qwen2.5-VL-7B & DM SFT checkpoint \\
Trainable components & LoRA (linear layers) & FETA adapter + heads & LoRA (linear layers) & LoRA (linear layers) \\
Input & image & layer-19 features & image + records + route & image + records + route \\
Training data & balanced 3-way labels & 3-way labels (+ masks) & 2{,}465 routed cases & 5{,}000 routed cases \\
Supervision & SFT targets & image-level + spatial & SFT targets & decision + format reward \\
LoRA rank / alpha & 16 / 32 & -- & 16 / 32 & 16 / 32 \\
Learning rate & $1{\times}10^{-4}$ & $3{\times}10^{-4}$ & $1{\times}10^{-4}$ & $3{\times}10^{-7}$ \\
Batch size / epochs & $1{\times}32$ / 1 epoch & 128 / 1 epoch & $1{\times}32$ / 1 epoch & $1{\times}4$ / 220 steps \\
Max completion length & 512 & -- & 512 & 512 \\
Sampling / augmentation & -- & blur aug. $p{=}0.5$ & -- & 8 per prompt, $T{=}1.25$, top-$p{=}1.0$ \\
\bottomrule
\end{tabular}
\end{table*}

\subsection{Reproducibility Details}
\label{suppl:reproducibility}

This section summarizes the preprocessing settings, per-stage training configuration, and compute setup used in our experiments.
Tab.~\ref{sup:table5} lists the training configuration of each stage.
Training is conducted on 2 NVIDIA RTX 5090 GPUs.

\noindent\textbf{Image Preprocessing.}
Images are resized and normalized with the standard Qwen2.5-VL-7B preprocessing pipeline for all VLM-based modules, while the provider-side feature grid is kept at the VLM patch resolution.
Predicted boxes and provider support boxes are converted to the normalized coordinate format used by the Decision Maker, \texttt{<|box\_start|>(x1,y1),(x2,y2)<|box\_end|>}, with coordinates defined on a $224\times224$ canvas.
For localization evaluation, predicted boxes are converted to masks using the same fixed SAM2~\cite{DBLP:journals/corr/abs-2408-00714} configuration for all box-output methods.

\noindent\textbf{GRPO Reward.}
The GRPO reward is defined as
\begin{equation}
\begin{split}
R ={}& \lambda_{\mathrm{auth}} R_{\mathrm{auth}}
     + \lambda_{\mathrm{scope}} R_{\mathrm{scope}}
     + \lambda_{\mathrm{arb}} R_{\mathrm{arb}} \\
    &+ \lambda_{\mathrm{loc}} R_{\mathrm{loc}}
     + \lambda_{\mathrm{fmt}} R_{\mathrm{fmt}},
\end{split}
\end{equation}
with
\begin{equation}
\lambda_{\mathrm{auth}}=\lambda_{\mathrm{scope}}=\lambda_{\mathrm{arb}}=\lambda_{\mathrm{loc}}=1.0,
\qquad
\lambda_{\mathrm{fmt}}=0.1.
\end{equation}
$R_{\mathrm{auth}}$ rewards binary \textsc{Real}/\textsc{Fake} correctness.
$R_{\mathrm{scope}}$ rewards the final three-way scope decision and penalizes confusion between \textsc{Full\_Synthetic} and \textsc{Tampered}.
$R_{\mathrm{arb}}$ rewards correct source arbitration, including selecting the correct evidence source when exactly one is correct, and rejecting both sources when neither matches the ground truth.
$R_{\mathrm{loc}}$ applies only to \textsc{Tampered} cases and uses an IoU-scaled localization score, with penalties for missing or invalid boxes.
$R_{\mathrm{fmt}}$ checks that the required fields are present and mutually consistent; its weight is kept small so that output validity is encouraged without dominating the decision reward.

\noindent\textbf{SAM2 Configuration.}
For all box-output methods, predicted boxes are converted to masks
using SAM2.1-Tiny~\cite{DBLP:journals/corr/abs-2408-00714}, with the \texttt{sam2.1\_hiera\_t.yaml}
configuration and the \texttt{sam2.1\_hiera\_tiny.pt} checkpoint.
The predicted box is used as the sole prompt, without additional point or mask prompts.
The same SAM2 configuration is used for all methods, and we do not apply any additional post-processing outside SAM2, such as largest-component filtering or morphological cleanup.

\section{Limitations}
\label{sec:limitations}

Although So-Fake broadens the scale, semantic coverage, and OOD evaluation of social-media image forgery detection, several limitations remain.

First, So-Fake is still a finite snapshot of a rapidly changing social-media environment.
So-Fake-Set is constructed mainly from public image datasets to support scalable and reproducible training, while So-Fake-OOD covers five platforms and a limited set of held-out commercial generators.
It therefore cannot exhaustively represent all platform distributions, post-processing pipelines, geographic contexts, or future synthesis and editing models.
The link-only release protocol for X and Instagram also follows platform policies but may affect long-term accessibility if original posts are removed.
In addition, the REAL label denotes images that passed multi-reviewer screening rather than provenance-level authentication, which remains infeasible for in-the-wild platform content.
Consequently, residual label noise from undetected synthetic or edited images cannot be fully excluded.

Second, the local-forgery pipeline is inherently more constrained than unconstrained real-world manipulation: most TAMPERED samples are produced through caption-guided object selection, segmentation, and inpainting.
Although So-Fake-OOD employs held-out commercial editing models, it
retains a similar upstream mask-construction process.
As a result, the benchmark may underrepresent copy--move edits,
multi-source compositing, subtle retouching, geometric deformation,
repeated editing, and manual manipulations performed by real users.

Finally, So-Fake-R1 is subject to structural limitations in its
evidence sources and routing design.
Because the Visual Observer and the Forensic Evidence Provider share
the same underlying VLM backbone, their errors may be correlated.
Moreover, while the Evidence Gate reduces unnecessary arbitration by
directly accepting sufficiently supported agreement, it may not recover cases in which both evidence sources fail confidently in the same direction.
Addressing these limitations will require broader platform and
generator coverage, more diverse manipulation processes, and evidence arbitration mechanisms that remain robust under correlated source errors.
\section{Ethics and Data Governance}
\label{sec:ethics}
This study uses publicly accessible social-media imagery and includes human annotation and preference-evaluation components.
No institutional ethics review was conducted because the human
component was limited to low-risk image annotation and comparative
assessment, no sensitive participant information was collected, and
participant responses were analyzed and reported only in aggregate.
All annotators and evaluators were informed of the study procedure,
the intended use of their responses, and the voluntary nature of
participation, and provided consent before taking part.
Authentic images in So-Fake-OOD were collected from publicly
accessible posts through official APIs or source links, subject to the platform-specific access and release constraints summarized in
Tab.~\ref{sup:table1}.
Individual consent was not obtained from persons depicted in these
public posts because the study involved no direct interaction with
account holders and used only publicly accessible content.
Reviewers excluded content containing sensitive personal information, harmful material, or elevated privacy and safety risks before release (Appendix Sec.~\ref{suppl:platform-collection}).
Images from X and Instagram are released as source links only.
Redistributed dataset artifacts are provided for non-commercial
academic research.
Individuals may request the removal of a sample and any synthetic or tampered variants derived from it by contacting the authors.
\section{Qualitative Analysis}
\label{sec:qualitative-analysis}
We further provide qualitative examples to illustrate how So-Fake-R1 handles different evidence-routing outcomes.
The selected cases cover direct evidence acceptance, targeted arbitration, full arbitration, rejection of both evidence records, and one representative failure mode.
For each case, we show the original image, the Visual Observer record, the Forensic Evidence Provider record, and the final routed decision.
These examples highlight that So-Fake-R1 does not treat the two evidence sources as votes: clean and well-supported agreement can be accepted directly, while conflicting or unsafe evidence is routed to the Decision Maker for image-conditioned arbitration.

The failure case reveals a limitation of direct evidence acceptance.
When both evidence records consistently support the same local-forgery hypothesis, the gate may accept the agreement without invoking the Decision Maker; if the shared hypothesis is wrong, the system inherits this error.
This trade-off reduces unnecessary arbitration on clean agreement cases, but limits correction when both evidence sources fail in the same direction.
Representative qualitative cases are shown in Tabs.~\ref{sup:table6}--\ref{sup:table10}.

\begin{table*}[b]
\centering
\caption{\textbf{Direct acceptance.} Both evidence records localize the same edited region, so the gate directly accepts the TAMPERED result without invoking image-side arbitration. GT is \texttt{TAMPERED}; final prediction is \texttt{TAMPERED}.}
\label{sup:table6}

\begin{tcolorbox}[exbox]
\footnotesize
\setlength{\parskip}{1.5pt}
\noindent
\begin{minipage}[t]{0.74\linewidth}
\Obs{Visual Observer} ($\mathcal{O}$)

\Obs{Reasoning:} The image depicts a vintage car parked on a street, with a woman in a dress standing beside it. The composition includes a building in the background, and the overall scene appears to be a candid black-and-white street photograph. The woman stands out because her texture and sharpness differ from the surrounding scene. Her figure appears slightly cleaner and more digitally defined than the car, road, and background, and the lighting and shadows on her body do not fully align with the rest of the photograph. These inconsistencies suggest that the woman may have been inserted or locally altered, so the image is more consistent with localized tampering than with a fully authentic photograph.

\Obs{Authenticity $\hat{a}^{\mathrm{obs}}$:} \textbf{FAKE} \qquad
\Obs{Answer $\hat{s}^{\mathrm{obs}}$:} \textbf{TAMPERED} \qquad
\Obs{Bbox $\hat{B}^{\mathrm{obs}}$:} \textbf{(57, 61, 184, 174)}

\vspace{5pt}

\Evi{Forensic Evidence Provider} ($\mathcal{P}$)

\Evi{Descriptors $\mathbf{d}^{\mathrm{for}}$:} mass $5.59$: anomaly evidence is present but not full-frame dominant; peak $0.923$: at least one region responds strongly; concentration $1.000$: the response is concentrated in high-activation patches; support area $0.065$: the candidate region is localized; entropy $2.37$: the response is spatially organized.

\Evi{Basis:} Evidence provider answer is TAMPERED. The provider gives a localized bbox and a moderate local-edit signal. Evidence provider bbox is (57, 61, 184, 174).

\Evi{Answer $\hat{s}^{\mathrm{for}}$:} \textbf{TAMPERED} \qquad
\Evi{Confidence $c^{\mathrm{for}}$:} \textbf{0.800} \qquad
\Evi{Bbox $\hat{B}^{\mathrm{for}}$:} \textbf{(57, 61, 184, 174)}
\end{minipage}
\hfill
\begin{minipage}[t]{0.20\linewidth}
\centering
\vspace{0pt}
\includegraphics[height=2.6cm]{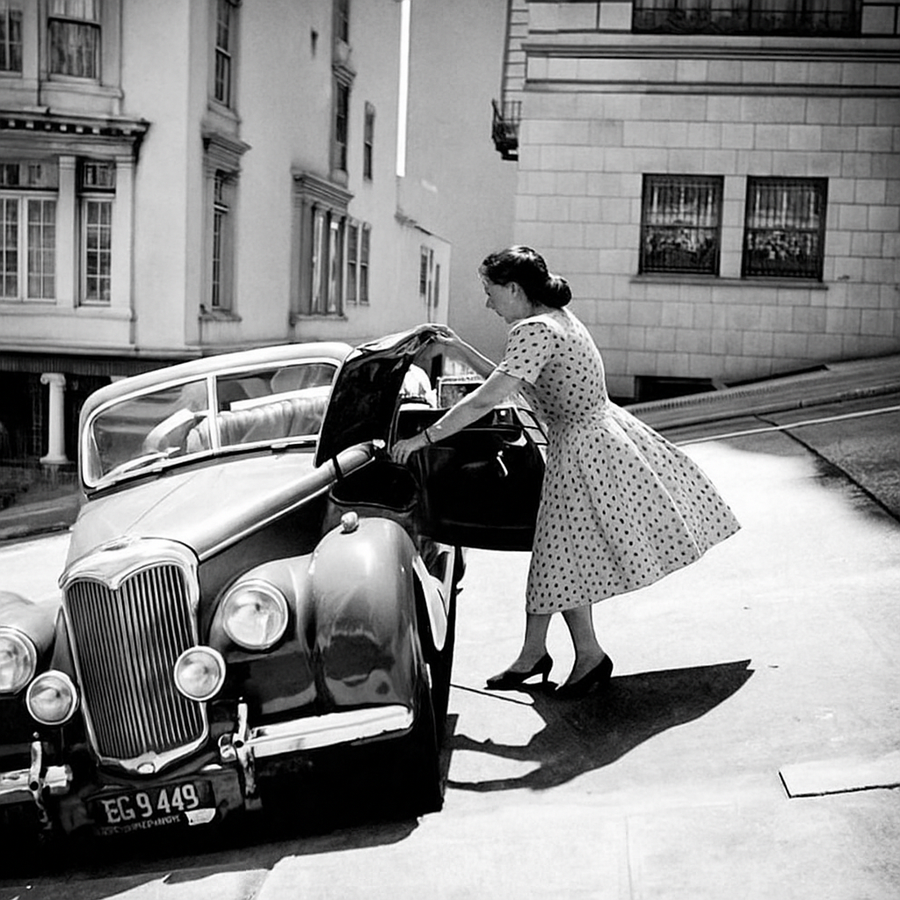}

\vspace{3pt}

\includegraphics[height=2.6cm]{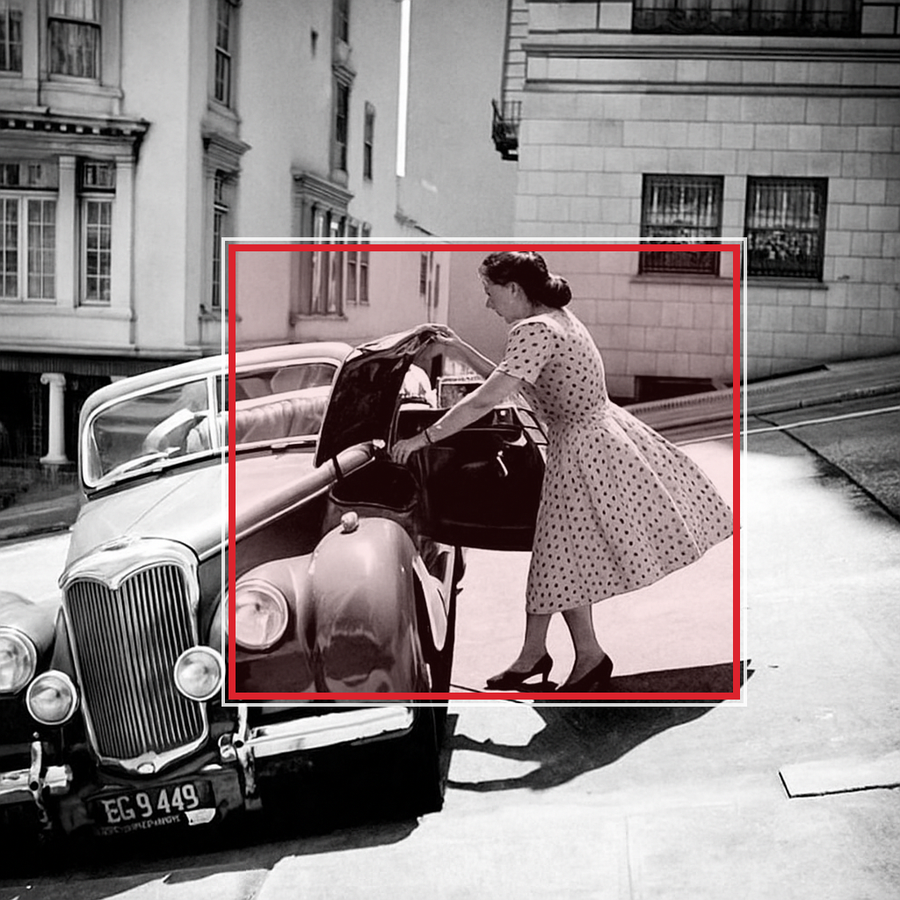}
\end{minipage}

\vspace{6pt}

\Dec{Evidence Gate} ($\mathcal{D}$)

\Dec{Route:} \textbf{DIRECT-ACCEPTANCE}

\Dec{Final explanation $r^{\mathrm{dec}}$:} The two reports give the same TAMPERED answer and the route is DIRECT-ACCEPTANCE. The available support is sufficient for direct acceptance under clean agreement, so image-side VLM recheck is not used. The final assessment is TAMPERED.

\Dec{Final authenticity $\hat{a}$:} \textbf{FAKE} \qquad
\Dec{Final answer $\hat{s}$:} \textbf{TAMPERED} \qquad
\Dec{Final bbox $\hat{B}$:} \textbf{(57, 61, 184, 174)}
\end{tcolorbox}
\end{table*}

\begin{table*}[t]
\centering
\caption{\textbf{Targeted arbitration.} The observer predicts REAL while the provider raises a tentative FULL\_SYNTHETIC claim; targeted arbitration keeps the decision within REAL versus FULL\_SYNTHETIC and accepts REAL. GT is \texttt{REAL}; final prediction is \texttt{REAL}.}
\label{sup:table7}

\begin{tcolorbox}[exbox]
\footnotesize
\setlength{\parskip}{1.5pt}
\noindent
\begin{minipage}[t]{0.74\linewidth}
\Obs{Visual Observer} ($\mathcal{O}$)

\Obs{Reasoning:} The image depicts a black-and-white cat lying on a patterned carpet and holding a small colorful toy with a string in its paw. The cat's fur appears smooth and well-defined, and the carpet texture remains clear across the scene. The background includes a dark piece of furniture, possibly a chair or table, and the lighting appears consistent. Upon closer inspection, the image shows no visible boundary artifacts, blending seams, or regions with inconsistent texture, shadow, or sharpness that would suggest local manipulation or full-image synthesis. The scene is therefore most consistent with a real photograph.

\Obs{Authenticity $\hat{a}^{\mathrm{obs}}$:} \textbf{REAL} \qquad
\Obs{Answer $\hat{s}^{\mathrm{obs}}$:} \textbf{REAL} \qquad
\Obs{Bbox $\hat{B}^{\mathrm{obs}}$:} \textbf{NONE}

\vspace{5pt}

\Evi{Forensic Evidence Provider} ($\mathcal{P}$)

\Evi{Descriptors $\mathbf{d}^{\mathrm{for}}$:} mass $2.64$: anomaly evidence is weak-to-moderate; peak $0.885$: a small number of patches respond strongly; concentration $0.995$: the response is sharply concentrated; support area $0.017$: the activated region is very small; entropy $2.00$: the response is compact rather than diffuse.

\Evi{Basis:} Evidence provider answer is FULL\_SYNTHETIC. The image-level fake read is tentative because confidence or descriptor magnitude is limited. Evidence provider bbox is NONE.

\Evi{Answer $\hat{s}^{\mathrm{for}}$:} \textbf{FULL\_SYNTHETIC} \qquad
\Evi{Confidence $c^{\mathrm{for}}$:} \textbf{0.569} \qquad
\Evi{Bbox $\hat{B}^{\mathrm{for}}$:} \textbf{NONE}
\end{minipage}
\hfill
\begin{minipage}[t]{0.20\linewidth}
\centering
\vspace{0pt}
\includegraphics[height=2.6cm]{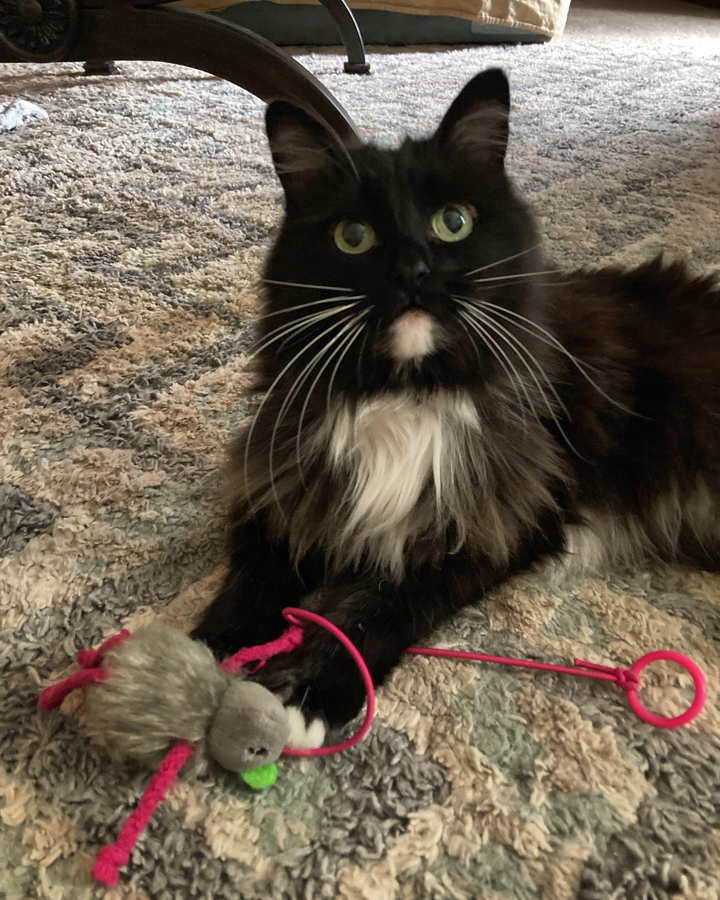}
\end{minipage}

\vspace{6pt}

\Dec{Decision Maker Output} ($\mathcal{D}$)

\Dec{Route:} \textbf{TARGETED-ARBITRATION}

\Dec{Final explanation $r^{\mathrm{dec}}$:} The black and white cat with long fur, yellow eyes, and a pink toy is lying on a textured carpet. The REAL claim is supported by the indoor setting, the cat's relaxed pose, the toy, and the carpet texture. The FULL\_SYNTHETIC claim overcalls the ordinary indoor scene. The FULL\_SYNTHETIC claim fails because the targeted content does not form a coherent whole-image generation failure. The REAL claim is accepted because the scene details remain mutually consistent after the targeted arbitration. The final assessment is REAL.

\Dec{Final authenticity $\hat{a}$:} \textbf{REAL} \qquad
\Dec{Final answer $\hat{s}$:} \textbf{REAL} \qquad
\Dec{Final bbox $\hat{B}$:} \textbf{NONE}
\end{tcolorbox}
\end{table*}

\begin{table*}[t]
\centering
\caption{\textbf{Full arbitration.} The observer treats a social-media screenshot as REAL, while the provider detects a bounded anomaly; full arbitration accepts TAMPERED with a localized box. GT is \texttt{TAMPERED}; final prediction is \texttt{TAMPERED}.}
\label{sup:table8}

\begin{tcolorbox}[exbox]
\footnotesize
\setlength{\parskip}{1.5pt}
\noindent
\begin{minipage}[t]{0.74\linewidth}
\Obs{Visual Observer} ($\mathcal{O}$)

\Obs{Reasoning:} The image shows a social-media post by Mark Zuckerberg, with a profile picture and a timestamp indicating that the post was made 35 minutes ago. The post highlights two pieces of information: the most liked Instagram post by Leo Messi and a record number of WhatsApp messages sent during the World Cup final. The layout is typical of a social-media platform, with engagement metrics such as likes, comments, and shares displayed below the post. The image appears to be a standard screenshot of a real social-media interface, and there are no obvious visual seams, inconsistent fonts, distorted UI elements, or localized artifacts that would indicate manipulation.

\Obs{Authenticity $\hat{a}^{\mathrm{obs}}$:} \textbf{REAL} \qquad
\Obs{Answer $\hat{s}^{\mathrm{obs}}$:} \textbf{REAL} \qquad
\Obs{Bbox $\hat{B}^{\mathrm{obs}}$:} \textbf{NONE}

\vspace{5pt}

\Evi{Forensic Evidence Provider} ($\mathcal{P}$)

\Evi{Descriptors $\mathbf{d}^{\mathrm{for}}$:} mass $9.84$: anomaly evidence is substantial; peak $1.000$: at least one region has a saturated response; concentration $1.000$: energy is concentrated in high-response patches; support area $0.008$: the candidate region is very small and bounded; entropy $2.53$: the response remains spatially organized.

\Evi{Basis:} Evidence provider answer is TAMPERED. The local anomaly is bounded, but confidence or descriptor strength keeps the provider support moderate. Evidence provider bbox is (14, 13, 28, 55).

\Evi{Answer $\hat{s}^{\mathrm{for}}$:} \textbf{TAMPERED} \qquad
\Evi{Confidence $c^{\mathrm{for}}$:} \textbf{0.880} \qquad
\Evi{Bbox $\hat{B}^{\mathrm{for}}$:} \textbf{(14, 13, 28, 55)}
\end{minipage}
\hfill
\begin{minipage}[t]{0.20\linewidth}
\centering
\vspace{0pt}
\includegraphics[height=2.6cm]{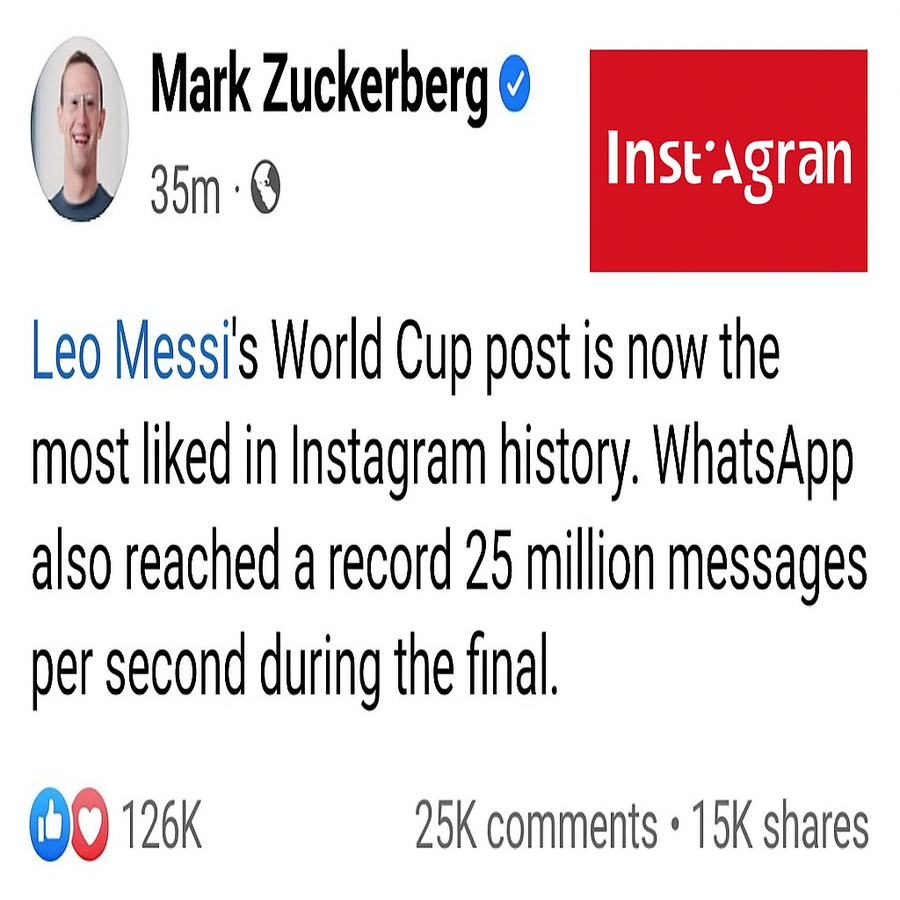}

\vspace{3pt}

\includegraphics[height=2.6cm]{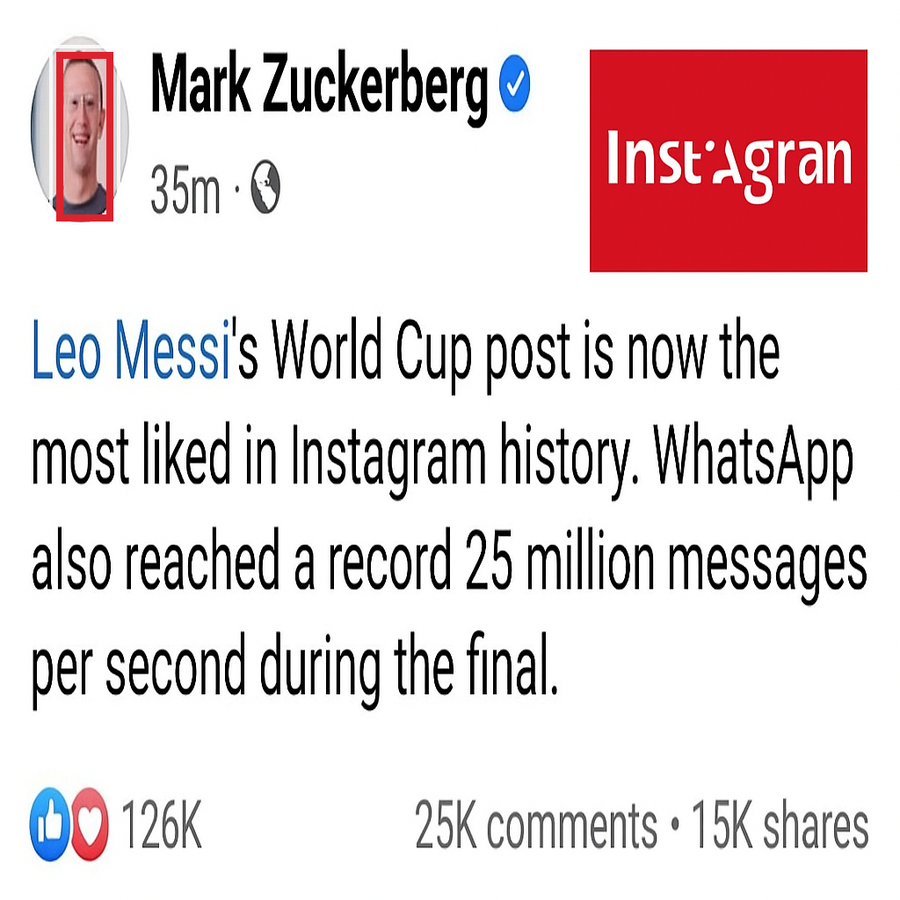}
\end{minipage}

\vspace{6pt}

\Dec{Decision Maker Output} ($\mathcal{D}$)

\Dec{Route:} \textbf{FULL-ARBITRATION}

\Dec{Final explanation $r^{\mathrm{dec}}$:} This is a FULL-ARBITRATION case: one evidence record frames the image as authentic while the other claims a local edit, so the recheck must resolve both authenticity and local scope instead of accepting either side directly. The Instagram post by Mark Zuckerberg is a real screenshot, but the local anomaly belongs to the post itself rather than the surrounding content. The profile picture, timestamp, and engagement metrics remain coherent with the post. The TAMPERED claim is supported because the fake evidence is localized within the post content. The REAL claim misses the local edit area.

\Dec{Final authenticity $\hat{a}$:} \textbf{FAKE} \qquad
\Dec{Final answer $\hat{s}$:} \textbf{TAMPERED} \qquad
\Dec{Final bbox $\hat{B}$:} \textbf{(14, 13, 28, 55)}
\end{tcolorbox}
\end{table*}

\begin{table*}[t]
\centering
\caption{\textbf{Full arbitration.} The observer predicts REAL and the provider predicts TAMPERED, but image-side arbitration rejects both and outputs FULL\_SYNTHETIC. GT is \texttt{FULL\_SYNTHETIC}; final prediction is \texttt{FULL\_SYNTHETIC}.}
\label{sup:table9}

\begin{tcolorbox}[exbox]
\footnotesize
\setlength{\parskip}{1.5pt}
\noindent
\begin{minipage}[t]{0.74\linewidth}
\Obs{Visual Observer} ($\mathcal{O}$)

\Obs{Reasoning:} Observer answer is REAL. The painting depicts a serene landscape with a body of water in the background, surrounded by rolling hills and scattered trees. The foreground features a field of wildflowers, and the sky is filled with soft, diffused clouds. The frame is ornate, suggesting a traditional style. Upon closer inspection, there are no visible boundary artifacts, blending seams, or regions with differing textures or edges. The overall appearance is consistent throughout.

\Obs{Authenticity $\hat{a}^{\mathrm{obs}}$:} \textbf{REAL} \qquad
\Obs{Answer $\hat{s}^{\mathrm{obs}}$:} \textbf{REAL} \qquad
\Obs{Bbox $\hat{B}^{\mathrm{obs}}$:} \textbf{NONE}

\vspace{5pt}

\Evi{Forensic Evidence Provider} ($\mathcal{P}$)

\Evi{Descriptors $\mathbf{d}^{\mathrm{for}}$:} mass $3.61$: anomaly evidence is limited; peak $0.984$: a local patch responds strongly; concentration $1.000$: the response is highly concentrated; support area $0.000$: the support is too small to explain a broad local edit; entropy $1.63$: the response is narrow and sparse.

\Evi{Basis:} Evidence provider answer is TAMPERED. The provider class points to a local edit, but the local-edit evidence is weak despite the proposed bbox. Evidence provider bbox is (58, 92, 182, 180).

\Evi{Answer $\hat{s}^{\mathrm{for}}$:} \textbf{TAMPERED} \qquad
\Evi{Confidence $c^{\mathrm{for}}$:} \textbf{0.600} \qquad
\Evi{Bbox $\hat{B}^{\mathrm{for}}$:} \textbf{(58, 92, 182, 180)}
\end{minipage}
\hfill
\begin{minipage}[t]{0.20\linewidth}
\centering
\vspace{0pt}
\includegraphics[height=2.6cm]{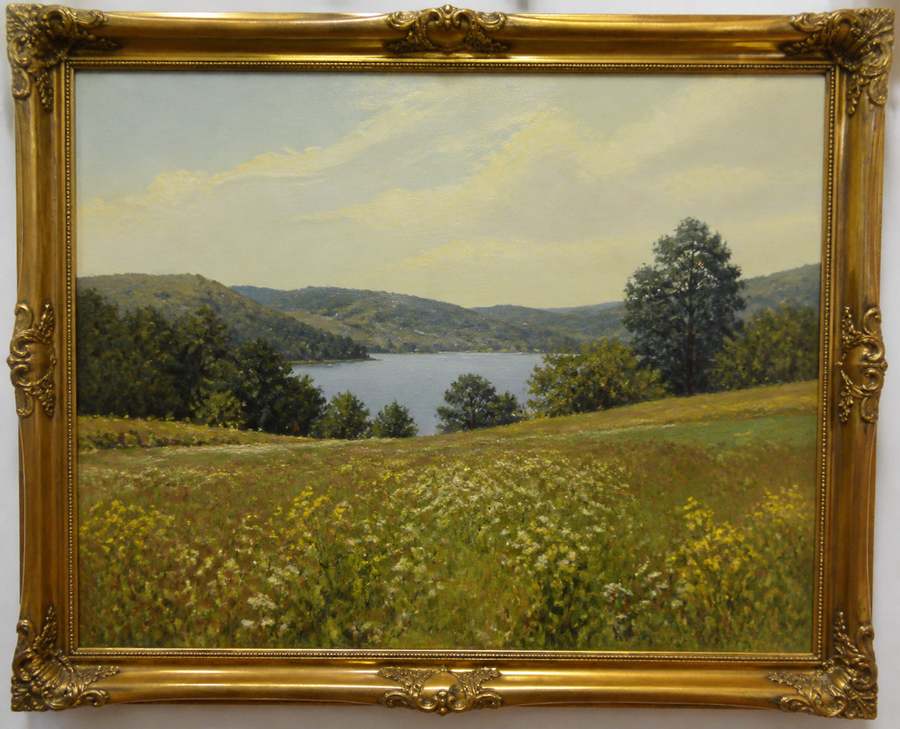}

\vspace{3pt}

\includegraphics[height=2.6cm]{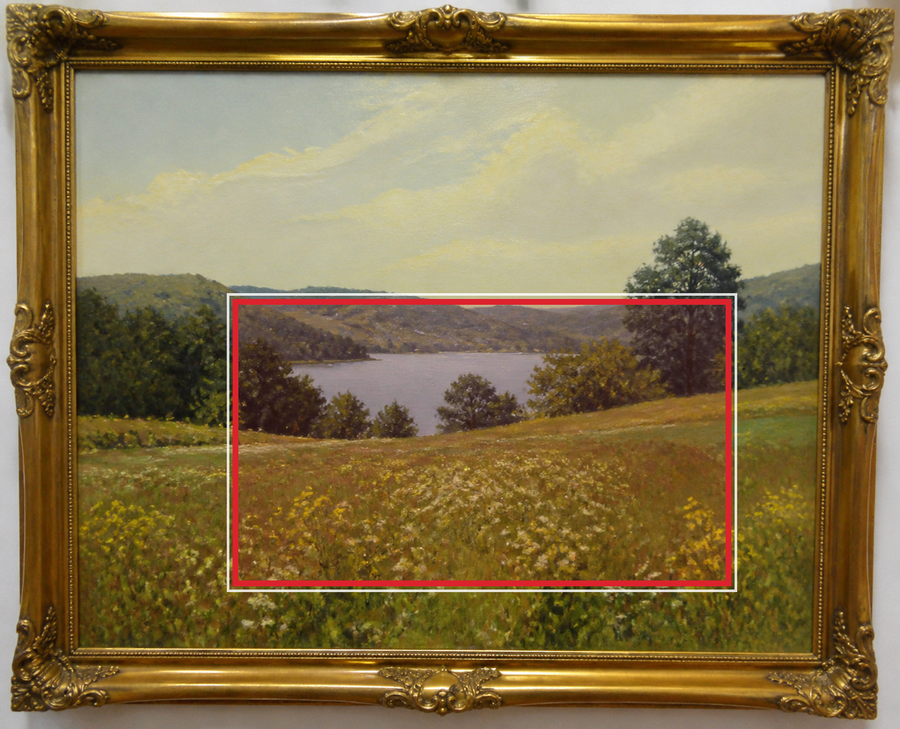}
\end{minipage}

\vspace{6pt}

\Dec{Decision Maker Output} ($\mathcal{D}$)

\Dec{Route:} \textbf{FULL-ARBITRATION}

\Dec{Final explanation $r^{\mathrm{dec}}$:} This is a FULL-ARBITRATION case: one evidence record frames the image as authentic while the other claims a local edit, so the decision must resolve both authenticity and local scope instead of accepting either side directly. The landscape painting has a gold frame, a grassy foreground, a lake, distant hills, and a cloudy sky. The REAL claim treats the frame and landscape as a coherent photograph, but this ordinary appearance does not rule out whole-image synthesis. The TAMPERED claim points to a local box, but that box is too small to explain the image-wide synthetic pattern. The final decision rejects both evidence records and assigns FULL\_SYNTHETIC because the suspicious cues span the framed object and the painted scene rather than one separable edited region.

\Dec{Final authenticity $\hat{a}$:} \textbf{FAKE} \qquad
\Dec{Final answer $\hat{s}$:} \textbf{FULL\_SYNTHETIC} \qquad
\Dec{Final bbox $\hat{B}$:} \textbf{NONE}
\end{tcolorbox}
\end{table*}

\begin{table*}[t]
\centering
\caption{\textbf{Failure case.} Both evidence records predict TAMPERED on a REAL image, so the gate accepts the wrong consensus without calling the Decision Maker. GT is \texttt{REAL}; final prediction is \texttt{TAMPERED}.}
\label{sup:table10}

\begin{tcolorbox}[exbox]
\footnotesize
\setlength{\parskip}{1.5pt}
\noindent
\begin{minipage}[t]{0.74\linewidth}
\Obs{Visual Observer} ($\mathcal{O}$)

\Obs{Reasoning:} Observer answer is TAMPERED. The image depicts a man holding a young girl outdoors. The observer focuses on the girl's dress and reports texture and sharpness differences around that region, interpreting it as a local manipulation.

\Obs{Authenticity $\hat{a}^{\mathrm{obs}}$:} \textbf{FAKE} \qquad
\Obs{Answer $\hat{s}^{\mathrm{obs}}$:} \textbf{TAMPERED} \qquad
\Obs{Bbox $\hat{B}^{\mathrm{obs}}$:} \textbf{(84, 15, 191, 224)}

\vspace{5pt}

\Evi{Forensic Evidence Provider} ($\mathcal{P}$)

\Evi{Descriptors $\mathbf{d}^{\mathrm{for}}$:} mass $7.27$: anomaly evidence is noticeable; peak $0.971$: at least one region responds strongly; concentration $0.990$: the response is concentrated in high-activation patches; support area $0.189$: the candidate region covers a moderate part of the image; entropy $2.75$: the response remains spatially organized.

\Evi{Basis:} Evidence provider answer is TAMPERED. The local anomaly is bounded, but confidence or descriptor strength keeps the provider support moderate. Evidence provider bbox is (84, 15, 191, 224).

\Evi{Answer $\hat{s}^{\mathrm{for}}$:} \textbf{TAMPERED} \qquad
\Evi{Confidence $c^{\mathrm{for}}$:} \textbf{0.786} \qquad
\Evi{Bbox $\hat{B}^{\mathrm{for}}$:} \textbf{(84, 15, 191, 224)}
\end{minipage}
\hfill
\begin{minipage}[t]{0.20\linewidth}
\centering
\vspace{0pt}
\includegraphics[height=2.6cm]{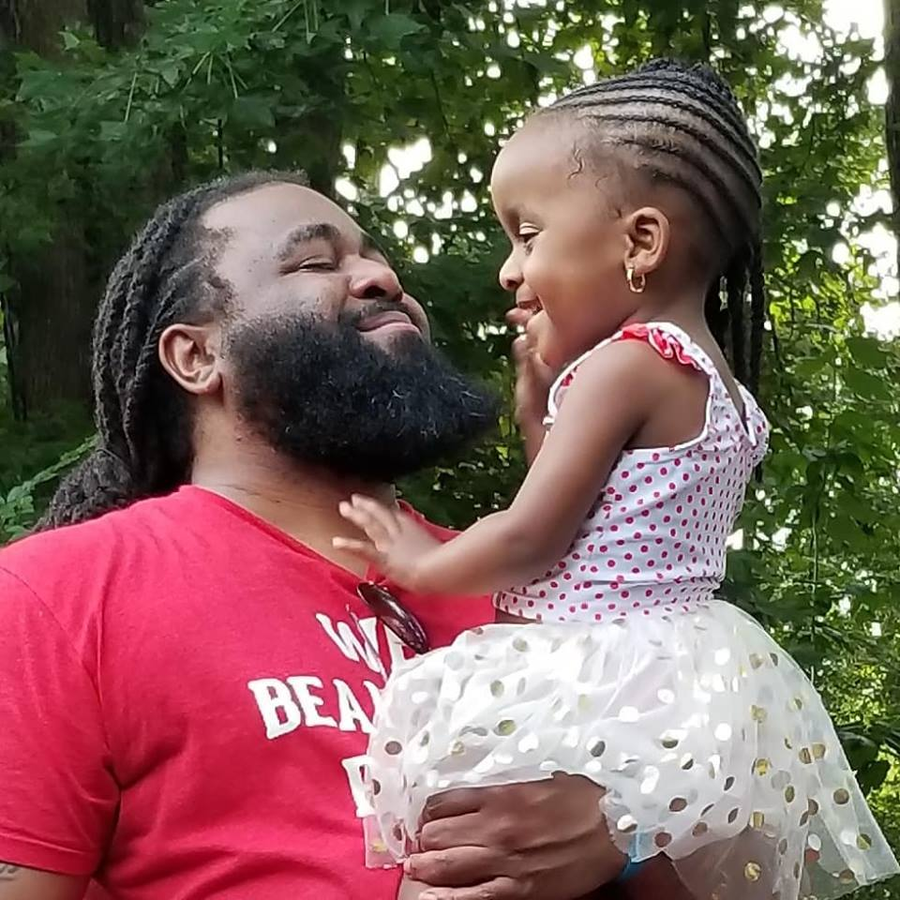}

\vspace{3pt}

\includegraphics[height=2.6cm]{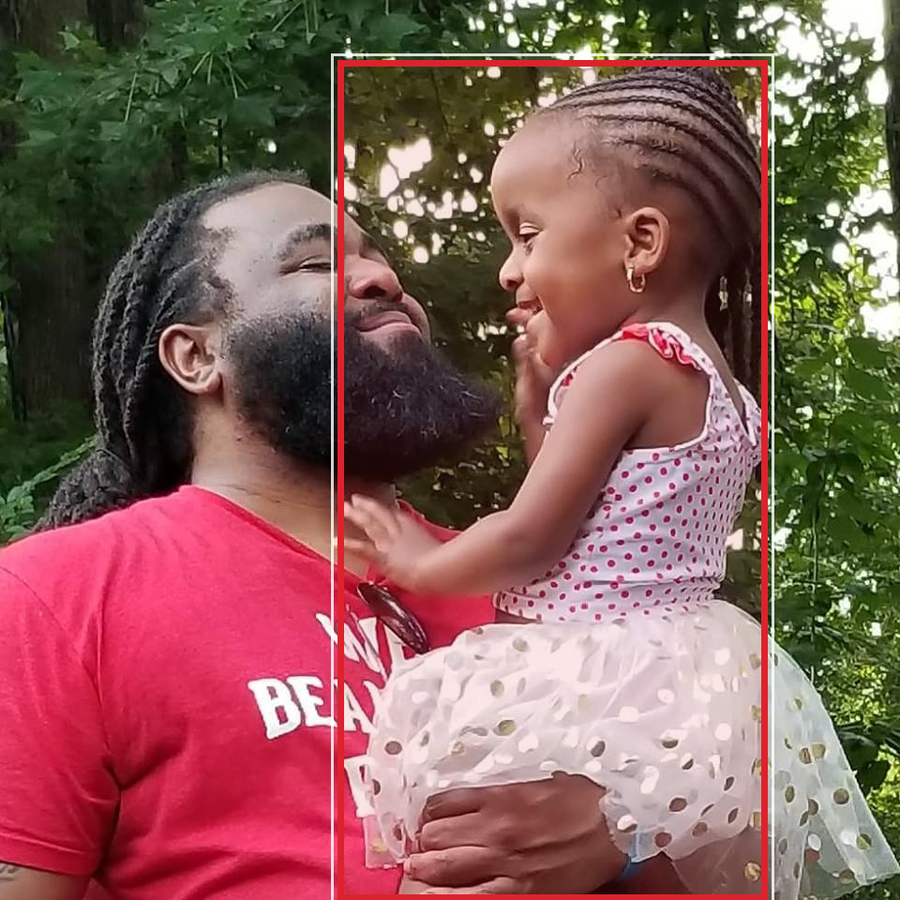}
\end{minipage}

\vspace{6pt}

\Dec{Evidence Gate} ($\mathcal{D}$)

\Dec{Route:} \textbf{DIRECT-ACCEPTANCE}

\Dec{Final explanation $r^{\mathrm{dec}}$:} The two reports give the same TAMPERED answer and the route is DIRECT-ACCEPTANCE. The available support is sufficient for direct acceptance under clean agreement, so image-side VLM recheck is not used. The final assessment is TAMPERED.

\Dec{Final authenticity $\hat{a}$:} \textbf{FAKE} \qquad
\Dec{Final answer $\hat{s}$:} \textbf{TAMPERED} \qquad
\Dec{Final bbox $\hat{B}$:} \textbf{(84, 15, 191, 224)}
\end{tcolorbox}
\end{table*}

\end{document}